\documentclass[lettersize,journal]{IEEEtran}
\usepackage{amsmath,amsfonts,amssymb}
\usepackage{algorithmic}
\usepackage{array}
\usepackage[caption=false,font=scriptsize,labelfont=sf,textfont=sf]{subfig} 
\usepackage{textcomp}
\usepackage{stfloats}
\usepackage{url}
\usepackage{verbatim}
\usepackage{graphicx}

\usepackage{tabularx}
\usepackage{marvosym}
\usepackage[dvipsnames, table]{xcolor}

\usepackage{cite}
\usepackage[colorlinks=true, urlcolor=Black,linkcolor=green, citecolor=green]{hyperref} 
\usepackage{orcidlink}
\usepackage{braket}
\usepackage{booktabs}
\usepackage{threeparttable}
\usepackage{multirow} 
\usepackage{makecell}
\usepackage{pifont}
\usepackage[normalem]{ulem}
\definecolor{darkgreen}{RGB}{0,128,0}  

\newcommand{\hhline}{\noalign{\vskip 1pt}\hline\noalign{\vskip 1pt}}

\newcommand{\degree}{^\circ}

\def\BibTeX{{\rm B\kern-.05em{\sc i\kern-.025em b}\kern-.08em
    T\kern-.1667em\lower.7ex\hbox{E}\kern-.125emX}}
    
\usepackage{balance}
\begin{document}
\title{Fixed-Length Dense Fingerprint Representation with Alignment and Robust Enhancement} 
\author{Zhiyu Pan$^{\orcidlink{0009-0000-6721-4482}}$,
  Xiongjun Guan$^{\orcidlink{0000-0001-8887-3735}}$, 
  Yongjie Duan$^{\orcidlink{0000-0003-3741-9596}}$,
	Jianjiang Feng$^{\orcidlink{0000-0003-4971-6707}}$, ~\IEEEmembership{Member, IEEE}, 
	and Jie Zhou$^{\orcidlink{0000-0001-7701-234X}}$, ~\IEEEmembership{Fellow, IEEE}
  \thanks{
    This work was supported in part by the National Natural Science Foundation of China under Grant 62376132 and 62321005. (\emph{Corresponding author: Jianjiang Feng}.)}
  \IEEEcompsocitemizethanks{
  \IEEEcompsocthanksitem
  The authors are with Department of Automation, Tsinghua University, Beijing 100084, China (e-mail: \url{pzy20@mails.tsinghua.edu.cn}; \url{gxj21@mails.tsinghua.edu.cn};
  \url{duanyj13@tsinghua.org.cn};  \url{jfeng@tsinghua.edu.cn}; \url{jzhou@tsinghua.edu.cn}).}
  \thanks{This paper has supplementary downloadable material available at \url{http://ieeexplore.ieee.org}., provided by the author. The material includes a supplementary PDF providing extended implementation details, network architectures, and additional experimental evaluations. Contact \url{pzy20@mails.tsinghua.edu.cn} for further questions about this work.}
}



\markboth{Journal of \LaTeX\ Class Files,~Vol.~18, No.~9, May~2025}%
{How to Use the IEEEtran \LaTeX \ Templates}

\maketitle

\begin{abstract}
  Fixed-length fingerprint representations, which map each fingerprint to a compact and fixed-size feature vector, are computationally efficient and well-suited for large-scale matching. However, designing a robust representation that effectively handles diverse fingerprint modalities, pose variations, and noise interference remains a significant challenge. In this work, we propose a fixed-length dense descriptor of fingerprints, and introduce FLARE—a fingerprint matching framework that integrates the \underline{F}ixed-\underline{L}ength dense descriptor with pose-based \underline{A}lignment and \underline{R}obust \underline{E}nhancement. This fixed-length representation employs a three-dimensional dense descriptor to effectively capture spatial relationships among fingerprint ridge structures, enabling robust and locally discriminative representations. To ensure consistency within this dense feature space, FLARE incorporates pose-based alignment using complementary estimation methods, along with dual enhancement strategies that refine ridge clarity while preserving the original fingerprint modality. The proposed dense descriptor supports fixed-length representation while maintaining spatial correspondence, enabling fast and accurate similarity computation. Extensive experiments demonstrate that FLARE achieves superior performance across rolled, plain, latent, and contactless fingerprints, significantly outperforming existing methods in cross-modality and low-quality scenarios. Further analysis validates the effectiveness of the dense descriptor design, as well as the impact of alignment and enhancement modules on the accuracy of dense descriptor matching. Experimental results highlight the effectiveness and generalizability of FLARE as a unified and scalable solution for robust fingerprint representation and matching. The implementation and code will be publicly available at our \href{https://github.com/Yu-Yy/FLARE}{GitHub repository}.
\end{abstract}

\begin{IEEEkeywords}
  Fingerprint recognition, fixed-length fingerprint representation, dense descriptor, fingerprint enhancement, pose-based alignment.
\end{IEEEkeywords}

\section{Introduction}
\IEEEPARstart{F}INGERPRINT as a biometric trait has several notable advantages, including ease of acquisition, high permanence, and enhanced privacy, making it broadly applicable in civilian and commercial fields \cite{maltoni2022handbook}. A typical fingerprint recognition system comprises three key components: image acquisition, feature extraction, and matching \cite{jain2004intro}. Acquisition methods and modalities vary significantly across different application scenarios and are commonly classified as rolled, plain, latent, and contactless fingerprints. Consequently, robust and efficient feature extraction and matching algorithms are essential to ensure reliable recognition performance across varying acquisition conditions and fingerprint types. Initial research in fingerprint recognition primarily focused on designing handcrafted descriptors using level-1 (e.g., orientation field, frequency map) and level-2 (e.g., ridge skeleton map, minutiae) fingerprint features \cite{jain1999fingercode,wang2007fingerprint,Kumar2011,bringer2010binary,cappelli2010minutia}. However, these conventional methods often struggle with partial or noisy fingerprints and exhibit limited generalization across fingerprint modalities. Benefiting from the efficiency and strong discriminative power of deep feature learning, deep learning has become the dominant paradigm in modern fingerprint matching systems.

\begin{figure}
  \centering
  \includegraphics[width=\linewidth]{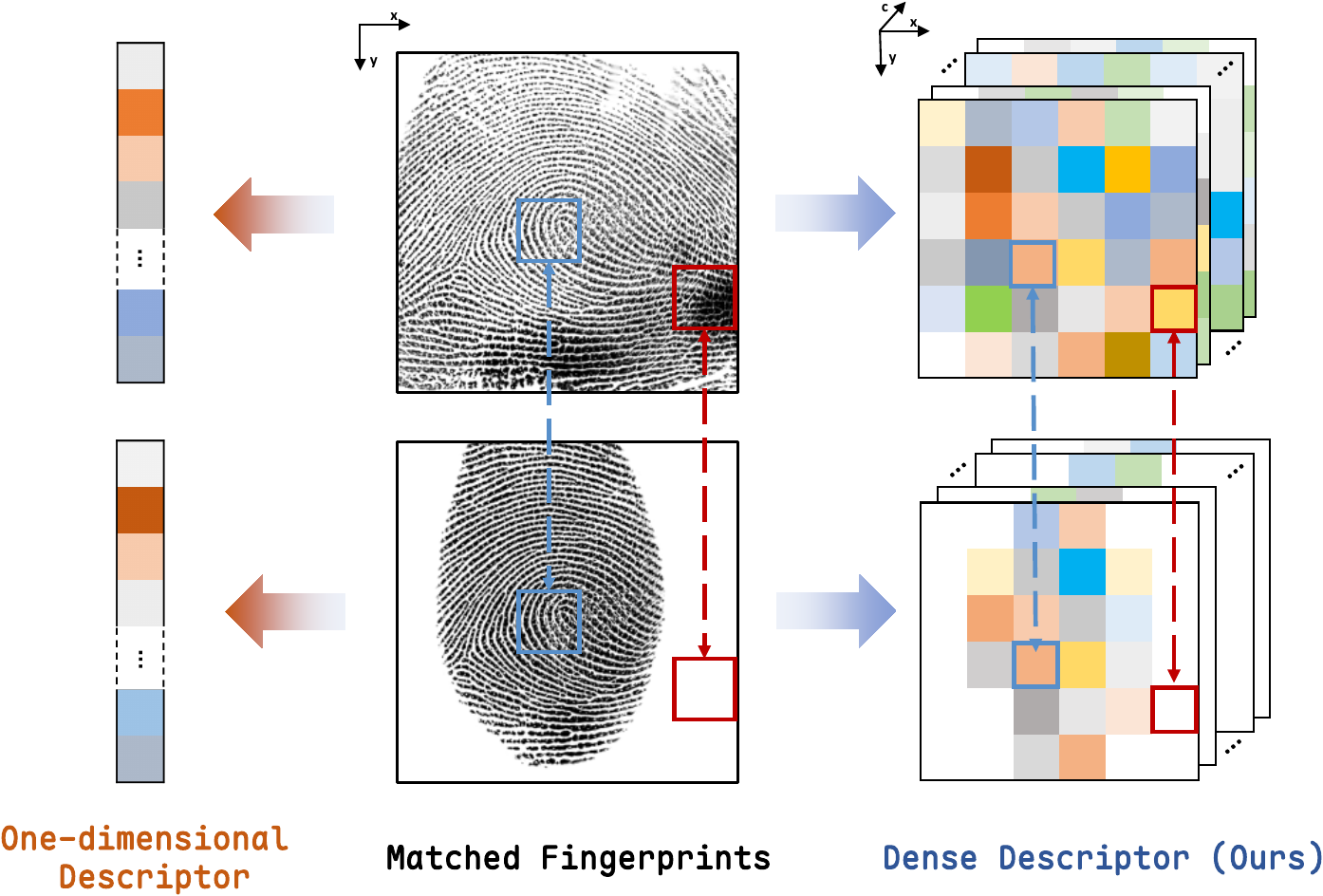}
  \caption{Comparison of one-dimensional and dense descriptors. One-dimensional descriptors lose spatial structural information, whereas dense descriptors maintain spatial correspondence, enhancing local sensitivity and discriminative power while mitigating background noise.}
  \label{fig:desc_compare}
\end{figure}

We categorize fingerprint matching methods based on deep learning into three types. The first is pairwise matching networks, which take two fingerprint images as input and directly produce a similarity score by jointly processing their features \cite{PFVNet, ifvit, guan2025joint}. While this enables fine-grained comparisons and high accuracy, the joint processing is computationally expensive and unsuitable for large-scale identification. The second category is local representation matching, where local descriptors are extracted from patches centered at detected minutiae or estimated orientation fields \cite{cao2019automated, MinNet, pan2024latent}. Each fingerprint is represented by a variable-length set of descriptors, and matching is performed by computing pairwise similarities. Although this method offers good accuracy, its efficiency is limited by the variable-length nature of the representation. The third is fixed-length representation matching, where a neural network encodes each fingerprint into a global fixed-length feature vector \cite{cao2017fingerprint,PDC, DeepPrint,grosz2024afrnet}. This allows for fast, many-to-many comparisons via matrix operations, making it highly efficient and scalable for large-scale matching. Nonetheless, the matching accuracy of fixed-length representations tends to degrade in challenging scenarios, such as latent fingerprints or partial and low-quality impressions with missing or distorted ridge structures \cite{grosz2023latent}. Motivated by these observations, we present FLARE — a fingerprint matching framework that combines a Fixed-Length dense descriptor with pose-based ridge Alignment and Robust Enhancement. Our approach preserves the efficiency of fixed-length matching while significantly improving robustness and accuracy under challenging fingerprint conditions, thereby enhancing overall generalizability.

Most existing fixed-length methods represent fingerprint descriptors as one-dimensional vectors\cite{DeepPrint,grosz2024afrnet,grosz2023latent}, which often fail to suppress interference from background regions (Fig. \ref{fig:desc_compare} left). As a result, their matching accuracy degrades when there are significant differences in fingerprint foreground areas or in the presence of strong background noise. Some approaches \cite{gu2022latent, zhang2023robust} attempt to mitigate these challenges by incorporating foreground attention masks during representation learning or by dividing fingerprint images into multiple regions for feature extraction and fusion. However, these strategies operate at the feature or region level and do not explicitly suppress background interference or adapt to variations across different foreground areas at the descriptor level. To overcome these limitations, we build upon the form of dense representation \cite{pan2024latent} and proposed a fixed-length dense descriptor, which serves as a localized deep representation of the fingerprint in the form of a three-dimensional tensor, where two spatial dimensions align with the original image coordinates (Fig. \ref{fig:desc_compare} right). The dense descriptor is defined only within the fingerprint foreground region, while background areas are left empty, effectively eliminating background interference. When two fingerprints are spatially aligned, matching is performed only within the overlapping foreground regions of the dense descriptors, allowing the method to handle partial fingerprints more effectively. Therefore, the dense representation offers enhanced robustness in challenging scenarios such as low-quality or incomplete fingerprints. 

Since dense descriptor matching relies on accurate spatial alignment, misalignment between fingerprint pairs can significantly degrade performance. To address this, FLARE adopts a pose-based alignment strategy built on 2D fingerprint pose estimation \cite{si2015detection, ouyang2017pose, yin2021joint, duan2023estimating, guan2025robust}. Each fingerprint is normalized into a unified coordinate system by aligning its estimated central location and orientation, requiring only a single transformation per image. This alignment strategy preserves the efficiency of fixed-length representation matching while enhancing robustness to pose variations. To improve alignment reliability under varying fingerprint quality and patterns, FLARE further integrates two complementary state-of-the-art pose estimation methods: the dense voting approach of Duan et al. \cite{duan2023estimating} and the region detection-based regression method of Guan et al. \cite{guan2025robust}. The former aggregates local predictions from ridge-level features, while the latter directly regresses global pose from coarse structural cues. By combining these two approaches, FLARE leverages both fine-grained local details and global spatial structure to achieve more robust and consistent alignment across fingerprint modalities.

\begin{figure}[t] 
  \centering
  \includegraphics[width=\linewidth]{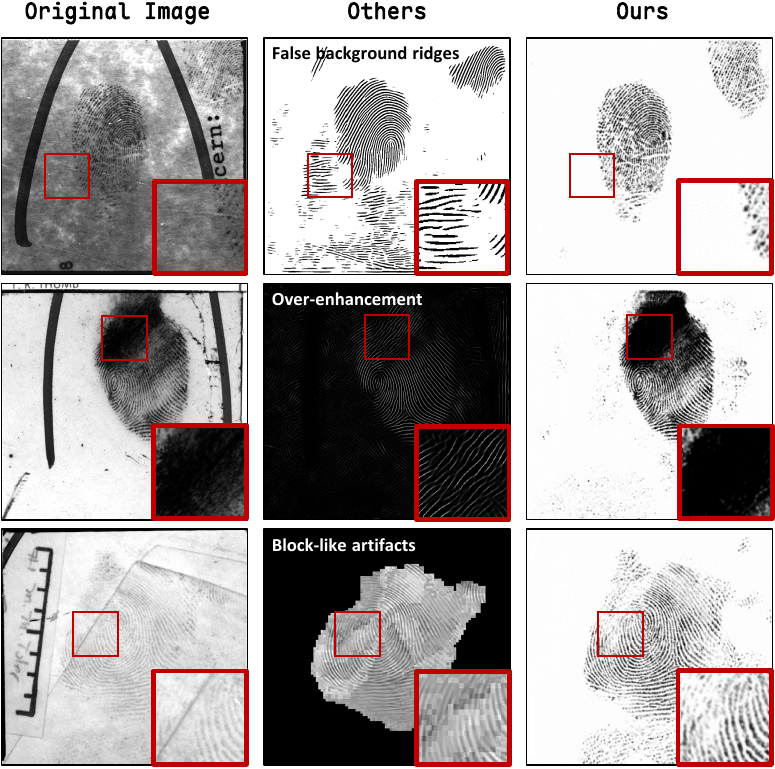}
  \caption{Visual comparison of enhancement methods. Each row shows an original fingerprint (left), the result from a representative existing method (middle), and our enhancement (right). Compared to previous methods, which may hallucinate background ridges (top) \cite{nist2020verifinger}, over-generate ridges in blurry regions (middle) \cite{FingerGAN}, or exhibit block artifacts (bottom) \cite{tang2017fingernet}, our approach preserves the original impression style, suppresses background interference, and avoids introducing spurious textures.}
  \label{fig:enh_comp_illus}
\end{figure}

Although dense descriptors are inherently robust to background clutter and partial observations, their performance can benefit from higher ridge clarity and more uniform local patterns. To this end, FLARE incorporates fingerprint enhancement that improves image quality by increasing ridge contrast and reducing common artifacts, thereby providing cleaner inputs that support more stable dense descriptor extraction. However, many existing enhancement methods \cite{tang2017fingernet, huang2020enh, nist2020verifinger, FingerGAN} transform fingerprint images into a new modality that retains only ridge-like structures. These approaches often rely on perceptual estimations of ridge patterns, which can be unreliable in noisy or low-quality regions. As a result, they may hallucinate artificial ridges or inadvertently remove valid fingerprint areas, reducing compatibility with descriptor networks. In contrast, our enhancement strategy directly operates on the original fingerprint texture, suppressing noise and improving contrast without altering ridge structures. This preserves structural fidelity and ensures compatibility with dense descriptor learning (Fig.~\ref{fig:enh_comp_illus}). Drawing inspiration from LFRNet \cite{grosz2023latent}, we propose UNetEnh, a UNet-based model \cite{ronneberger2015u} trained on simulated low-quality fingerprints. In addition, we introduce PriorEnh, which leverages a VQ-VAE \cite{vqvae} to learn a latent ridge-structure codebook from high-quality fingerprints. This prior guides the enhancement process and enables partial ridge reconstruction in degraded regions. Experimental results show that PriorEnh improves matching across a variety of fingerprint types, while UNetEnh performs particularly well on latent fingerprints. To exploit their complementary strengths, FLARE combines both enhancement methods, leading to improved robustness and accuracy.

We conduct comprehensive experiments across diverse fingerprint types and datasets, demonstrating that FLARE consistently outperforms existing fixed-length methods in matching accuracy. A major source of FLARE`s improvement lies in its fixed-length dense representation, which encodes spatially localized ridge patterns in a compact and structurally expressive form, offering strong robustness to noise, partial observations, and modality variations. Building on this foundation, we introduce a refined enhancement philosophy and develop new enhancement modules that improve ridge clarity while preserving the original image modality. In addition, we explore the complementary behavior of different pose estimation and enhancement strategies, forming a unified preprocessing pipeline that yields cleaner inputs for descriptor extraction and further improves matching accuracy. 
This work extends our previous conference paper FDD \cite{FDD} by incorporating complementary pose estimation methods, introducing newly designed fingerprint enhancement modules, and exploring their synergistic integration to strengthen the preprocessing pipeline. We also provide a more comprehensive analysis of the fixed-length dense descriptor through expanded experiments and ablation studies. Moreover, our experimental results show that FLARE consistently outperforms FDD in matching accuracy, with particularly notable gains on low-quality fingerprint datasets.
Concretely, the contributions of this research are as follows:
\begin{itemize}
  \item We introduce a fixed-length dense descriptor that preserves the spatial organization of fingerprint ridges in a compact 3D representation, enabling effective encoding of foreground-discriminative patterns while naturally downweighting background regions.
  \item We design two enhancement modules, UNetEnh and PriorEnh, which operate directly on the original fingerprint texture to reduce background noise interference and improve ridge clarity while preserving structural fidelity.
  \item We present FLARE, a fingerprint recognition framework built upon dense feature representation and strengthened by a preprocessing pipeline that integrates complementary pose estimation and enhancement methods, producing cleaner standardized inputs and boosting descriptor-level matching performance.
  \item We conduct comprehensive experiments and ablation studies across diverse fingerprint types and conditions, demonstrating the effectiveness of the dense representation, the advantages of the proposed enhancement modules, and the additional performance gains achieved through the integrated design of FLARE.
\end{itemize}

\section{Related Work}
\subsection{Fixed-length Fingerprint Representation}
Fixed-length fingerprint representations encode an entire fingerprint image into a compact vector of fixed dimension, allowing similarity computation via distance metrics. With the rise of deep learning, fixed-length representations have become more robust and discriminative. Song and Feng \cite{song2017fingerprint} proposed integrating multi-scale representations from pyramid deep convolutional features. Cao et al. \cite{cao2017fingerprint} introduced Spatial Transformer Networks (STNs) to assist fingerprint rectification. Engelsma et al. \cite{DeepPrint} further integrated STNs with compact descriptor networks, incorporating minutiae-aware features. Grosz et al. \cite{grosz2024afrnet} combined CNN and Transformer features under STN-based alignment to improve performance. However, these STNs are typically trained without ground-truth pose supervision, relying instead on matching losses. Gu et al. \cite{gu2022latent} addressed this by training a pose estimator with annotated 2D poses and extracting multi-scale features from different fingerprint regions. Another line of work seeks alignment-free representations by improving robustness to pose variations, such as fusing multiple local representations aligned using minutiae anchors \cite{song2019aggregating} or applying minutiae-supervised pooling \cite{wu2022minutiae}. However, these approaches often rely strongly on the accuracy of minutiae extraction, making them sensitive to the reliability of extracted minutiae and the availability of sufficient foreground ridge regions. In contrast, FLARE adopts the proposed pose-supervised dense fixed-length representation, which addresses these limitations by preserving spatial correspondence and suppressing background noise, thereby enhancing robustness and matching accuracy.

\subsection{Fingerprint 2D Pose Estimation}
Unlike other biometric traits such as face or palmprint, fingerprints lack stable anatomical landmarks for defining a 2D pose. Early works \cite{nilsson2003localization, jiang2004reference, cappelli2009spatial, yang2014localized} attempted to estimate pose based on singular points, ridge curvature, or contours, but the diversity of fingerprint patterns and sensitivity to ridge quality limited their generalization \cite{maltoni2022handbook}. Following the pose definition of Si et al. \cite{si2015detection}, where the center and orientation are derived from ridge flow characteristics, existing pose estimation methods can be broadly categorized into two types: global regression-based \cite{DeepPrint,ouyang2017pose,guan2025robust} and local voting-based approaches \cite{gu2018efficient,duan2023estimating}. Regression-based methods estimate the overall pose from the entire image, while voting-based methods aggregate local predictions from fingerprint patches. In FLARE, we integrate a global regression-based estimator\cite{guan2025robust} with a local voting-based method \cite{duan2023estimating} to leverage their complementary strengths for improving matching performance.

\subsection{Fingerprint Enhancement}
Fingerprint enhancement improves ridge clarity and supports better matching, particularly for low-quality images. Classical methods often rely on Gabor filtering. Cappelli et al. \cite{cappelli2009semi} used adaptive Gabor filters for local orientation and frequency; Feng et al. \cite{feng2013orientation} and Yang et al. \cite{yang2014localized} improved robustness by introducing global and local orientation dictionaries combined with pose estimation. With deep learning, Tang et al. \cite{tang2017fingernet} modeled Gabor filtering as learnable convolutions to enhance latent fingerprints. GAN-based approaches have also emerged: Huang et al. \cite{huang2020enh} adopted a PatchGAN to refine latent fingerprints; Zhu et al. \cite{FingerGAN} framed enhancement as a modality translation task using FOMFE \cite{wang2007fingerprint} and minutiae supervision. These methods, while effective for traditional minutiae-based matching, often transform the fingerprint into a new modality focused solely on ridge perception. This transformation risks introducing hallucinated patterns that compromise the quality of global deep representation extraction. Moreover, such methods exhibit limited robustness to structural variations and noise artifacts. In contrast, we argue that fingerprint enhancement should operate within the original image modality, preserving structural fidelity while suppressing background noise and enhancing ridge contrast. Following this principle, we develop two complementary enhancement modules integrated in FLARE: UNetEnh, a denoising-based model, and PriorEnh, a prior-guided reconstruction model learned from high-quality fingerprints. Their fusion further improves robustness and matching performance across fingerprint modalities.

\begin{figure*}
  \centering
  \includegraphics[width=.9\linewidth]{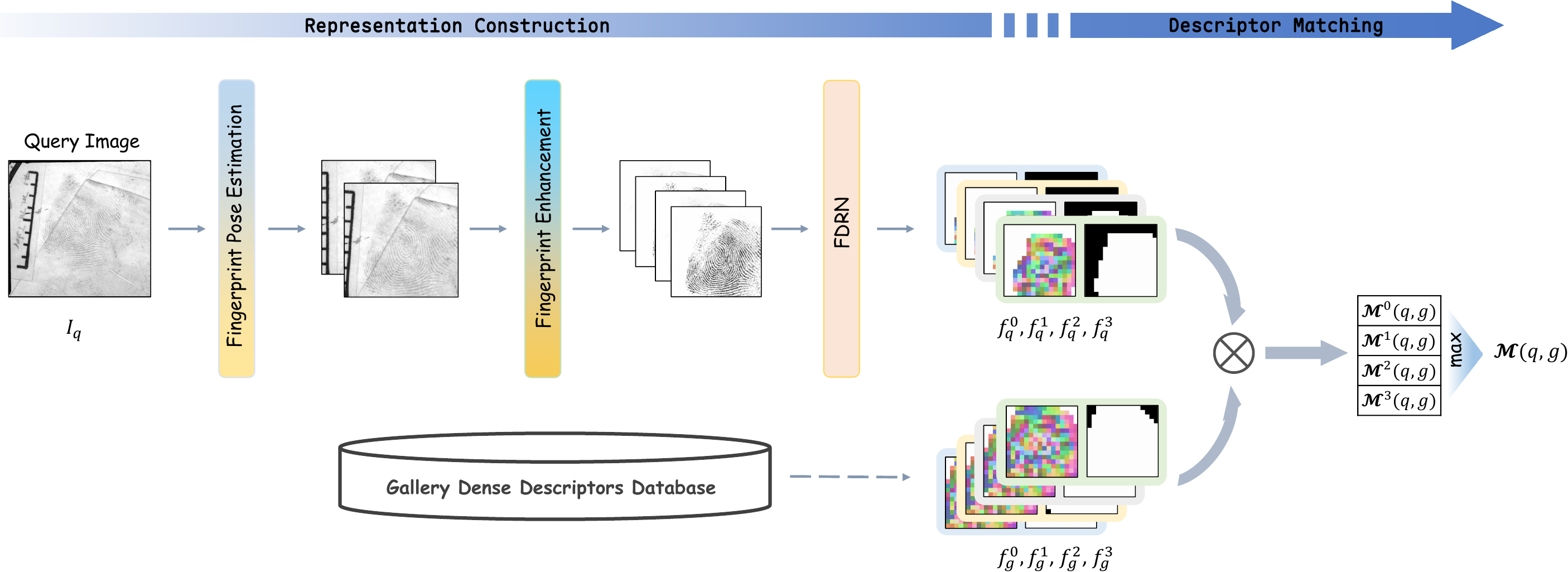}
  \caption{FLARE matching pipeline. Each image is processed through two pose estimators and two enhancers, yielding four descriptor pairs. The final score is the maximum of four cosine similarities.}
  \label{fig:FLARE_pipeline}
\end{figure*}

\section{Methodology}

\subsection{Overview of FLARE}
The overall fingerprint matching pipeline of FLARE is illustrated in Fig.~\ref{fig:FLARE_pipeline}. For each input fingerprint, FLARE first estimates the 2D pose using two complementary estimators and aligns the image into standardized spatial configurations according to each estimator's interpretation. The aligned images are then enhanced through two distinct strategies\footnote{The rationale for performing pose estimation prior to enhancement is discussed in Sec.~\ref{sec:discussions}.}, followed by descriptor extraction to  produce four sets of outputs, each consisting of a dense, fixed-length representation and a corresponding foreground mask. Matching between two fingerprints is performed by computing cosine similarity between descriptors within overlapping foreground regions. The four sets of descriptors are independently compared, yielding four similarity scores, and the maximum score is selected as the final matching result. The fingerprint pose estimation, enhancement, and dense descriptor extraction modules are trained independently. The following sections provide detailed descriptions of the pose estimation process, fingerprint enhancement, fixed-length dense descriptor extraction, and the matching procedure.

\subsection{Fingerprint 2D Pose Estimation}
For fingerprint pose estimation, FLARE adopts a lightweight global regression-based method proposed by Guan et al. \cite{guan2025robust} and a local voting-based strategy proposed by Duan et al. \cite{duan2023estimating}. The regression-based method predicts the fingerprint center and orientation directly from the overall fingerprint contour, while the voting-based method infers the pose by aggregating dense local predictions based on fine-grained ridge structures. By integrating both approaches, FLARE leverages complementary global and local information to achieve more reliable spatial alignment. Architectural and training details follow the original implementations.

To further enhance the robustness of pose estimation under background noise, we employ a data augmentation strategy that randomly inserts background patterns during training. This approach substantially improves performance on latent fingerprints with heavy background interference. In this part,  background augmentation is performed using images from the MSRA-TD500 dataset \cite{MSRA_TD500}.

\begin{figure}[t]
  \centering
  \includegraphics[width=.9\linewidth]{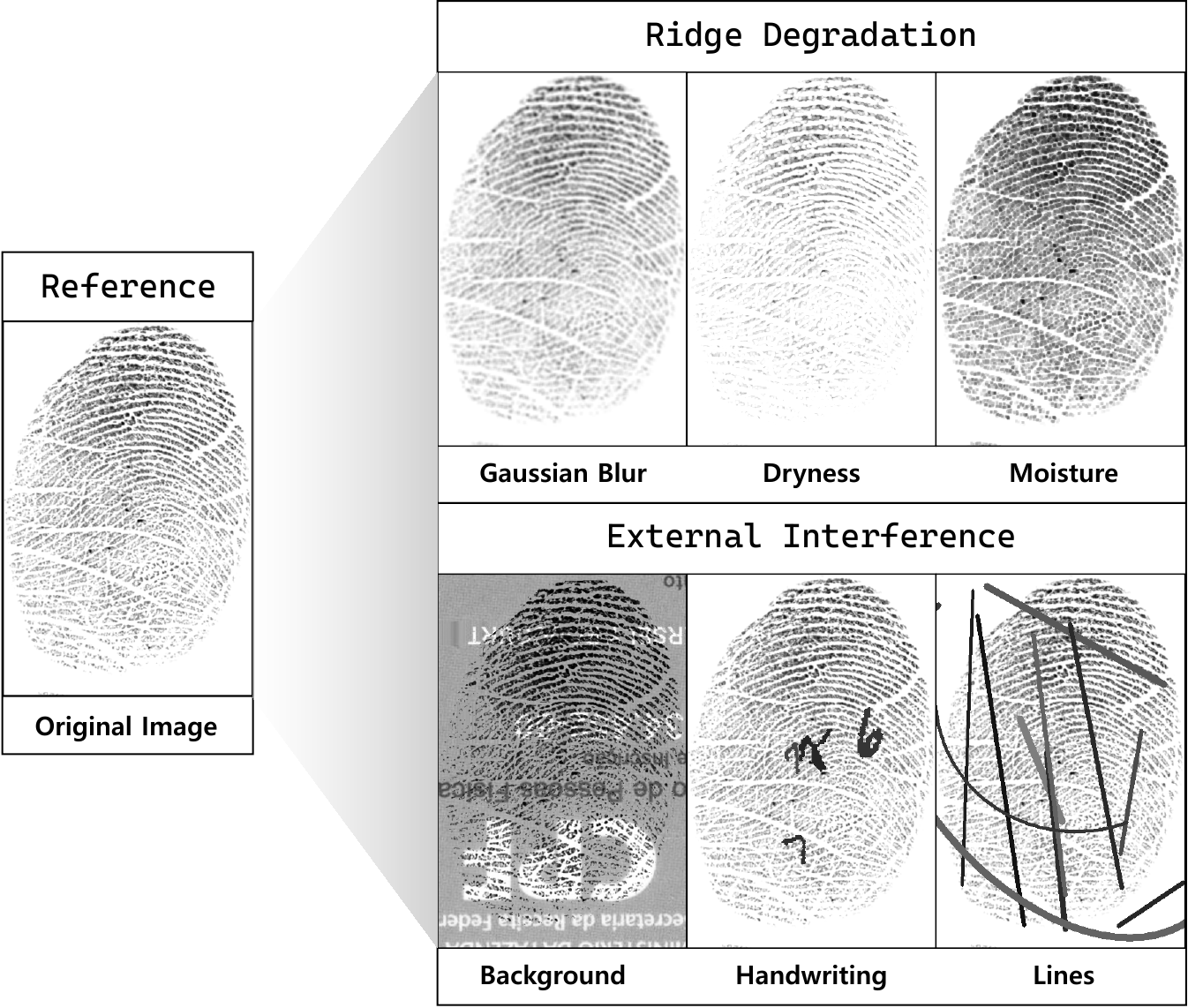}
  \caption{Examples of fingerprint degradation simulation.}
  \label{fig:noise_examples}
\end{figure}

\subsection{Fingerprint Enhancement} \label{sec:enh_method}
Our enhancement design aims to suppress background noise while enhancing the contrast of foreground ridge patterns. To train the enhancement network, we use optically captured high-quality fingerprints and simulate low-quality inputs through a set of handcrafted degradations. These include: (1) ridge degradation, such as Gaussian blur, dry or moist fingerprint artifacts that reduce ridge quality; and (2) external interference, such as background overlays from BID \cite{BID} and occlusions with line patterns or handwritten digits from MNIST \cite{deng2012mnist}. These degradations can be combined to increase diversity. Fig.~\ref{fig:noise_examples} shows examples of high-quality fingerprints and their corresponding degraded versions. This process produces paired training samples $(I_\text{LQ}, I_\text{HQ})$ for learning the enhancement network.

\begin{figure}
  \centering
  \includegraphics[width=\linewidth]{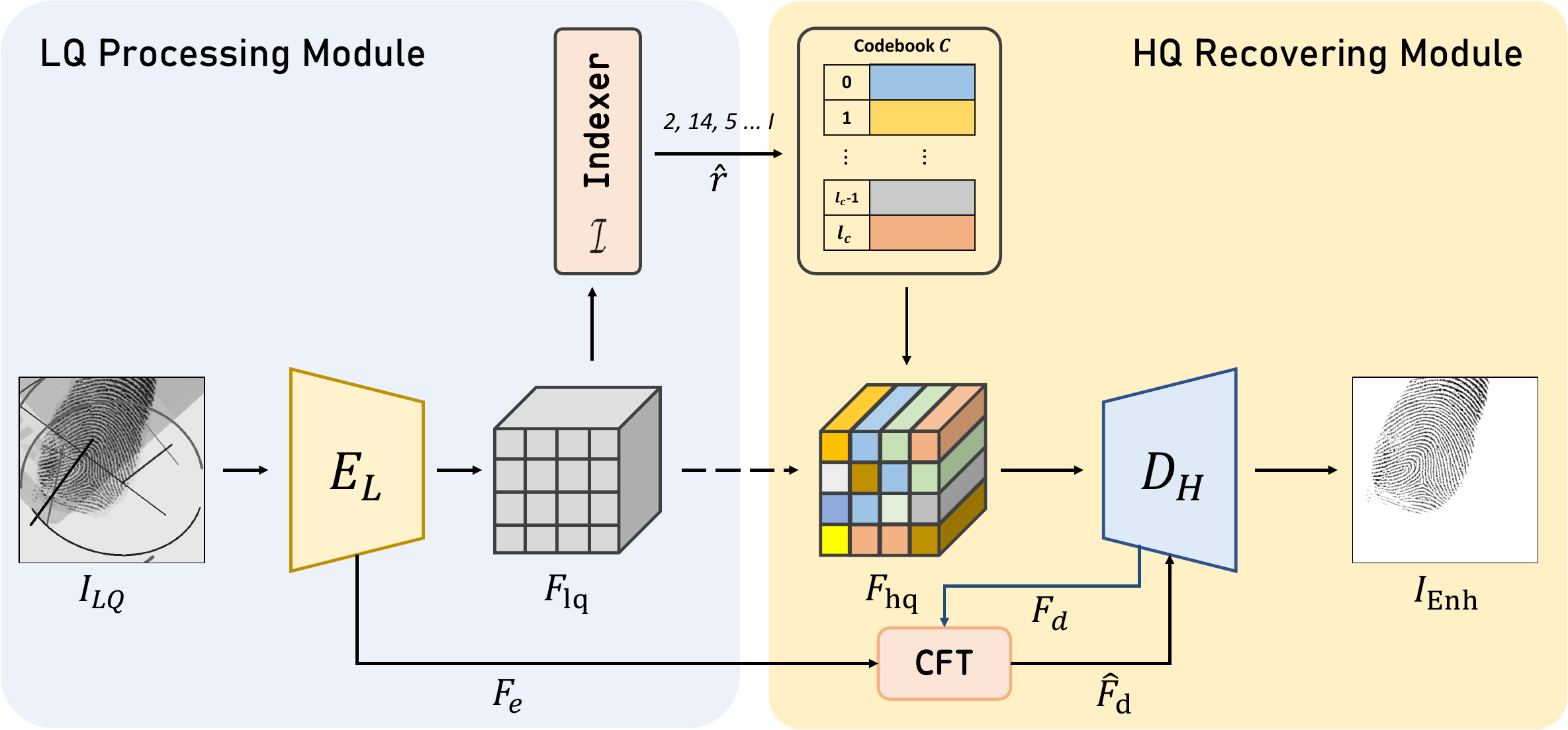}
  \caption{The architecture illustration of PriorEnh.}
  \label{fig:PriorEnh_structure}
\end{figure}

To enhance fingerprint quality while preserving structural fidelity, we propose two complementary modules: UNetEnh, a U-Net–based model that directly suppresses image noise, and PriorEnh, which incorporates ridge-structure priors to guide enhancement.UNetEnh adopts the U-Net architecture \cite{ronneberger2015u}, taking simulated noisy fingerprints $I_\text{LQ}$ as input and producing enhanced outputs $I_\text{Enh}$.PriorEnh is trained in two stages (Fig.~\ref{fig:PriorEnh_structure}). In the first stage, a VQ-VAE \cite{vqvae} is trained in a self-reconstruction manner using high-quality fingerprints $I_\text{HQ}$ to learn a latent codebook $\mathcal{C} \in \mathbb{R}^{l_c \times d_c}$ that captures fine-grained ridge priors. In the second stage, the pretrained codebook $\mathcal{C}$ and decoder $D_H$ are frozen. A U-Net–shaped encoder $E_L$, a GatedPixelCNN-based indexer $\mathcal{I}$ \cite{GatedPixelCNN}, and a controllable feature transformation (CFT) module \cite{wang2018recovering} are jointly trained with $I_\text{LQ}$ as input and $I_\text{HQ}$ as the target. The encoder and indexer reconstruct latent representations, while the CFT module modulates decoder features using spatial transformations from encoder activations, improving output fidelity.

To address the computational cost of operating on relatively large latent maps, we adopt GatedPixelCNN \cite{GatedPixelCNN} as the indexer $\mathcal{I}$ that produce the indexes to the codebook in a $l_c$ classification form as $\hat{r}$. The CFT module enables the encoder features $F_e$ to modulate the decoder features $F_d$ through spatial feature transformation: 
\begin{equation} \label{eq:cft}
  \hat{F}_d = F_d + (\alpha \odot F_d + \beta) \times w\;; \quad \alpha, \beta = \text{CFT}(F_d, F_e)\;, 
\end{equation}
where $\alpha$ and $\beta$ are the affine parameters, and $w \in [0,1]$ is a weighting factor that controls the degree of influence. Together with the pretrained codebook and decoder from the first stage, these components constitute the full PriorEnh network. 

The enhancement loss $\mathcal{L}_\text{Enh}$ for UNetEnh can be simply calculated by the Mean Square Error (MSE) between the predicted enhanced fingerprint image $I_\text{Enh}$ and $I_\text{HQ}$. In the case of PriorEnh, the first-stage training loss $\mathcal{L}_\text{Enh}^{s1}$ is consistent with that of the standard VQ-VAE \cite{vqvae}, and thus is not elaborated here. For the second training stage, we supervise the reconstruction of the latent feature map using the following loss:
\begin{equation} \label{eq:latent_rec}
   \mathcal{L}_\text{ind} = \left\| F_\text{lq} - \mathrm{sg}(F_\text{c}) \right\|_2^2 + \lambda_\text{ind} \sum_{i=0}^{hw-1} -r_i^\ast \log(\hat{r}_i)\;,
 \end{equation}
where $F_\text{lq} \in \mathbb{R}^{h\times w \times d_c}$ denotes the latent feature map produced by the low-quality encoder $E_L$, and $F_\text{c}$ is the corresponding latent feature map generated by the first-stage encoder $E_H$ with high-quality input $I_\text{HQ}$. The operator $\mathrm{sg}$ indicates stop-gradient. $\hat{r}_i$ denotes the predicted index probability of the $i$-th spatial location, and $r_i^\ast$ is the corresponding ground truth generated from $F_\text{c}$. We set $\lambda_\text{ind} = 0.5$ in our case. The total loss for the second training stage is defined as:
\begin{equation} \label{eq:priorenh}
  \mathcal{L}_\text{Enh}^{s2} = \left\|I_\text{Enh} - I_\text{HQ}\right\|_1 + \left\|\mathcal{P}(I_\text{Enh}) - \mathcal{P}(I_\text{HQ})\right\|_2^2 + \mathcal{L}_\text{ind}, 
\end{equation} 
where $\mathcal{P}(\cdot)$ denotes the perceptual similarity measured by the LPIPS metric \cite{LPIPS}. 

\begin{figure}
  \centering
  \includegraphics[width=\linewidth]{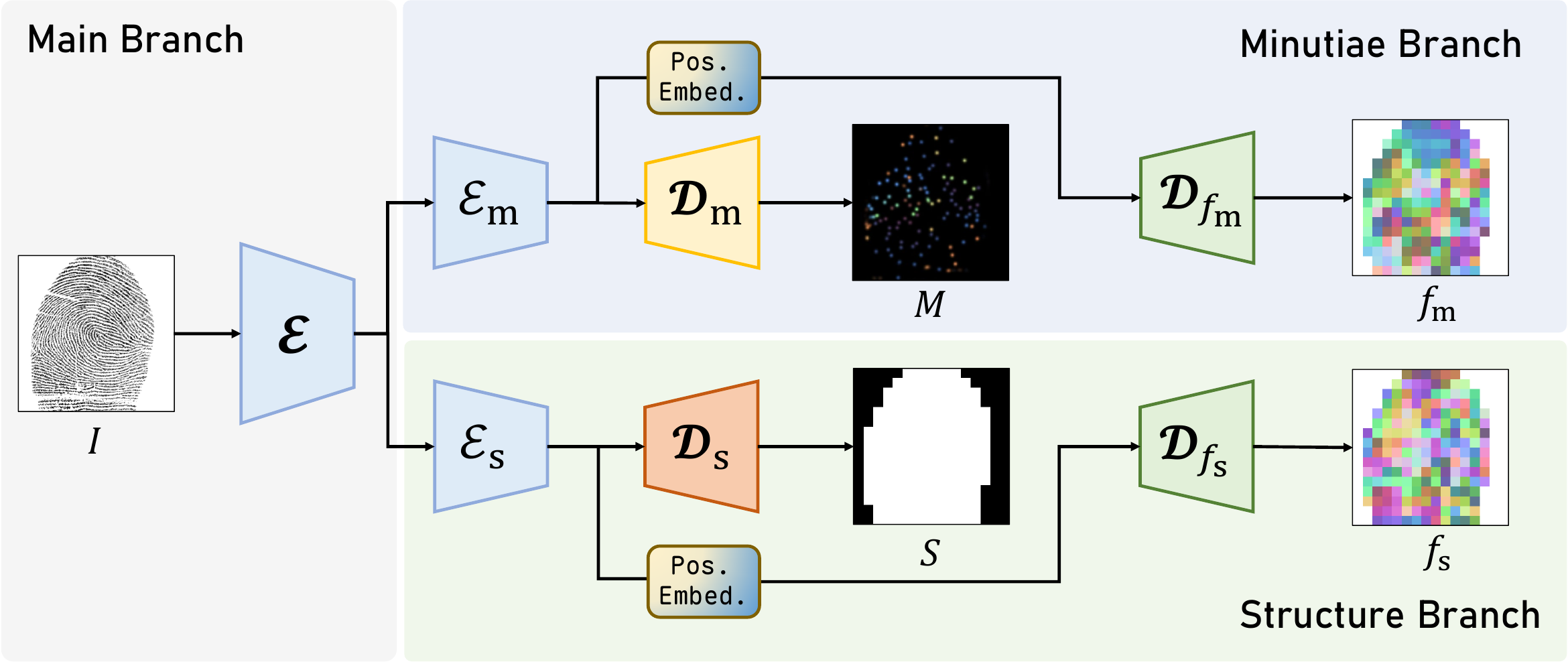}
  \caption{The architecture illustration of FDRN.}
  \label{fig:FDRN_structure}
\end{figure}

\subsection{Fixed-length Dense Representation Extraction}
To handle various fingerprint foreground area and mitigate background interference, we adopt a three-dimensional dense representation, with two axes corresponding to spatial coordinates. Unlike previous approaches that compress the entire fingerprint into a global abstract one-dimensional vector, this representation retains local structural details and facilitates foreground-background separation via an associated validity mask that identifies meaningful fingerprint regions. Based on this design, we propose the Fixed-length Dense Representation Network (FDRN) to extract a fixed-length dense descriptor that captures both minutiae-related and structure-related features, thereby enhancing discriminative capability.

The FDRN is built upon ResNet-34 \cite{resnet}, with the first max-pooling layer removed to preserve fine-grained details. Furthermore, we modify it into a dual-branch structure to capture complementary aspects of the fingerprint representation. Since the input fingerprint images have been aligned to a canonical coordinate system through pose estimation, the dense descriptor space is also spatially aligned. To further enhance spatial awareness, we incorporate a classical 2D sinusoidal positional embedding module \cite{vaswani2017attention} into each dense representation decoder, enabling the network to extract more distinct and location-sensitive features. The network architecture is illustrated in Fig. \ref{fig:FDRN_structure}.

Specifically, we formulate the two branches as multi-task learning modules. Given an input fingerprint image $I \in \mathbb{R}^{1\times H \times W}$, the first branch (Minutiae branch) outputs a minutiae map $M \in \mathbb{R}^{6 \times \frac{H}{2} \times \frac{W}{2}}$ \cite{DeepPrint}, along with a fixed-length dense representation $f_\text{m} \in \mathbb{R}^{\mathcal{D} \times \frac{H}{16} \times \frac{W}{16}}$ that is related to the minutiae. The second branch (Structure branch) generates a foreground segmentation map $S \in \mathbb{R}^{1 \times \frac{H}{16} \times \frac{W}{16}}$, which corresponds to the spatial size of the dense descriptor, and produces a structure-aware representation $f_\text{s} \in \mathbb{R}^{\mathcal{D} \times \frac{H}{16} \times \frac{W}{16}}$ that focus on the foreground ridge patterns. The final fixed-length dense descriptor $f\in \mathbb{R}^{2\mathcal{D} \times \frac{H}{16} \times \frac{W}{16}}$ is obtained by concatenating the dense representations from both branches and applying the foreground mask as follows:
\begin{equation}\label{eq:dense_desc}
  f = (f_\text{s} \oplus f_\text{m}) \odot S\;,
\end{equation}
where $\oplus$ stands for the concatenation and $\odot$ represents the Hadamard product. 

The loss functions for the representation extraction network consist of two main components: one for learning the dense representations and the other for the auxiliary tasks. For the representation learning component, we adopt the CosFace loss \cite{wang2018cosface} as a classification loss $\mathcal{L}_\text{cls}$ to encourage each dense representations to acquire discriminative identity-related features. The two branches, $f_\text{m}$ and $f_\text{s}$, is supervised independently, leading to a total classification loss of $\mathcal{L}_\text{cls} = \mathcal{L}_\text{cls}^\text{m} + \mathcal{L}_\text{cls}^\text{s}$. To further enhance the robustness of dense descriptors against variations in foreground coverage, intensity distribution, and local distortion, we introduce a local consistency loss $\mathcal{L}_{\text{lc}}$ that enforces feature similarity across overlapping foreground regions of different impressions from the same identity. Specifically, we simulate an incomplete fingerprint $I_\text{inc}$ by applying a binary mask derived from a real plain fingerprint onto a rolled fingerprint $I_\text{rol}$, followed by elastic deformation and intensity shifting to emulate natural variations. Specifically, the loss function is defined by:
\begin{equation}\label{eq:sim}
  \mathcal{L}_\text{lc} = \frac{1}{|s_{\text{inc}\cap \text{rol}}|} \sum_{(i,j)\in S_{\text{inc}\cap \text{rol}}} \left\| f^\text{inc}(i,j) - f^\text{rol}(i,j) \right\|_2^2\;,
\end{equation}
where $f^\text{inc}$ and $f^\text{rol}$ denotes the dense descriptors extracted from $I_\text{inc}$ and $I_\text{rol}$ respectively. $s_{\text{inc}\cap \text{rol}}$ represents the overlapping foreground regions. 

For the auxiliary tasks, we supervise the network to predict both the minutiae map and the foreground mask. The auxiliary loss consists of a binary cross-entropy loss $\mathcal{L}_\text{mask}$ for the mask prediction and a MSE loss $\mathcal{L}_\text{mnt}$ for the minutiae map regression. The ground-truth minutiae maps and foreground masks are extracted using VeriFinger v12.0 \cite{nist2020verifinger}. The overall loss for the Fixed-length Dense Representation Network (FDRN) is defined as:
\begin{equation} \label{eq:fdrn}
  \mathcal{L}_\text{FDRN} = \mathcal{L}_\text{cls} + \lambda_\text{ls}\mathcal{L}_\text{ls} + \lambda_\text{mask}\mathcal{L}_\text{mask} + \lambda_\text{mnt}\mathcal{L}_\text{mnt}\;,    
\end{equation}
where we set $\lambda_\text{mask}=1$, $\lambda_\text{mnt}=0.01$, and $\lambda_\text{ls}=0.00125$ to achieve roughly the same decreasing rate for these loss functions. 

\subsection{Fingerprint Matching}
In practical deployments, dense representations for all gallery fingerprints are extracted offline through the complete FLARE pipeline, including pose estimation, enhancement, and descriptor extraction. These fixed-length representations are then stored in a database for efficient matching. 

During inference, a query fingerprint $I_q$ undergoes the same processing steps. Given a 500 ppi query image $I_q$ and a gallery image $I_g$, we first apply pose estimation and alignment, followed by cropping to $512 \times 512$ pixels. The aligned images are then enhanced using two pose estimation methods and two enhancement strategies, producing four augmented versions of each image: $\{(I_q^i, I_g^i)\;|\;i=0,1,2,3\}$. To balance computational efficiency and matching accuracy, each enhanced image is downsampled to $256 \times 256$ pixels before being input to the Fixed-length Dense Representation Network (FDRN), which produces dense descriptors $f_q^i, f_g^i \in \mathbb{R}^{2\mathcal{D} \times 16 \times 16}$ and foreground masks $S_q^i, S_g^i \in \mathbb{R}^{1 \times 16 \times 16}$ (see Eq.~\ref{eq:dense_desc}). The matching score is computed by measuring the cosine similarity between the flattened dense representations ${f_q^i}^\prime, {f_g^i}^\prime \in \mathbb{R}^{512\mathcal{D}}$, where only the overlapping foreground regions are considered. Specifically, the score for the $i$-th fingerprint pair $(I_q^i, I_g^i)$ is defined as:
\begin{equation}\label{eq:matching_score}
  \mathcal{M}^i(q,g) = \frac{{{f_q^i}^\prime}^T \cdot {f_g^i}^\prime}{\left\|f_q^i \odot S_g^i \right\|_F \left\|f_g^i \odot S_q^i \right\|_F}\;,
\end{equation}
where $\left\|\cdot\right\|_F$ stands for Frobenius norm. And the final matching score for fingerprint images $(I_q, I_p)$ is obtained by taking the maximum over the four matching scores derived from different pose-enhancement combinations:
\begin{equation}\label{eq:fuse_score}
  \mathcal{M}(q,g) = \max_{i\in\{0,1,2,3\}}\mathcal{M}^i(q,g)\;.
\end{equation}

\begin{table}[t]
  \caption{Training configurations of the blocks in FLARE. BS: batch size, LR: learning rate.}
  \label{tab:train_details}
  \centering
  \begin{tabular}{lcc
    *{1}{>{\centering\arraybackslash}p{0.1\linewidth}}
    c}
    \toprule
     \textbf{Block} & \textbf{\makecell{BS}} & \textbf{\makecell{LR}} & \textbf{\makecell{Epochs}} & \textbf{GPU} \\
    \midrule
    Duan et al.\cite{duan2023estimating} & 64 & $3.5\times10^{-4}$ & 80 & 1 $\times$ RTX 3090 \\ 
    Guan et al.\cite{guan2025robust} & 64 & $1\times10^{-3}$ & 80 & 1 $\times$ RTX 3090 \\
    \hhline
    UNetEnh & 32 & $1\times10^{-4}$ & 50 & 1 $\times$ RTX 3090 \\
    PriorEnh s1 & 12 & $7\times10^{-5}$ & 40 & 6 $\times$ RTX 4090 \\ 
    PriorEnh s2 & 4 & $1\times10^{-5}$ & 50 & 2 $\times$ RTX 3090 \\
    \hhline
    FDRN & 24 & $1\times10^{-4}$ & 200 & 1 $\times$ RTX 3090 \\
    \bottomrule 
  \end{tabular}
\end{table}

\subsection{Implementation Details} \label{sec:implementation}
To improve generalization and prevent overfitting, we adopt a data augmentation strategy that simulates multiple impressions of the same finger. Specifically, we use the fingerprint distortion model proposed by Si et al. \cite{si2015detection} to generate random distortion fields, which are applied to real fingerprints to synthesize impressions with varied geometric deformations. In addition, histogram matching adjusts the grayscale distribution of synthetic images to simulate a broader range of visual conditions. We also incorporate mild geometric transformations. For pose estimation and enhancement modules, random geometric transformations are applied to the fingerprints of size $512 \times 512$ at 500 ppi, including
random translations in the range $[-80, 80]$ pixels along both axes and rotations within $[-45^\circ, 45^\circ]$. For FDRN training, we adopt smaller transformations with translations in $[-10, 10]$ pixels and rotations in $[-5^\circ, 5^\circ]$. Furthermore, for training the enhancement modules, we adopt the simulated low-quality fingerprint strategy described in Sec.~\ref{sec:enh_method}, where realistic degradations are applied to high-quality prints to generate paired training data. The codebook size $l_c$ is set to 4096, with an embedding dimension $d_c$ of 3. The feature dimension $\mathcal{D}$ for the fixed-length dense representation is set to 6. Additional training details for each FLARE module are summarized in Tab.~\ref{tab:train_details}, 
and the supplementary material further provides full network hyperparameter settings along with additional data augmentation details for training the pose estimation and enhancement modules.

Training uses the AdamW optimizer \cite{adamw} with a plateau-based learning rate scheduler that reduces the learning rate by a factor of 0.1 with a patience of 10 epochs. To monitor overfitting and select hyperparameters, 20\% of the training samples are held out as a validation set. After choosing the best-performing configuration, the model is retrained on the full training set and trained to convergence, defined as the point where the scheduled learning rate decays below $1 \times 10^{-6}$. The final converged model is used for evaluation.

\begin{table*}[!t]
  \begin{threeparttable}
    \centering
    \caption{Fingerprint datasets used in our work.}
    \label{tab:datasets}
    \begin{tabular}{llllccc}
      \toprule
      \textbf{Type} & \textbf{Dataset} & \textbf{Sensor} & \textbf{Description} & \textbf{Usage} & \textbf{Genuine Pairs} & \textbf{Imposter Pairs} \\
      \midrule
      \multirow{2}{*}{Rolled} & NIST SD14 & Inking & 24,000 pairs of rolled fingerprints & train\tnote{a} & - & -\\
      & NIST SD4 & Inking & 2,000 pairs of rolled fingerprints & test & 2,000 & 3,998,000 \\
      \hhline
      \multirow{3}{*}{Plain} & FVC2004 DB1A\tnote{b}  & Optical & 100 fingers $\times$ 8 impressions & test & 2,800 & 4,950 \\
      & N2N Plain & Optical & 2,000 pairs (plain and rolled fingerprints) & test & 2,000 & 3,998,000 \\
      & DPF & Optical & 776 rolled and 40,112 plain fingerprints with diverse poses & train\tnote{c} & - & - \\
      \hhline
      \multirow{2}{*}{Partial} & FVC2002 DB3A\tnote{b} & Capacitive & 100 fingers $\times$ 8 impressions & test & 2,800 & 4,950 \\
      & FVC2006 DB1A\tnote{d}  & Electric field  & 140 fingers $\times$ 12 impressions & test & 9,240 & 9,730 \\
      \hhline
      \multirow{2}{*}{Latent} & NIST SD27\tnote{e} & - & 258 pairs (latent fingerprints from crime scene) & test & 258 & 2,764,470 \\
      & THU Latent10K & - & 10,458 pairs (latent fingerprints from crime scene) & test & 10,458 & 109,359,306 \\
      \hhline
      Contactless & PolyU CL2CB\tnote{f} & Optical/Camera & 336 fingers $\times$ 6 contact-based and contactless fingerprints & test & 12,096 & 4,052,160 \\
      \bottomrule
    \end{tabular}
    \begin{tablenotes}
      \item[a] It is used for training the FDRN, and the first 3,200 gallery fingerprints are also used to train the pose estimation module.
      \item[b] The number of genuine pairs is $100 \times \binom{8}{2} = 2,800$, and the number of imposter pairs is $\binom{100}{2} = 4,950$.
      \item[c] It is used to train both the pose estimation and fingerprint enhancement modules.
      \item[d] The number of genuine pairs is $140 \times \binom{12}{2} = 9,240$, and the number of imposter pairs is $\binom{140}{2} = 9,730$.
      \item[e] The gallery portion (10,458 plain or rolled fingerprints) of THU Latent10K is appended to the original gallery set.
      \item[f] The number of genuine pairs is $336 \times 6 \times 6 = 12,096$, and the number of imposter pairs is $336 \times 335 \times 6 \times 6 = 4,052,160$.
    \end{tablenotes}
  \end{threeparttable}
\end{table*}

\begin{figure*}[!t]
  \centering
  \subfloat[]{\includegraphics[height=.09\linewidth]{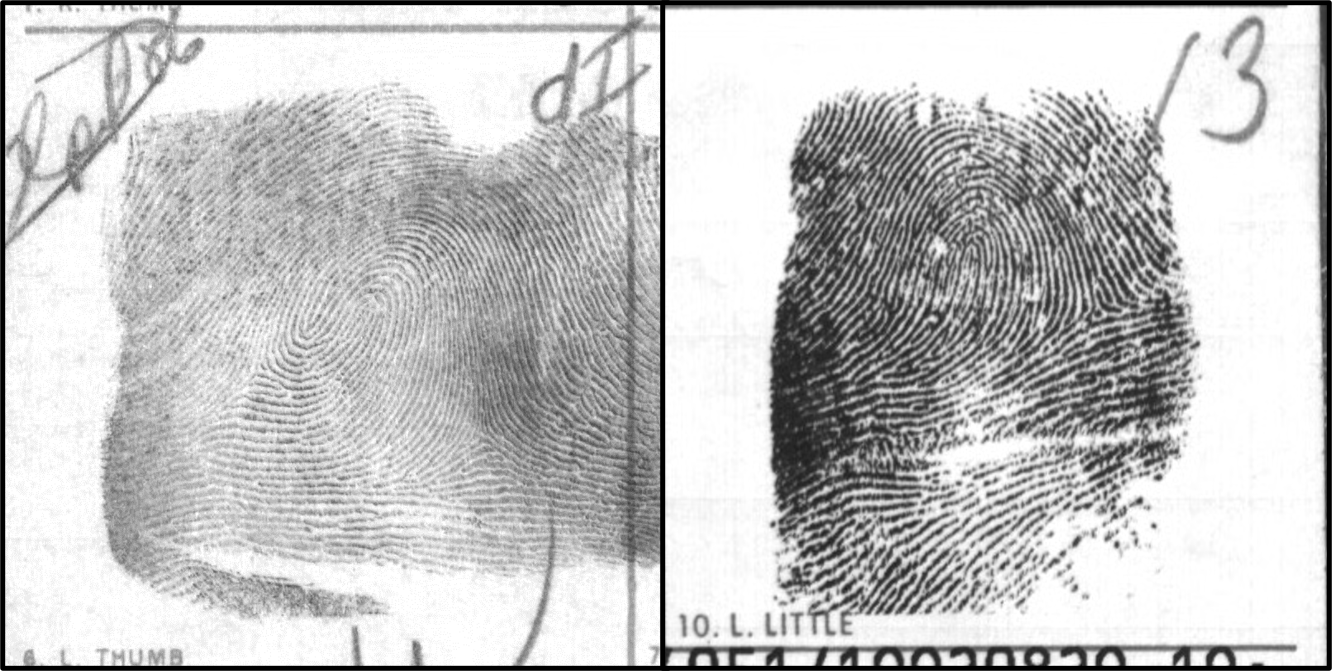}} \hfil
  \subfloat[]{\includegraphics[height=.09\linewidth]{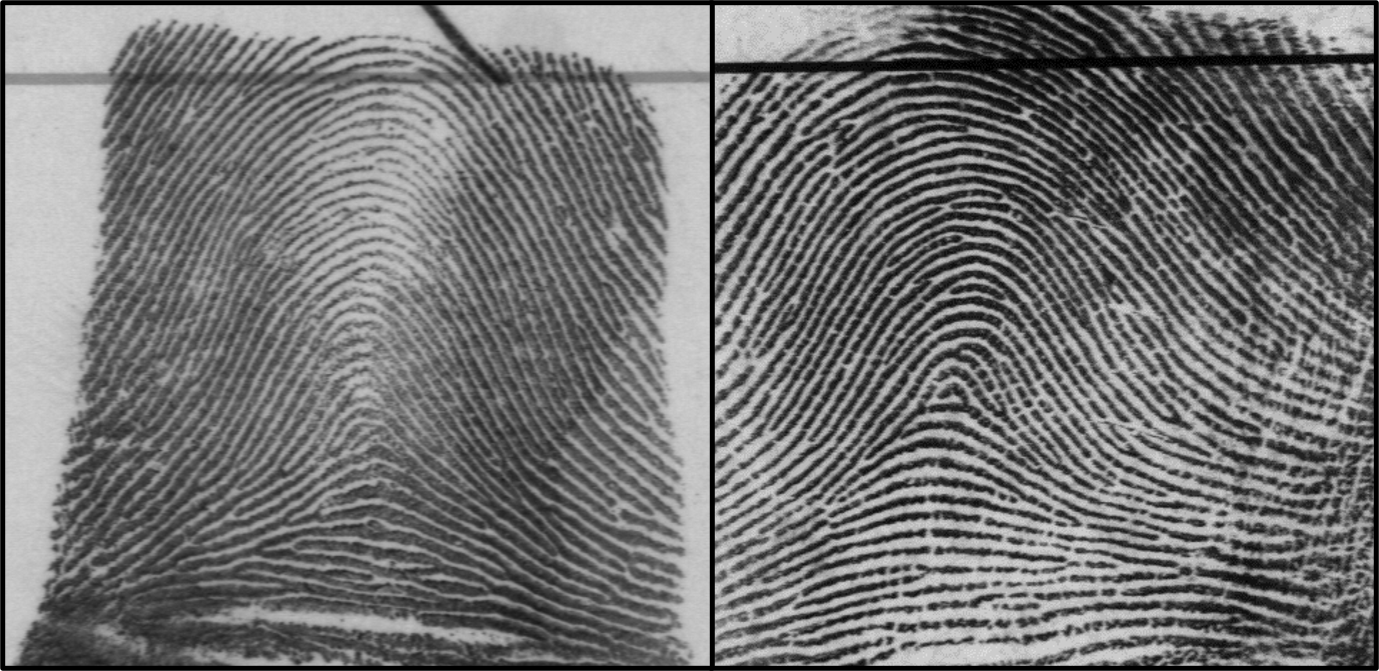}} \hfil
  \subfloat[]{\includegraphics[height=.09\linewidth]{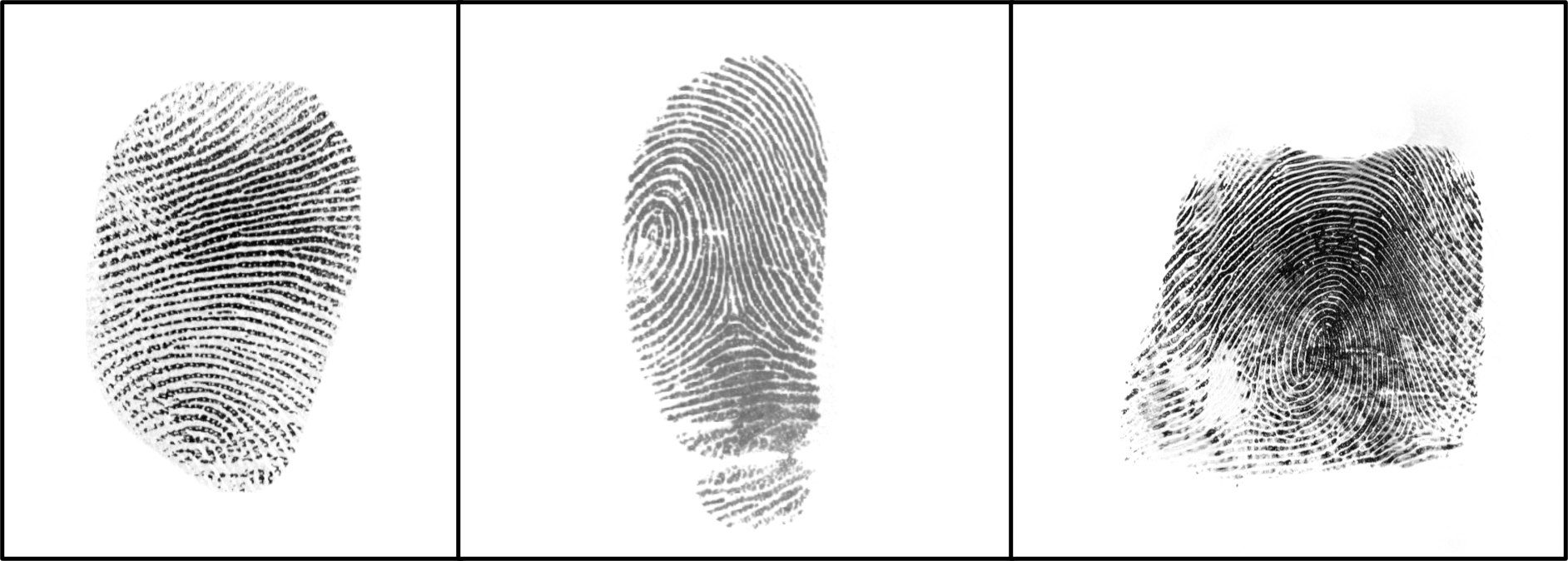}} \hfil
  \subfloat[\label{fig:n2nplain_ex}]{\includegraphics[height=.09\linewidth]{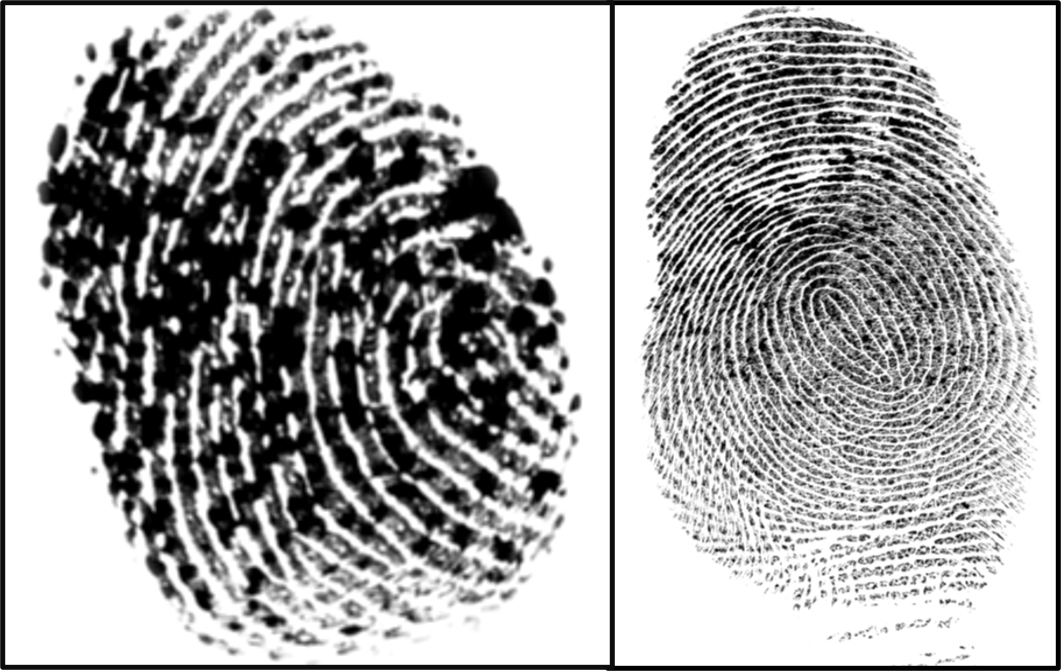}} \hfil
  \subfloat[]{\includegraphics[height=.09\linewidth]{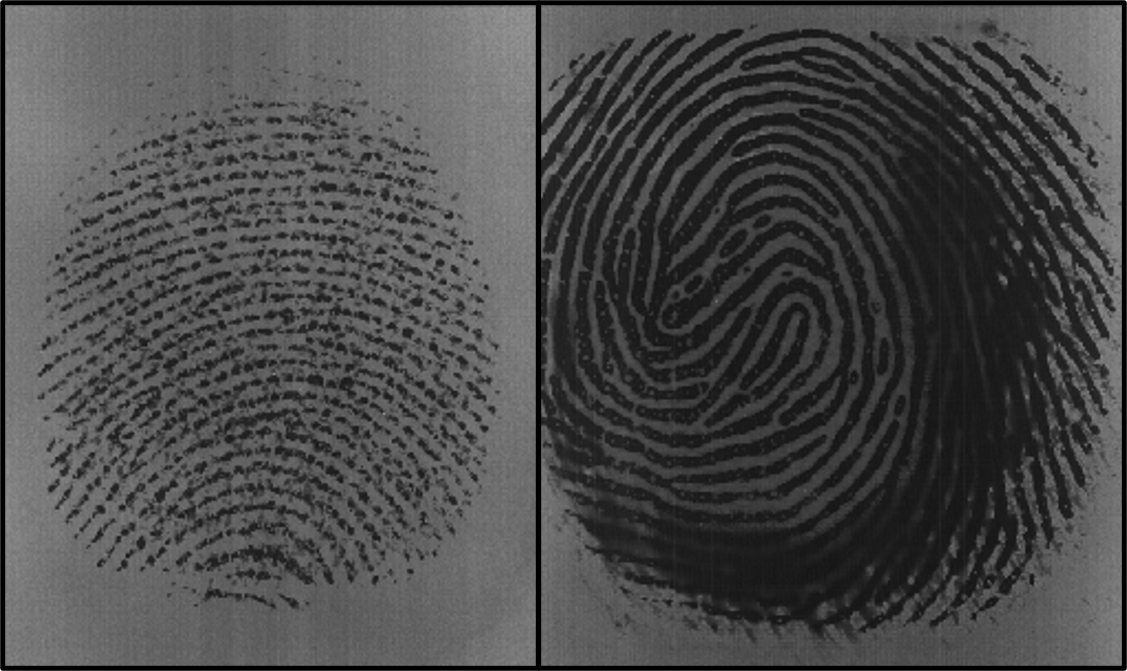}} \hfil
  \subfloat[]{\includegraphics[height=.09\linewidth]{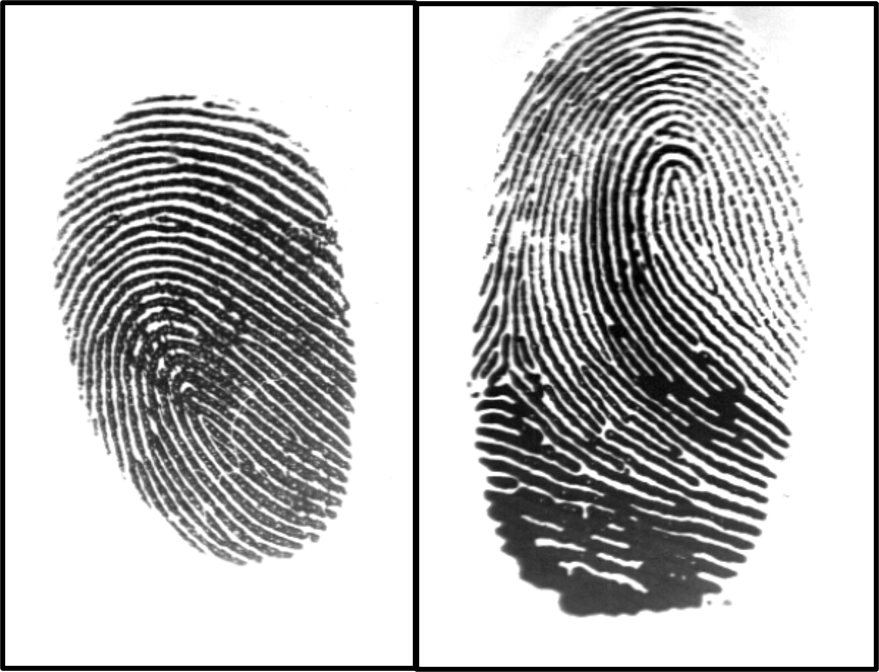}} \hfil
  \subfloat[]{\includegraphics[height=.09\linewidth]{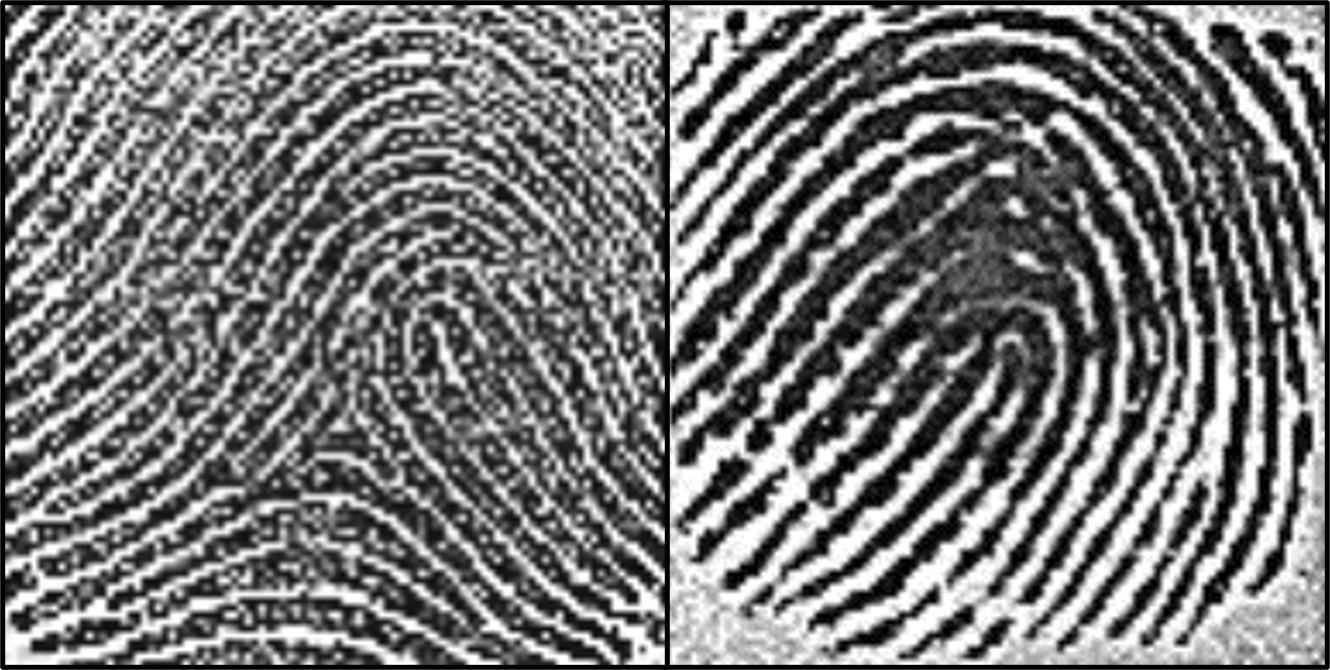}} \hfil
  \subfloat[]{\includegraphics[height=.09\linewidth]{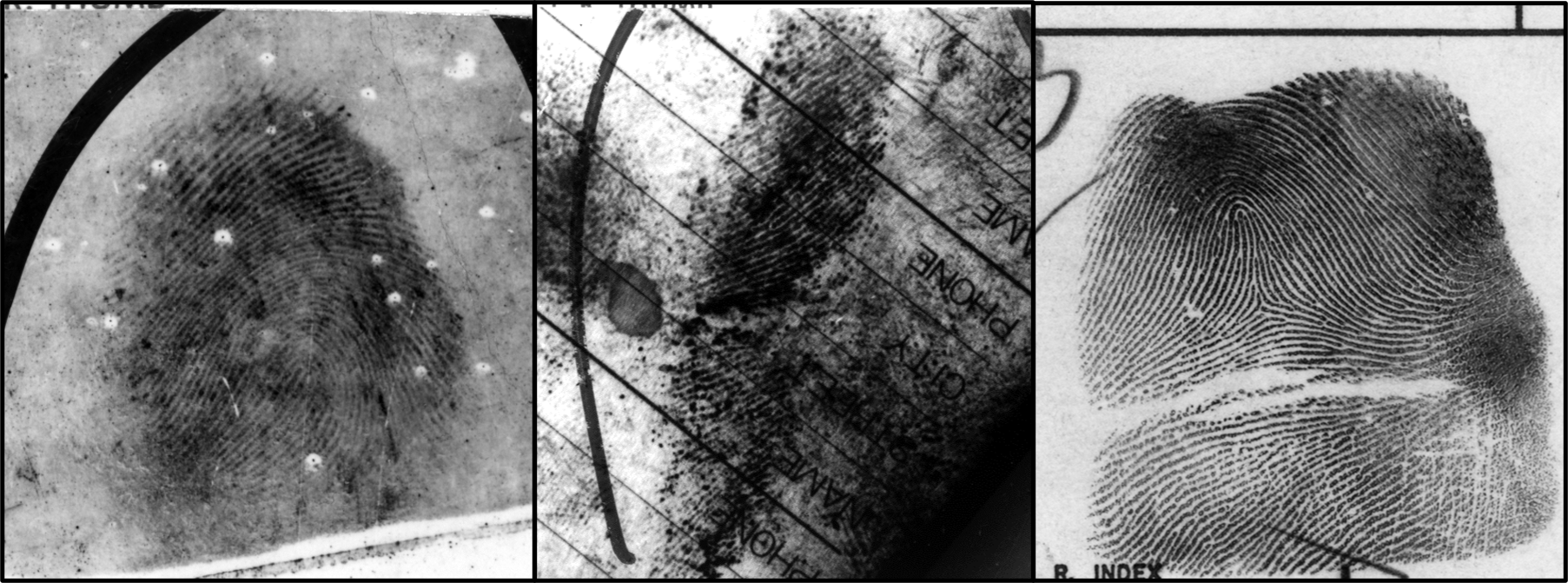}} \hfil
  \subfloat[]{\includegraphics[height=.09\linewidth]{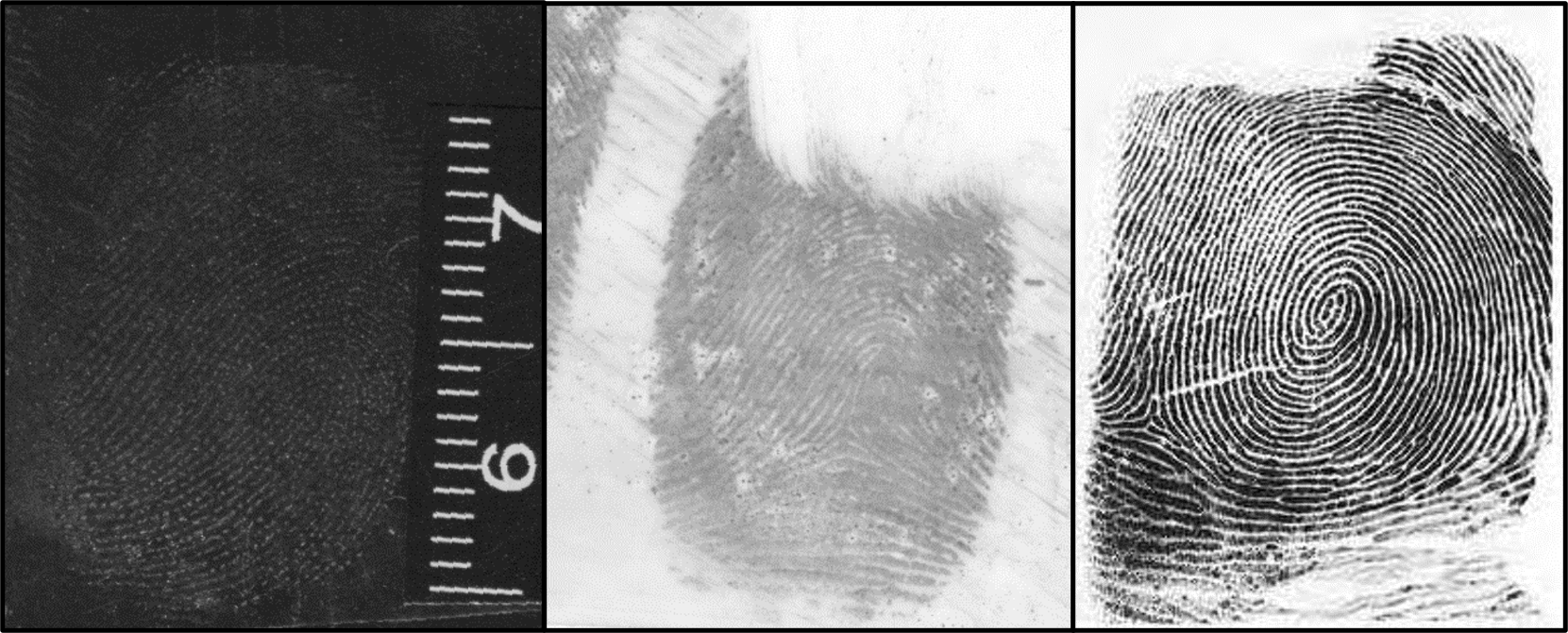}} \hfil
  \subfloat[]{\includegraphics[height=.09\linewidth]{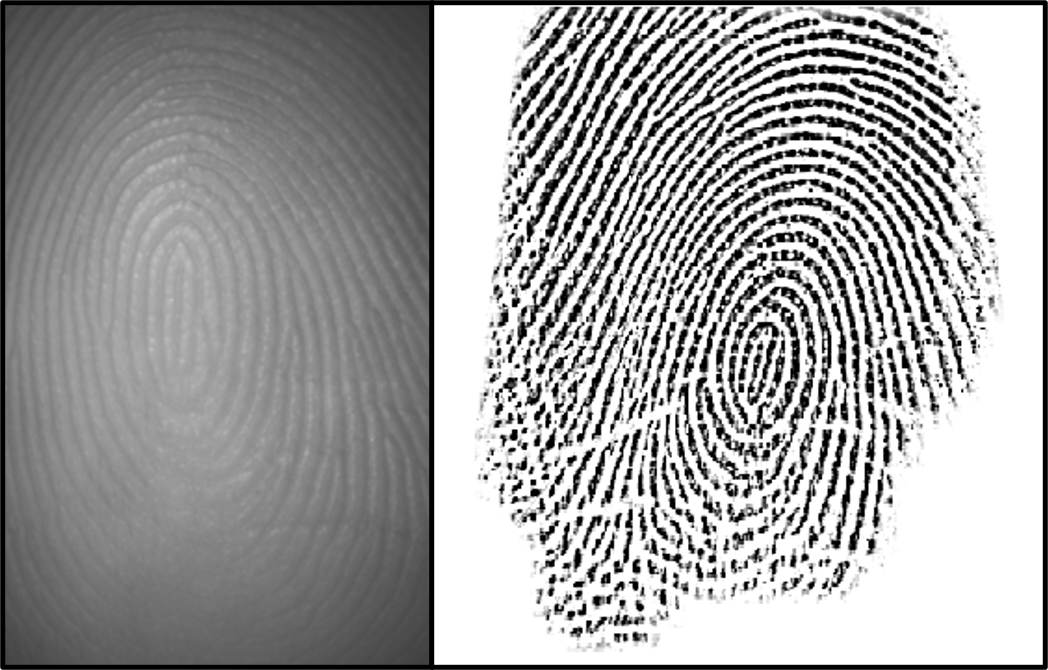}} \hfil
  \caption{Fingerprint examples from different fingerprint datasets (a) NIST SD14, (b) NIST SD4, (c) DPF, (d) N2N Plain, (e) FVC2002 DB3A, (f) FVC2004 DB1A, (g) FVC2006 DB1A, (h) NIST SD27, (i) THU Latent10K, (j) PolyU CL2CB. The input images are at 500 ppi, but have been rescaled in this figure to facilitate clearer visualization.}
  \label{fig:datasets}
\end{figure*}

\section{Experiment}
In this section, we first describe the datasets used in our experiments and then evaluate FLARE`s matching performance against representative fixed-length representation methods. We next present ablation studies and provide a detailed analysis of the proposed fixed-length dense representation. We further examine the influence of pose estimation and enhancement and analyze how they work together to improve performance. Finally, we assess the space and computational complexity of the dense representation and the overall FLARE system.
\subsection{Datasets}
We employ a variety of fingerprint datasets in this work, covering multiple modalities including rolled, plain, partial, latent, and contactless fingerprints, collected using diverse sensors and under varying acquisition conditions. All fingerprint images not originally in 500 ppi resolution are rescaled to 500 ppi. Tab.~\ref{tab:datasets} summarizes the details of all the datasets used in our experiments, and representative fingerprint samples from each are illustrated in Fig. \ref{fig:datasets}.

For training, we use only high-quality rolled and plain fingerprints to emphasize generalization capability. Specifically, we adopt the Diverse Pose Fingerprint (DPF) dataset \cite{duan2023estimating}, which contains 776 rolled fingerprints and 40,112 plain fingerprints with diverse pose variations, along with the first 3,200 rolled fingerprints from the gallery portion of NIST SD14, to train the pose estimation module. The same DPF dataset is also used to train the fingerprint enhancement modules, and the codebook for PriorEnh is constructed based on it as well. For training the fixed-length dense representation, we follow our previous conference work \cite{FDD} and use the first 24,000 pairs of rolled fingerprints from NIST SD14. Additionally, 32,676 plain fingerprints from 633 fingers in DPF are randomly selected and combined with their segmentation masks to simulate incomplete fingerprints as described in Eq.~\ref{eq:sim}. It is important to note that all training is performed exclusively on high-quality contact-based rolled or plain fingerprints, without exposure to other modalities. Evaluation is conducted directly on fingerprints of various types without fine-tuning.

To evaluate the generalization and robustness of our method, we conduct experiments on several benchmark and diverse fingerprint datasets covering a wide range of modalities. For the FVC benchmark, we select three representative datasets: FVC2002 DB3A, FVC2004 DB1A, and FVC2006 DB1A. For these datasets, we followed the same experimental settings for splitting the imposter and genuine pairs as used in previous works \cite{cappelli2010minutia, su2016fingerprint}. The NIST SD302 dataset \cite{nist302} contains 2,000 fingers from 200 subjects. We use subset U (2,000 rolled fingerprints) as the gallery, and combine subsets R and S (2,000 plain fingerprints) as the query, forming the N2N Plain. For evaluation on the NIST SD27 latent fingerprint dataset, we expand the gallery by including 10,458 rolled fingerprints from the gallery portion of our internal dataset, THU Latent10K. For the contactless-to-contact setting, we follow the preprocessing protocol of Cui et al. \cite{cui2023monocular} on the PolyU CL2CB dataset \cite{liu20143d}, where contactless fingerprints are scaled to match the mean ridge period of contact-based fingerprints without applying any geometric warping. 

\begin{table}[!t]
  \centering
  \caption{Description of the compared fixed-length representation methods.}
  \label{tab:method_desc}
  \renewcommand{\arraystretch}{1.2}
  \begin{threeparttable}
    \resizebox{\linewidth}{!}{
    \begin{tabular}{lll}
      \toprule
      \textbf{Approach} & \textbf{Preprocessing} & \textbf{Representation} \\
      \midrule
      DeepPrint \cite{DeepPrint} & STN alignment & Minutiae/Texture (1D)\\
      AFRNet \cite{grosz2024afrnet} & STN alignment & CNN/Transformer (1D) \\
      MultiScale \cite{gu2022latent} & Pose alignment & Multi-region (1D) \\
      FDD \cite{FDD} & Pose alignment & Minutiae/Structure (dense)\\
      FLARE & Dual Pose/Enh.\tnote{$\ddagger$} & Minutiae/Structure (dense)\\
      \bottomrule
    \end{tabular}}
    \begin{tablenotes}
      \item[$\ddagger$] It stands for dual fingerprint pose alignment and dual fingerprint enhancement.
    \end{tablenotes}
\end{threeparttable}
\end{table}

\begin{table*}
  \centering
  \caption{Matching accuracy (\%) on several fingerprint datasets. Unless otherwise specified, TAR@FAR = 0.1\% is reported. Bold indicates the best performance, and italic denotes the second-best.}
  \label{tab:compared_method}
  \renewcommand{\arraystretch}{1.2}
  \begin{threeparttable}
    \resizebox{\textwidth}{!}{
      \begin{tabular}{l
        c*{2}{>{\centering\arraybackslash}p{0.05\linewidth}}
        c*{1}{>{\centering\arraybackslash}p{0.05\linewidth}}
        c*{1}{>{\centering\arraybackslash}p{0.05\linewidth}}
        c*{4}{>{\centering\arraybackslash}p{0.05\linewidth}}
        c*{4}{>{\centering\arraybackslash}p{0.05\linewidth}}
        }
        \toprule
        \multirow{2}{*}{\textbf{Method}} & \multicolumn{2}{c}{\textbf{NIST SD4}} & \multicolumn{2}{c}{\textbf{N2N Plain}} & \multicolumn{1}{c}{\textbf{FVC02}\tnote{$\ddagger$}} & \multicolumn{1}{c}{\textbf{FVC04}\tnote{$\ddagger$}} & \multicolumn{1}{c}{\textbf{FVC06}\tnote{$\ddagger$}} & \multicolumn{1}{c}{\textbf{PolyU}\tnote{$\ddagger$}} & \multicolumn{2}{c}{\textbf{NIST SD27}} & \multicolumn{2}{c}{\textbf{THU Latent10K}} \\
        \cmidrule(lr){2-3} \cmidrule(lr){4-5} \cmidrule(lr){6-9} \cmidrule(lr){10-11} \cmidrule(lr){12-13}  
        & \textbf{Rank-1} & \textbf{TAR}\tnote{$\dagger$} & \textbf{Rank-1} & \textbf{TAR}\tnote{$\dagger$} & \multicolumn{4}{c}{\textbf{TAR}} & \textbf{Rank-1} & \textbf{TAR} & \textbf{Rank-1} & \textbf{TAR} \\
        \midrule
        DeepPrint \cite{DeepPrint} & 98.65 & 95.95 & 79.45 & 71.60 & 60.36 & 85.07 & 77.85 & 53.41 & 24.03 & 33.33 & 57.71 & 68.44 \\
        P-DeepPrint$^*$ \cite{DeepPrint} & 99.45 & 98.65 &  97.90 & 97.50 & 67.75 & 92.25 & 71.33 & 54.22 & 29.46 & 43.41 & 67.22 & 79.65 \\
        MultiScale \cite{gu2022latent} & 99.05 & 98.00 & 98.10 & 97.70 & 77.04 & 96.86 & 79.69 & 47.34 & 29.07 & 36.05 & 69.07 & 76.75 \\
        AFRNet \cite{grosz2024afrnet} & 97.45 & 94.70 & 96.80 & 96.05 & 83.68 & 98.61 & \textbf{90.42} & 70.13 & 31.78 & 41.86 & 68.91 & 78.65 \\
        P-AFRNet$^*$ \cite{grosz2024afrnet} & 99.00 & 97.70 & 95.85 & 92.95 & 52.54 & 78.54 & 71.57 & 12.36 & 18.60 & 28.68 & 63.04 & 50.19 \\
        \rowcolor{gray!7}
        FDD (binary) \cite{FDD} & 99.65 & \textbf{99.70} & 98.65 & 98.80 & 95.82 & 99.25 & 84.65 & 86.56 & 42.25 & 32.56 & 79.83 & 88.17 \\
        \rowcolor{gray!7}
        FDD \cite{FDD} & \textbf{99.75} & 99.60 & 98.65 & 98.75 & 95.86 & \textit{99.50} & 89.62 & 88.62 & \textit{51.94} & \textit{62.02} & 82.46 & \textit{90.79} \\
        \rowcolor{gray!20}
        FLARE (binary) & \textbf{99.75} & \textbf{99.70} & \textit{98.75} & \textit{98.90} & \textit{95.93} & 99.36 & 78.78 & \textit{97.43} & 50.00 & 54.26 & \textit{82.50} & 90.70 \\
        \rowcolor{gray!20}
        FLARE & \textbf{99.75} & \textbf{99.70} & \textbf{98.85} & \textbf{98.95} & \textbf{96.36} & \textbf{99.57} & \textit{89.95} & \textbf{98.11} & \textbf{60.85} & \textbf{69.38} & \textbf{85.12} & \textbf{93.33} \\
        \bottomrule
      \end{tabular}
      }
      \begin{tablenotes}
        \item[$\dagger$] TAR@FAR=0.01\%.
        \item[$\ddagger$] FVC02, FVC04, FVC06, and PolyU denote FVC2002 DB3A, FVC2004 DB1A, FVC2006 DB1A, and PolyU CL2CB, respectively. This naming convention is consistently used in Tab.~\ref{tab:enhancement_comp} and Tab.~\ref{tab:complementarity_quantitative}.
        \item[$*$] ``P-'' indicates the reimplemented methods in which the original STN is replaced by the pose estimation method of Duan et al.\cite{duan2023estimating}, consistent with the pose estimator used in MultiScale \cite{gu2022latent} and FDD \cite{FDD}.
      \end{tablenotes}
  \end{threeparttable}
\end{table*}

\begin{figure*}[!t]
  \centering
  \subfloat[PolyU CL2CB]{\includegraphics[height=.24\linewidth]{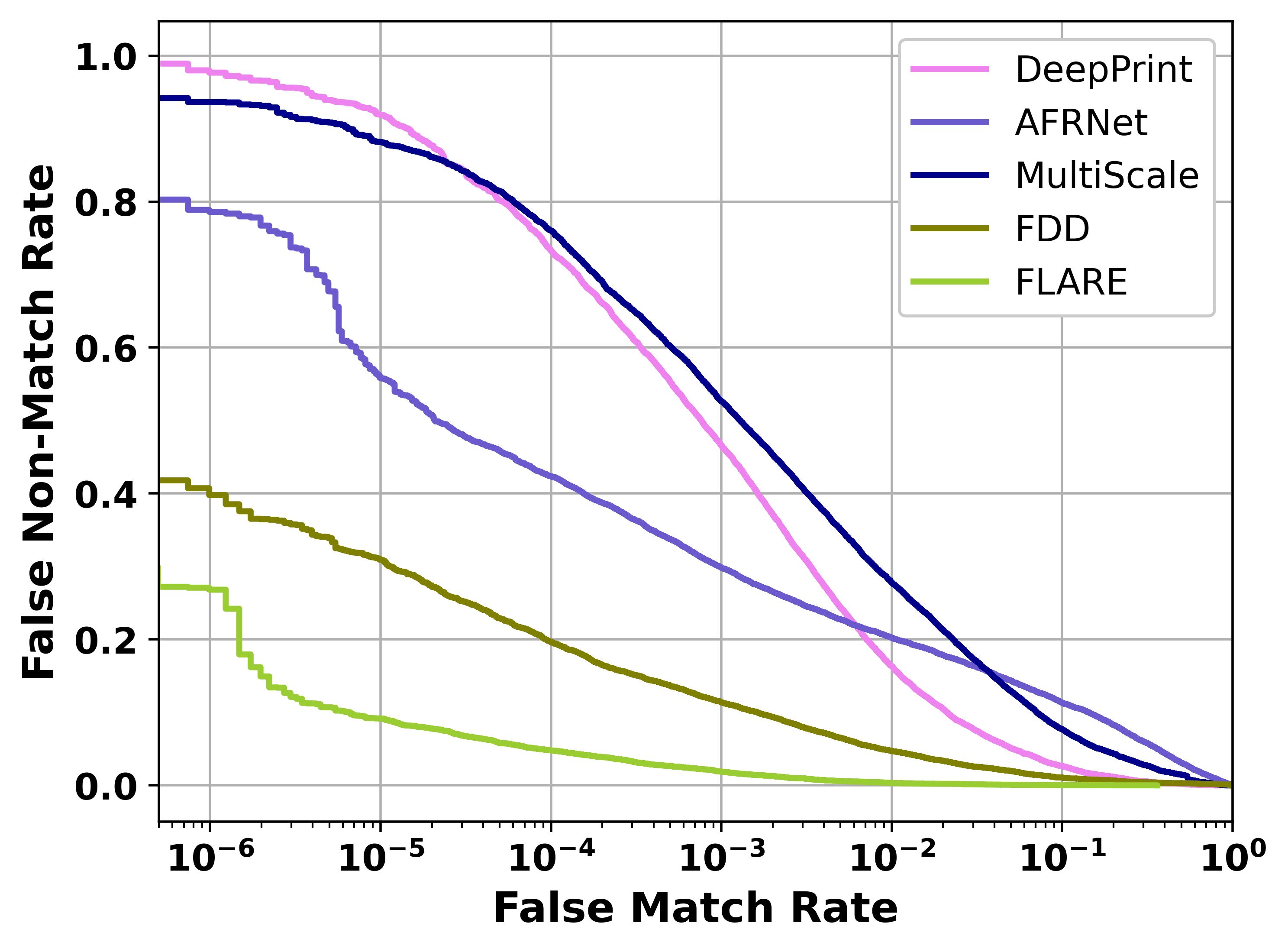}} \hfil
  \subfloat[NIST SD27]{\includegraphics[height=.24\linewidth]{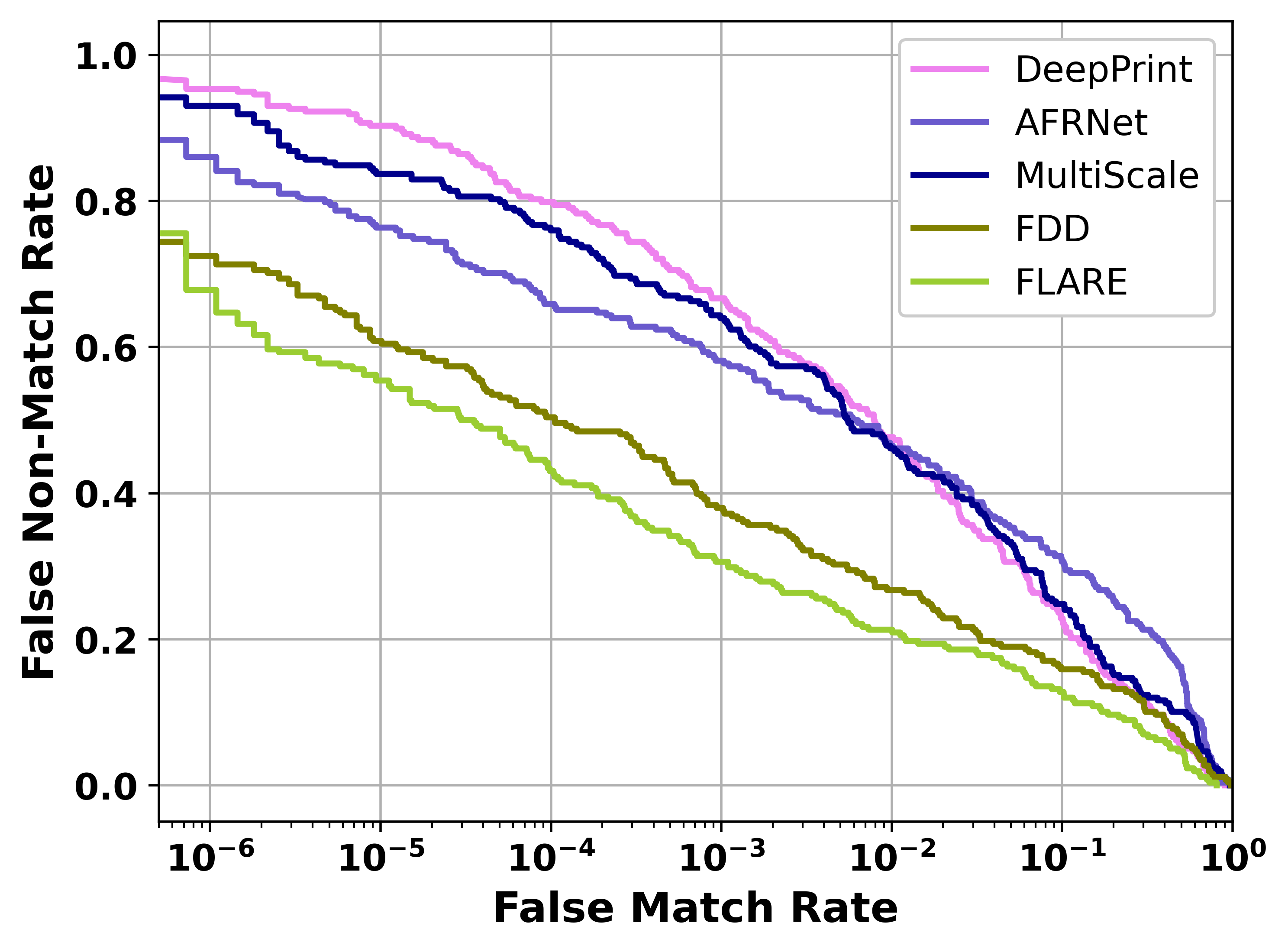}} \hfil
  \subfloat[THU Latent10K]{\includegraphics[height=.24\linewidth]{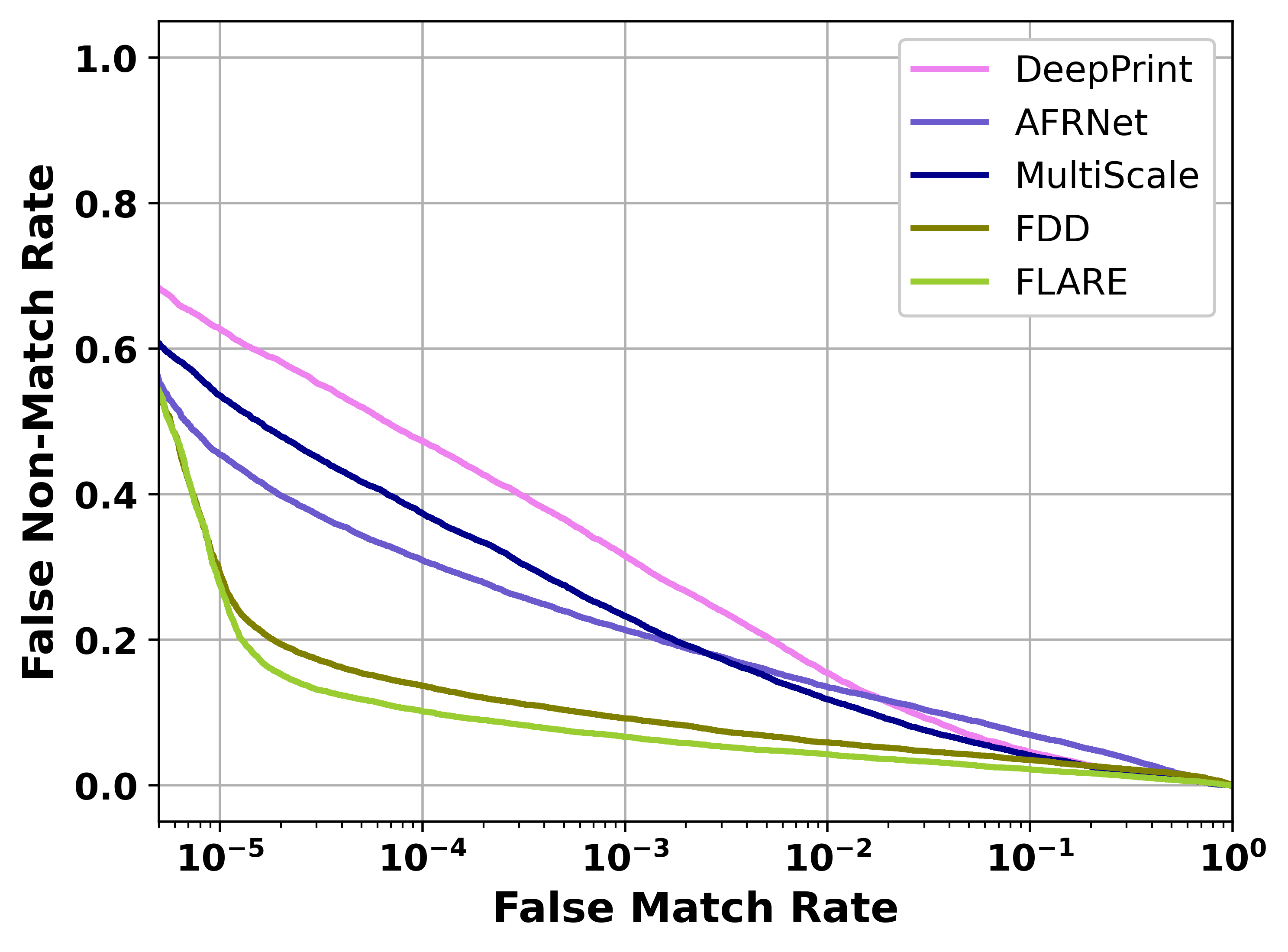}} \hfil
  \caption{DET curves for (a) contactless-to-contact matching on PolyU CL2CB, and (b, c) latent-to-contact matching on NIST SD27 and THU Latent10K, respectively.}
  \label{fig:det_curves}
\end{figure*}

\begin{figure}[!t]
  \subfloat[NIST SD27]{\includegraphics[width=.5\linewidth]{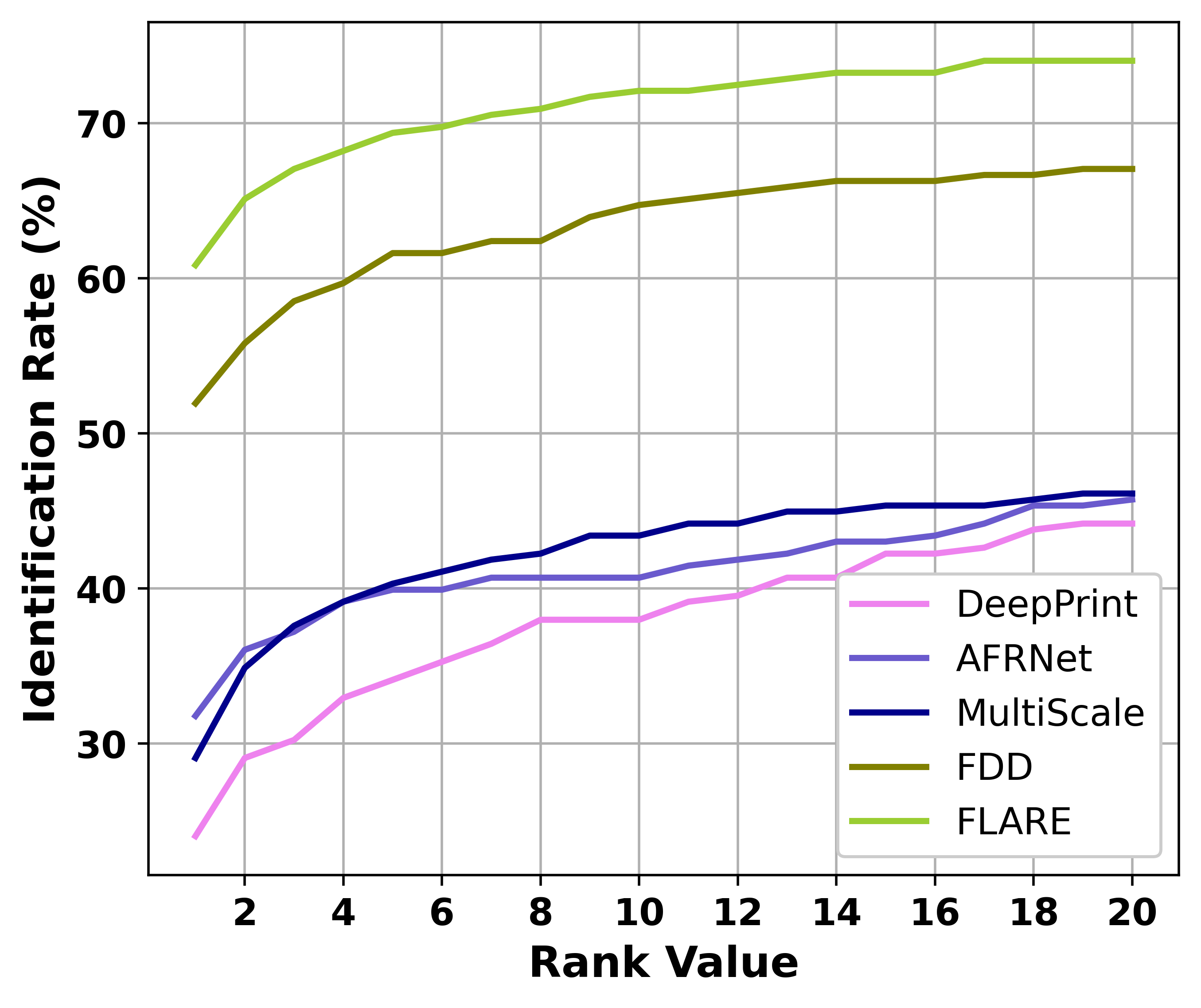}}  \hfil
  \subfloat[THU Latent10K]{\includegraphics[width=.5\linewidth]{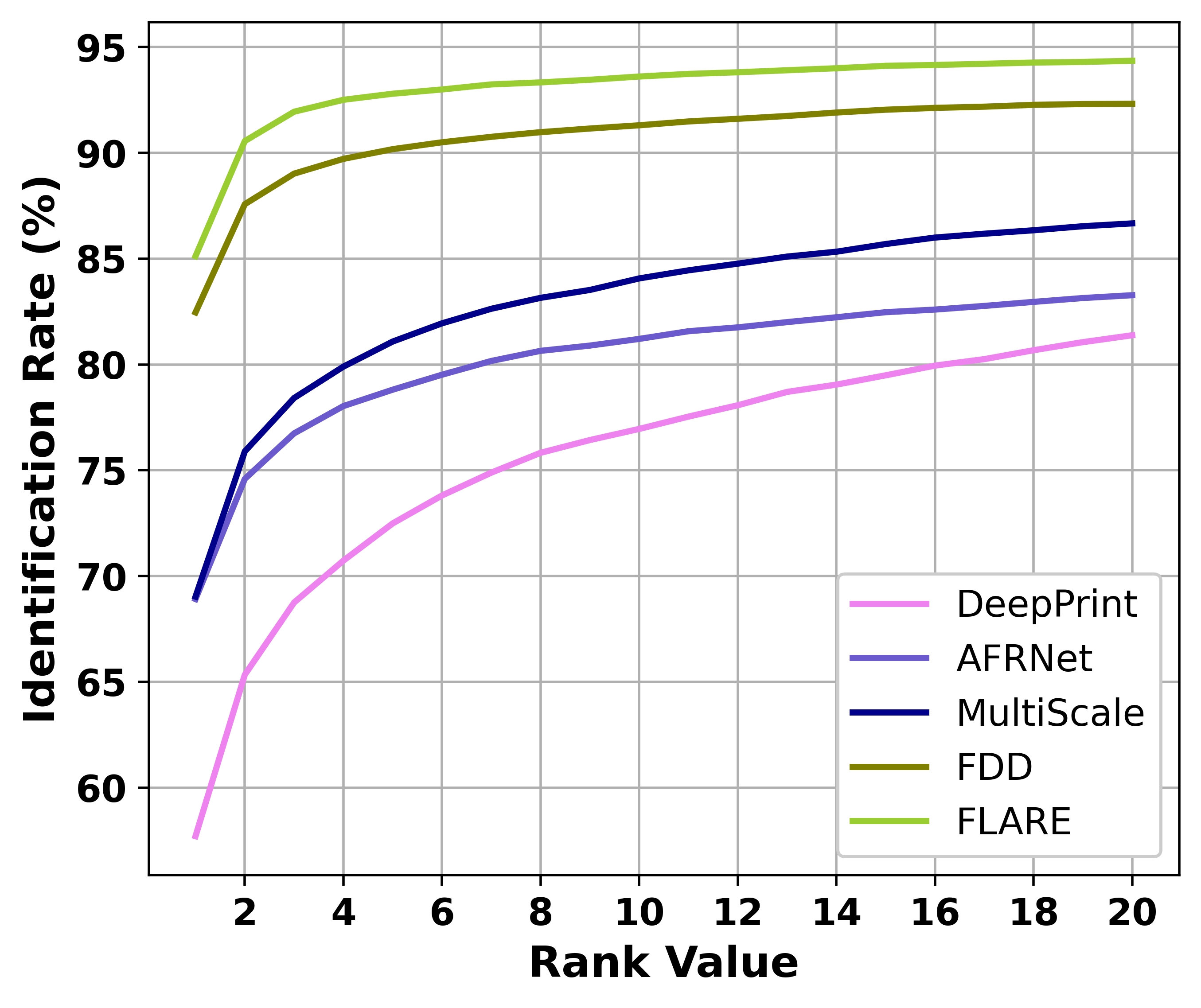}} \hfil
  \caption{CMC curves for latent-to-contact matching on (a) NIST SD27 and (b) THU Latent10K.}
  \label{fig:cmc_curves}
\end{figure}

\subsection{Fingerprint Matching Performance Comparison} %
We compare our method with several high-performing and representative fixed-length fingerprint matching approaches, including DeepPrint \cite{DeepPrint}, AFRNet \cite{grosz2024afrnet}, MultiScale \cite{gu2022latent}, and our previous conference work FDD \cite{FDD}. A summary of the key characteristics of these methods is provided in Tab. \ref{tab:method_desc}. Since the official implementations of DeepPrint \cite{DeepPrint} and AFRNet \cite{grosz2024afrnet} are not publicly available, we reimplemented them to the best of our ability based on the details provided in their respective papers. For DeepPrint \cite{DeepPrint}, we did not apply any additional quantization to its fingerprint representations, so as to preserve its best matching performance. For MultiScale \cite{gu2022latent} and FDD \cite{FDD}, which require fingerprint pose estimation as a preprocessing step, we employ the method of Duan et al. \cite{duan2023estimating} to estimate the 2D pose for all corresponding experiments. All fingerprint representation extraction methods are trained using the same dataset and data augmentation strategies. For AFRNet and DeepPrint, however, we additionally apply random rotations within $[-30\degree,30\degree]$ during training to better facilitate the learning of their STN-based alignment modules. To further isolate the influence of pose estimation on matching performance, we also reimplement DeepPrint and AFRNet without their STN modules and instead use the pose estimation of Duan et al. \cite{duan2023estimating} to normalize the fingerprint pose; these variants are denoted as P-DeepPrint and P-AFRNet, respectively. The evaluation metrics include Rank-1 accuracy for closed-set identification tasks, and True Accept Rates (TAR) at False Accept Rates (FAR) of 0.1\% and 0.01\% for open-set verification tasks. To provide a more comprehensive comparison of matching performance on challenging datasets such as latent fingerprints (which typically exhibit low image quality) and contactless fingerprints (which involve cross-modality matching), we plot the Detection Error Tradeoff (DET) curves for the relevant datasets, as shown in Fig. \ref{fig:det_curves}. In addition, we present the Cumulative Match Characteristic (CMC) curves on latent fingerprints to evaluate closed-set identification performance, as shown in Fig. \ref{fig:cmc_curves}. 

As shown in Tab.~\ref{tab:compared_method} and Fig.~\ref{fig:det_curves}--\ref{fig:cmc_curves}, both FDD and FLARE demonstrate strong performance across a wide range of fingerprint types—including rolled, plain, partial, contactless, and latent—outperforming prior fixed-length methods based on one-dimensional descriptors \cite{DeepPrint,gu2022latent,grosz2024afrnet} in most cases. To further isolate the effect of descriptor design from that of pose estimation, we also evaluate the performance of the P-DeepPrint and P-AFRNet variants. Their results show that replacing the STN with an external pose estimator does not consistently improve accuracy. This indicates that the gains achieved by FDD and FLARE primarily stem from our proposed fixed-length dense descriptors, which explicitly encode spatially localized fingerprint features, allowing the model to focus on foreground regions and suppress background noise. In contrast, one-dimensional descriptor approaches often rely on global aggregation without explicit modeling of foreground structures, making them more vulnerable to degraded inputs caused by noise or partial fingerprints. Moreover, we further evaluate the binary versions of FDD and FLARE, in which the dense descriptors are binarized using a simple zero-thresholding rule and matched with Hamming distance. Despite the drastic compression, both binary FDD and binary FLARE still outperform most existing methods on several benchmarks, further demonstrating the effectiveness and generality of dense representations.

Building upon FDD, FLARE further introduces fingerprint enhancement and incorporates multiple complementary pose estimation and enhancement strategies. These components not only improve the quality of the input images for descriptor extraction but also increase robustness against pose estimation inaccuracies. As a result, FLARE consistently achieves competitive or superior performance compared to FDD across all benchmarks. The benefits of enhancement are particularly pronounced in challenging scenarios. On fingerprints affected by significant noise or modality shifts—such as those in NIST SD27, THU Latent10K (latent), and PolyU CL2CB (contactless)—FLARE shows notable improvements, demonstrating enhanced ridge clarity and reduced background interference that translate to better matching accuracy. 

\subsection{Analysis of Dense Representations}
We conduct an ablation study of the fixed-length dense representation network. All experiments are conducted under the same configuration as FDD \cite{FDD}, using the pose estimation method of Duan et al. \cite{duan2023estimating} without applying any enhancement. We evaluate three variants: (1) removing the minutiae branch (w/o Mnt. Branch), (2) merging the two branches into a single one that jointly handles auxiliary tasks and descriptor extraction (Combined Branch), and (3) removing the 2D positional embedding (w/o Pos. Embedding). All variants retain the same descriptor dimensionality $f \in \mathbb{R}^{12\times16\times16}$ and are trained to convergence. As shown in Tab.~\ref{tab:ablation_fdrn}, excluding the minutiae branch leads to a notable performance drop in cross-modality matching, particularly on the contactless dataset PolyU CL2CB, underscoring the importance of the minutiae-aware representation in improving cross-modality matching generalization. Merging the branches reduces the overall representational capacity, especially under challenging conditions such as latent fingerprints. Finally, removing the positional embedding consistently impairs performance, highlighting its critical role in spatial encoding and matching reliability for degraded images. 

\begin{figure}[!t]
  \includegraphics[width=.9\linewidth]{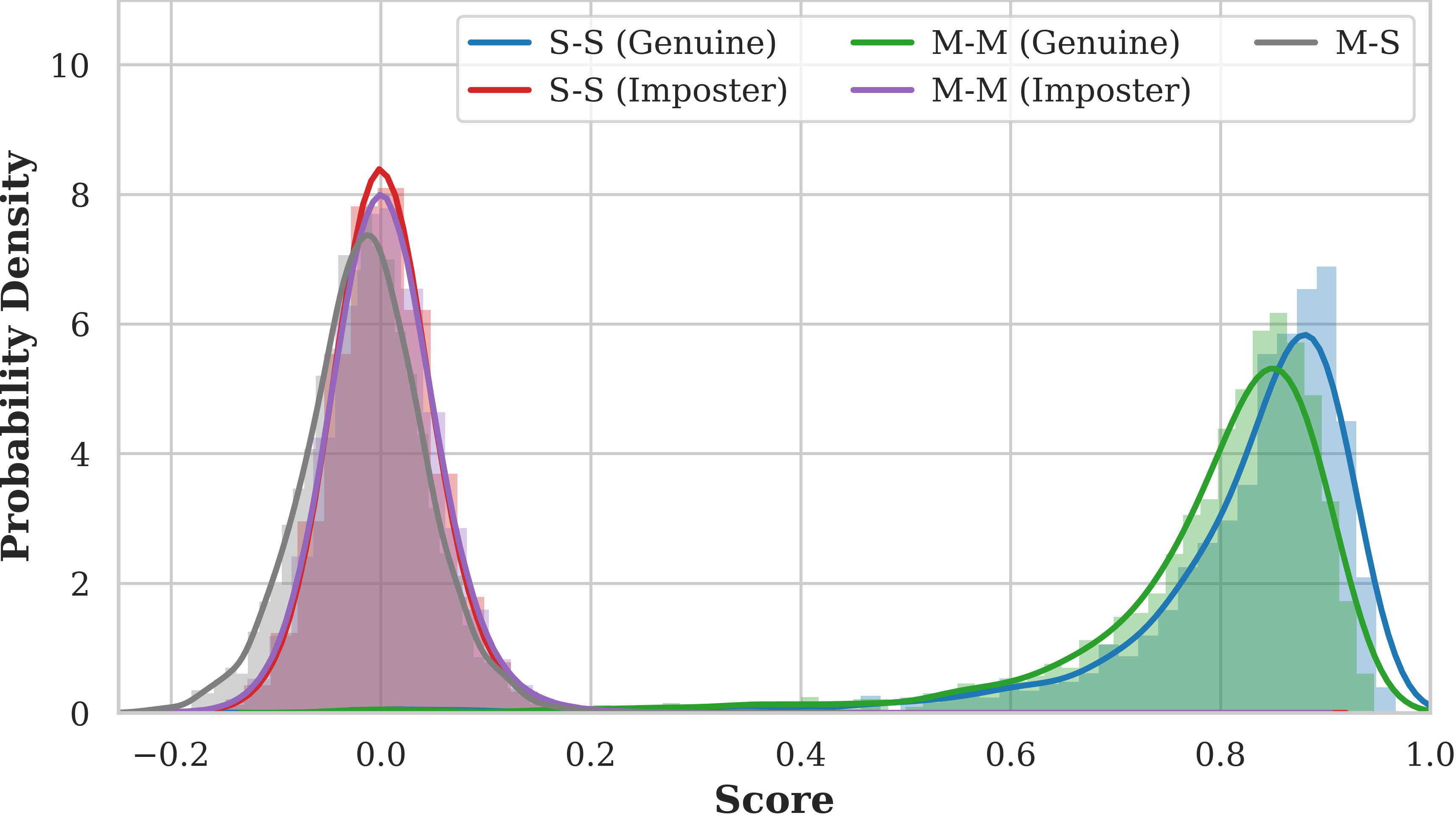}
  \caption{Distribution of cosine similarity scores between different types of descriptor pairs on the N2N Plain dataset. ``S-S'' and ``M-M'' denote structure-to-structure and minutiae-to-minutiae descriptor matching, respectively, while ``M-S'' indicates cross-matching between the two branches extracted from the same fingerprint.}
  \label{fig:ms_score_distribition}
\end{figure}

\begin{table}[!t]
  \centering
  \caption{Ablation evaluations on extraction of fixed-length dense representation network (FDRN). TAR@FAR=0.1\% is reported.}
  \label{tab:ablation_fdrn}
  \begin{tabular}{l|ccccc}
    \toprule
    \multirow{2}{*}{\textbf{Ablations}} & \multicolumn{2}{c}{\textbf{NIST SD27}} & \multicolumn{2}{c}{\textbf{THU Latent10K}} & \textbf{PolyU} \\
    \cmidrule{2-3} \cmidrule{4-5} \cmidrule{6-6} 
    & \textbf{Rank-1} & \textbf{TAR} & \textbf{Rank-1} & \textbf{TAR}  & \textbf{TAR} \\
    \midrule
    w/o Mnt. Branch & 42.25 & 54.65 & 78.65 & 87.71 & 72.76 \\
    Combined Branch & 43.02 & 53.49 & 78.69 & 88.34 & 87.76 \\
    w/o Pos. Embedding & 44.96 & 55.43 & 80.92 & 88.99 & 88.09 \\
    \hhline
    None & 51.94 & 62.02 & 82.46 & 90.79 & 88.62\\
    \bottomrule
  \end{tabular}
\end{table}

\begin{figure}[!t]
  \centering
  \subfloat[\label{fig:fdd_rolled}]{\includegraphics[width=.45\linewidth]{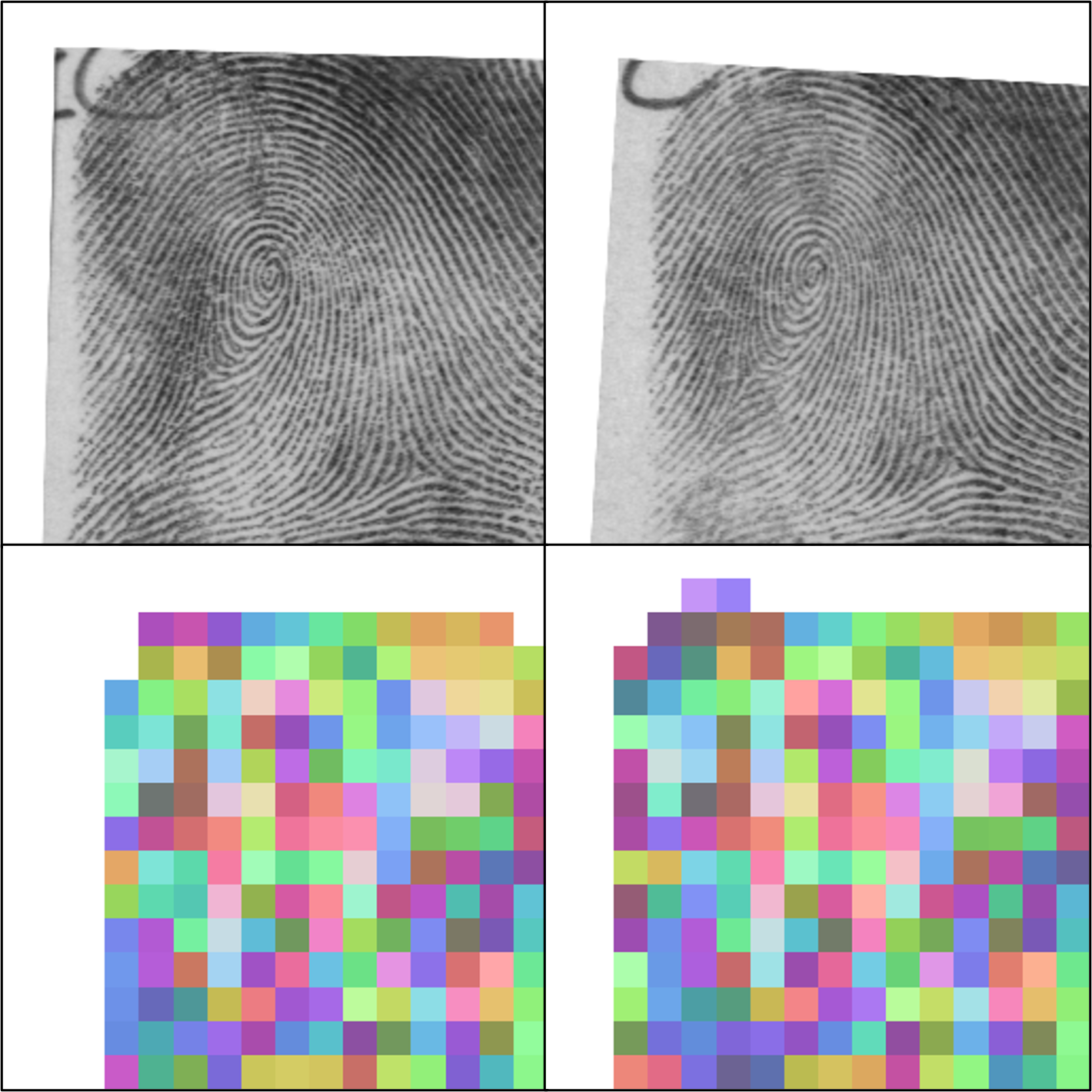}} \hfil
  \subfloat[\label{fig:fdd_plain}]{\includegraphics[width=.45\linewidth]{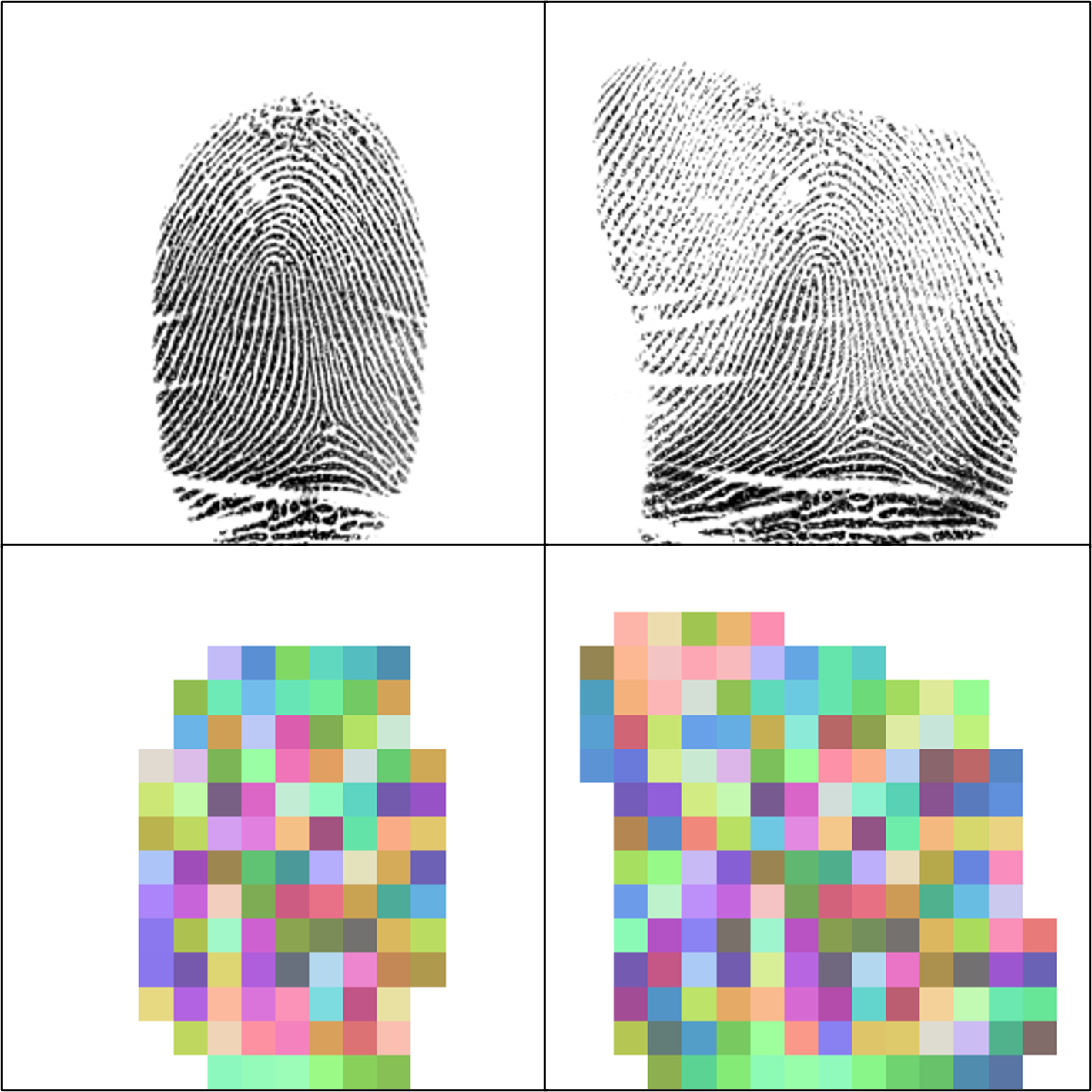}} \hfil
  \subfloat[\label{fig:fdd_partial}]{\includegraphics[width=.45\linewidth]{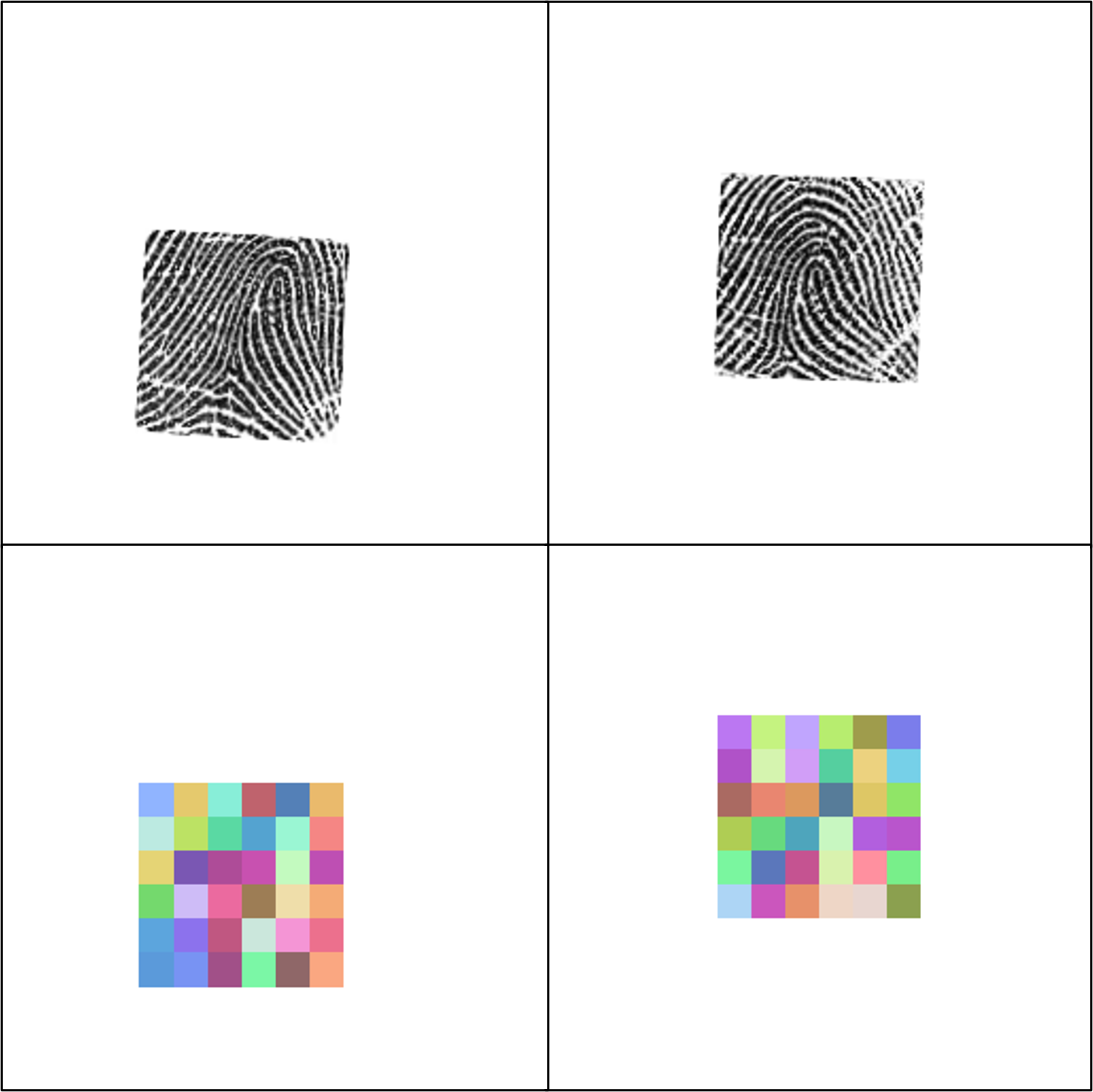}} \hfil
  \subfloat[\label{fig:fdd_contactless}]{\includegraphics[width=.45\linewidth]{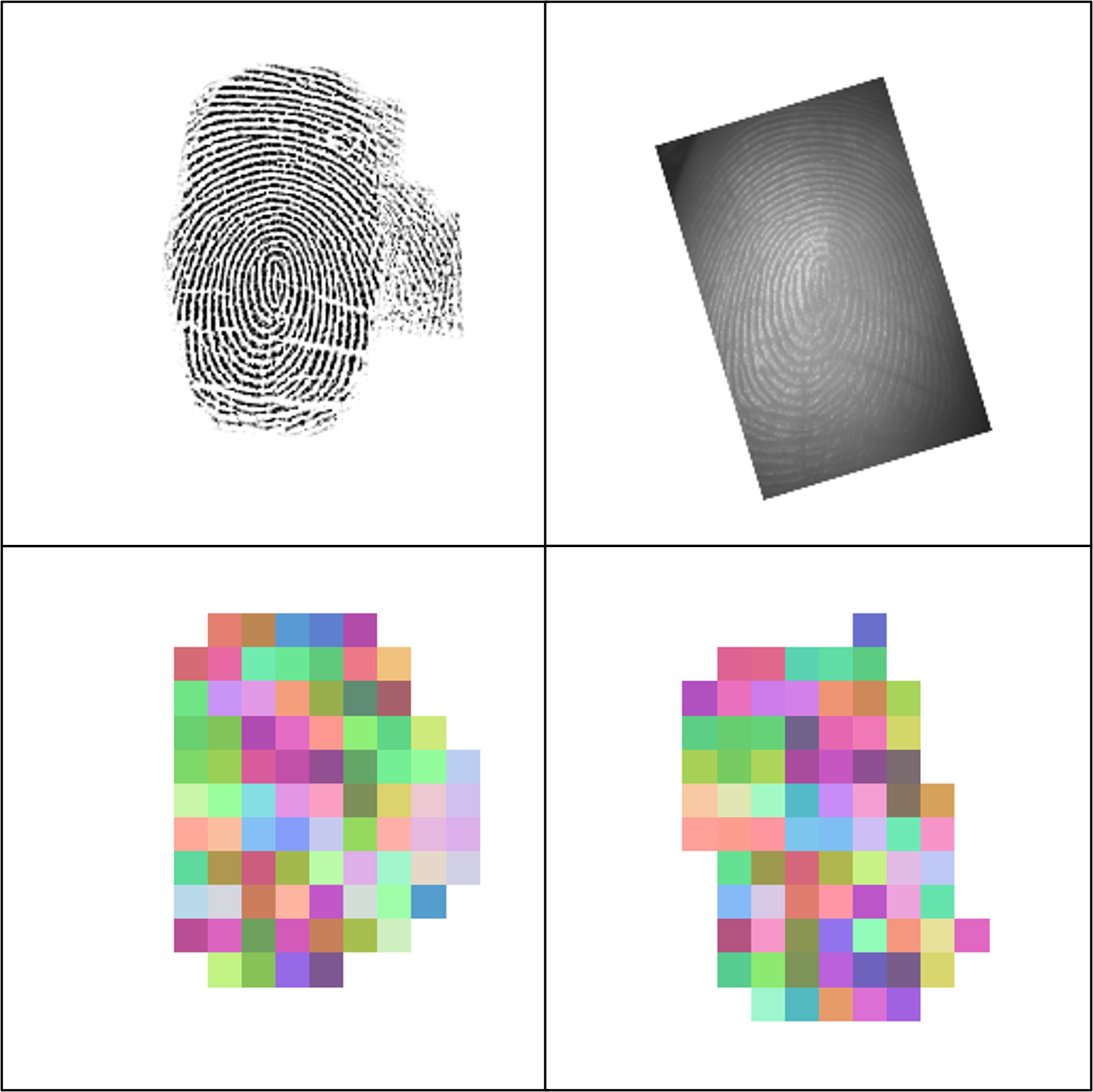}} \hfil
  \subfloat[\label{fig:fdd_latent1}]{\includegraphics[width=.45\linewidth]{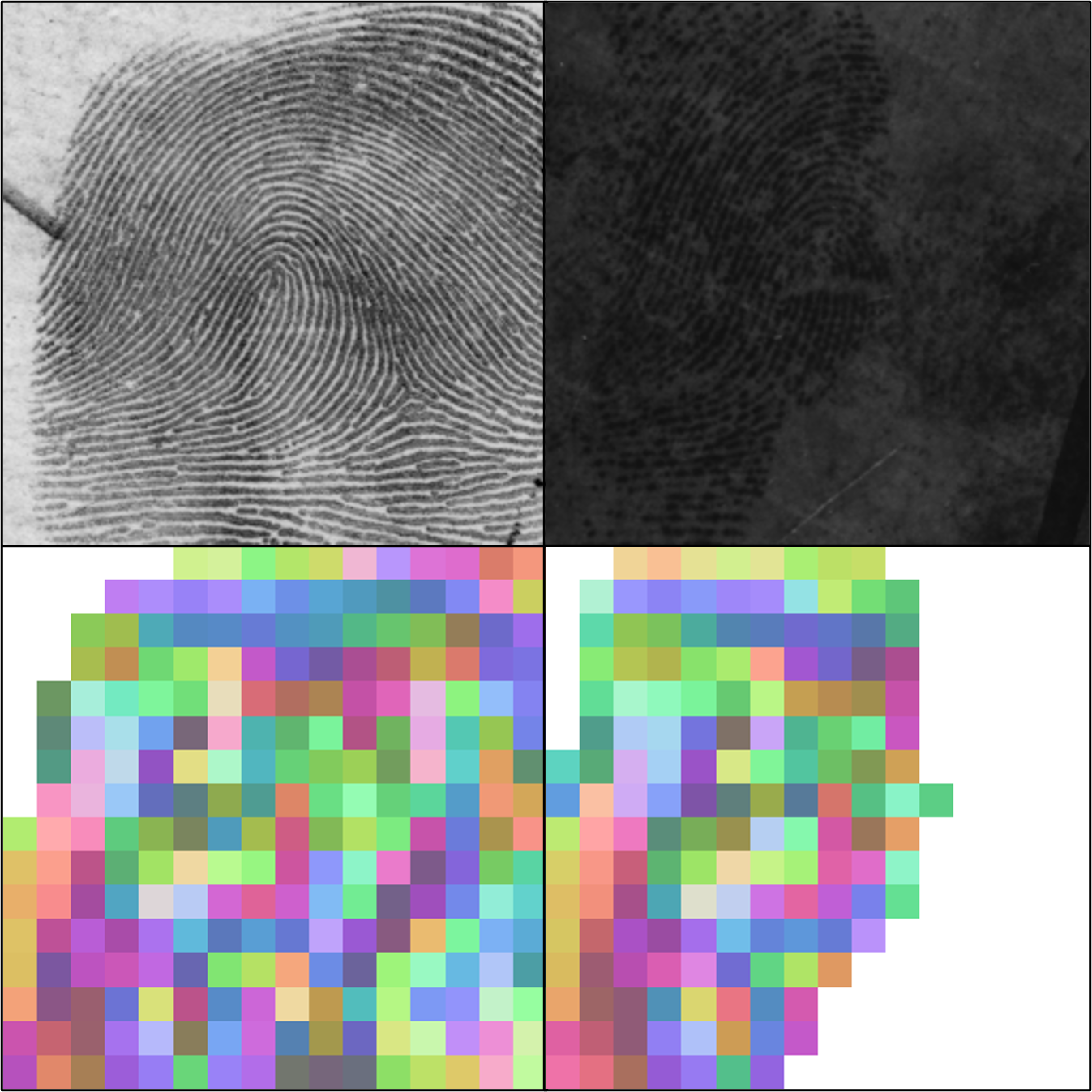}} \hfil
  \subfloat[\label{fig:fdd_latent2}]{\includegraphics[width=.45\linewidth]{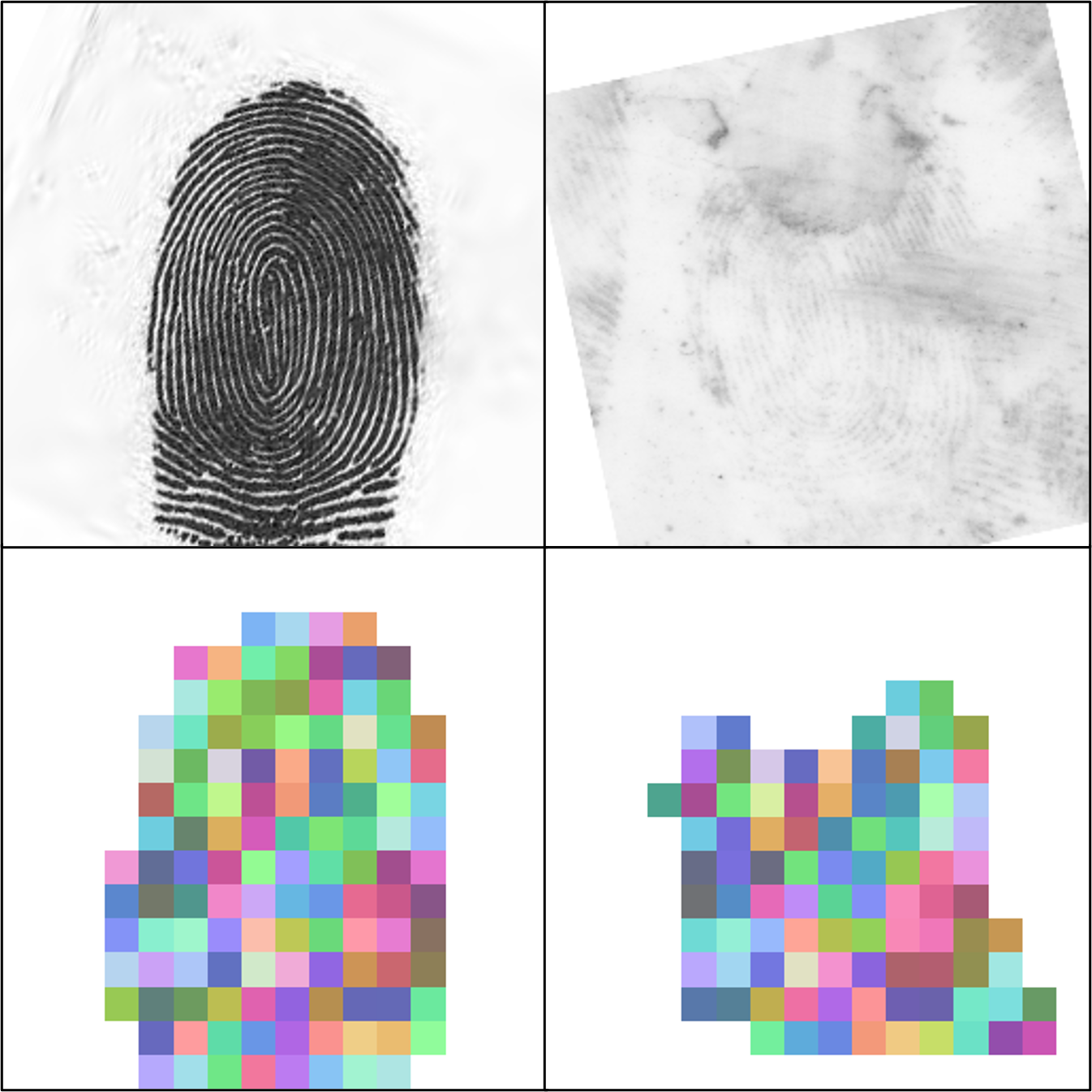}} \hfil
  \caption{Examples of the fixed-length dense representation extracted from genuine pairs of (a) NIST SD4, (b) N2N Plain, (c) FVC2006 DB1A, (d) PolyU CL2CB, (e) NIST SD27, (f) THU Latent10K. The fingerprint images shown in the figure have been aligned based on their estimated poses.}
  \label{fig:fdd_illus}
\end{figure}

\begin{figure}[!t]
  \centering
  \subfloat[\label{fig:sm1}]{\includegraphics[width=.9\linewidth]{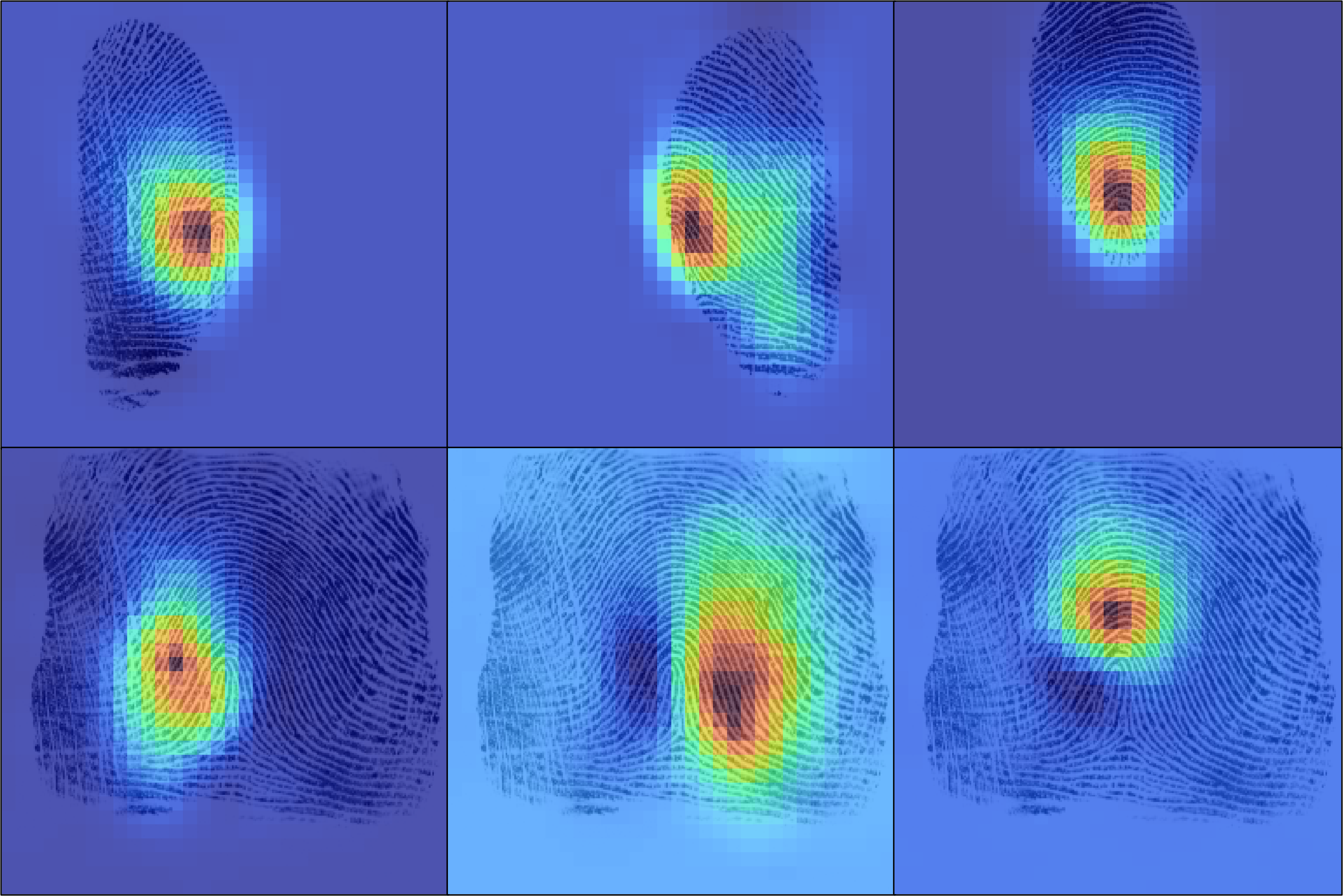}} \hfil
  \subfloat[\label{fig:sm2}]{\includegraphics[width=.9\linewidth]{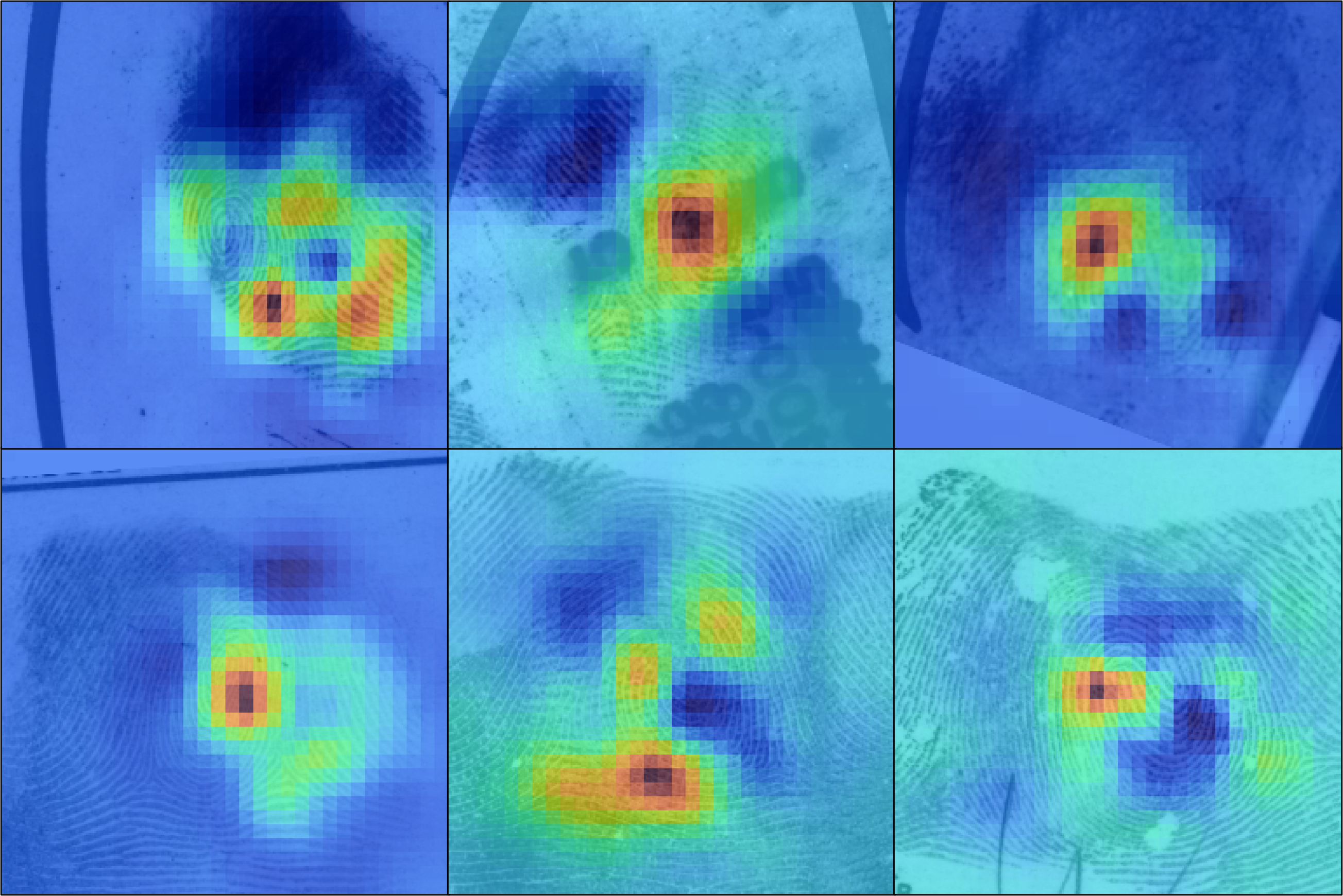}} \hfil
  \caption{Saliency map visualizations for: (a) same-identity fingerprints captured under different pressing poses; (b) incomplete latent fingerprints matched with their rolled reference.}
  \label{fig:saliency_map}
\end{figure}

To further validate the effectiveness of separately extracting minutiae-aware and structure-aware descriptors, we analyze the similarity distributions of the two branches. Specifically, we evaluate the matching scores between $f_\text{m}$ and $f_\text{s}$ extracted from the same image, as well as genuine and imposter pairs of $f_\text{m}$ (M-M) and $f_\text{s}$ (S-S) on the N2N Plain dataset. The score distributions are illustrated in Fig.~\ref{fig:ms_score_distribition}. We observe that the similarity between $f_\text{m}$ and $f_\text{s}$ resembles that of imposter distributions, indicating that the two branches capture complementary information without redundancy. 

Fig.~\ref{fig:fdd_illus} illustrates the spatial characteristics of the proposed dense representation, dense representations, where meaningful descriptor responses are concentrated within fingerprint foreground regions while background areas are suppressed. For clearer visualization of these properties, all inputs shown here are the pose-normalized fingerprint images used directly for descriptor extraction. These examples further highlight the robustness of the dense descriptors across a variety of matching scenarios. For rolled fingerprint matching (Fig.~\ref{fig:fdd_rolled}), the dense representations remain highly consistent across different impressions of the same finger. In plain fingerprint matching (Fig.~\ref{fig:fdd_plain}), strong correspondence is observed within the overlapping ridge regions. For partial fingerprints (Fig.~\ref{fig:fdd_partial}), the descriptors remain stable within the shared foreground area despite significant cropping. In contactless-to-contact matching (Fig.~\ref{fig:fdd_contactless}), the representations exhibit notable robustness to modality-induced appearance variations. For latent fingerprint matching, both examples shown in Fig.~\ref{fig:fdd_latent1} and Fig.~\ref{fig:fdd_latent2} demonstrate that background clutter is effectively suppressed and meaningful structural consistency is preserved within valid ridge regions.

In addition, we use sliding-window occlusion to generate saliency maps that further illustrate the matching behavior of the dense descriptors. As shown in Fig.~\ref{fig:saliency_map}, regions closer to the fingerprint core contribute more discriminative information and exert a greater influence on the final matching score. Moreover, in cases where the input fingerprint is incomplete—either due to variations in pressing pose (Fig.~\ref{fig:sm1}) or partial acquisition in latent fingerprints (Fig.~\ref{fig:sm2})—the saliency responses on the rolled reference concentrate only within the regions that spatially overlap with the valid portion of the query fingerprint. This behavior indicates that the dense descriptors naturally restrict effective matching to the overlapping foreground regions of the two fingerprints, reinforcing their spatial locality and robustness to missing areas.

These results highlight the intrinsic robustness of the dense descriptor. Furthermore, we analyze how enhancement strategies contribute to quality improvement for more reliable descriptor extraction and matching (Sec.~\ref{sec:enhancement_ex}), and how pose estimation improves robustness to input pose variations, as detailed in the supplementary material.

\begin{figure*}[!t]
  \centering
  \includegraphics[width=.9\linewidth]{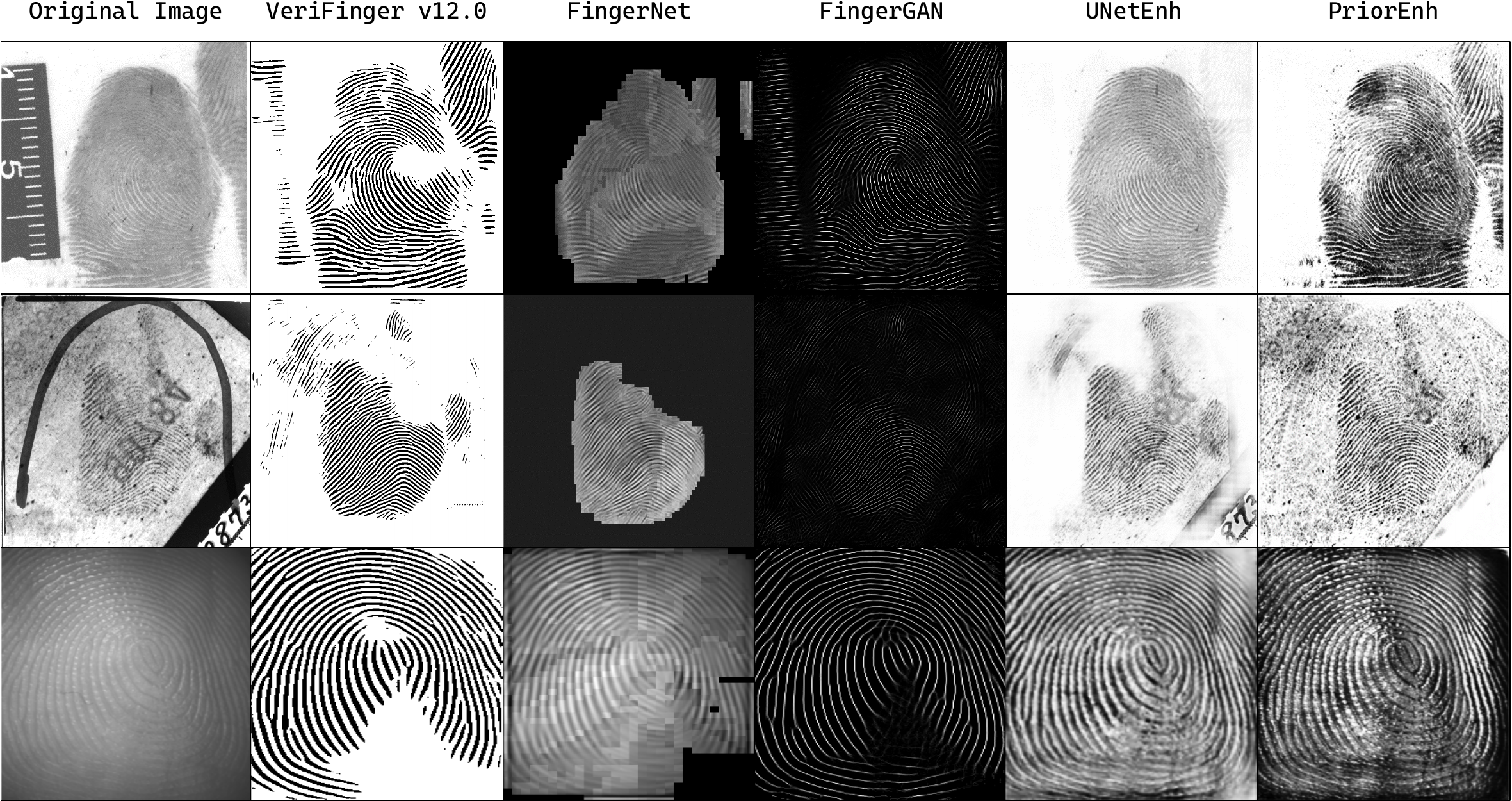}
  \caption{Qualitative comparison of fingerprint enhancement results across different methods. Each row shows a representative fingerprint from THU Latent10K, NIST SD27, and PolyU CL2CB, respectively.}
  \label{fig:enh_qualitive}
\end{figure*}

\begin{table*}[!t]
  \centering
  \caption{Matching Performance (\%) with Different Enhancement Methods. Values are reported as relative differences from the original image baseline. Unless otherwise specified, TAR@FAR = 0.1\% is reported.
  }
  \label{tab:enhancement_comp}
  \renewcommand{\arraystretch}{1.2}
  \begin{threeparttable}
    \resizebox{\textwidth}{!}{
      \begin{tabular}{l
        c*{2}{>{\centering\arraybackslash}p{0.05\linewidth}}
        c*{1}{>{\centering\arraybackslash}p{0.05\linewidth}}
        c*{1}{>{\centering\arraybackslash}p{0.05\linewidth}}
        c*{4}{>{\centering\arraybackslash}p{0.05\linewidth}}
        c*{4}{>{\centering\arraybackslash}p{0.05\linewidth}}}
        \toprule
        \multirow{2}{*}{\textbf{\makecell[l]{Enh.\\Method}}} & \multicolumn{2}{c}{\textbf{NIST SD4}} & \multicolumn{2}{c}{\textbf{N2N Plain}} & \multicolumn{1}{c}{\textbf{FVC02}} & \multicolumn{1}{c}{\textbf{FVC04}} & \multicolumn{1}{c}{\textbf{FVC06}} & \multicolumn{1}{c}{\textbf{PolyU}} & \multicolumn{2}{c}{\textbf{NIST SD27}} & \multicolumn{2}{c}{\textbf{THU Latent10K}} \\
        \cmidrule(lr){2-3} \cmidrule(lr){4-5} \cmidrule(lr){6-9} \cmidrule(lr){10-11} \cmidrule(lr){12-13}  
        & \textbf{Rank-1} & \textbf{TAR}\tnote{$\dagger$} & \textbf{Rank-1} & \textbf{TAR}\tnote{$\dagger$} & \multicolumn{4}{c}{\textbf{TAR}} & \textbf{Rank-1} & \textbf{TAR} & \textbf{Rank-1} & \textbf{TAR} \\
        \midrule
        Orginal Image & 99.75 & 99.60 & 98.65 & 98.75 & 95.86 & 99.50 & 89.62 & 88.62 & 51.94 & 62.02 & 82.46 & 90.79 \\
        \hhline
        VeriFinger \cite{nist2020verifinger} & \textcolor{red}{-0.90} & \textcolor{red}{-1.10} & \textcolor{red}{-1.35} & \textcolor{red}{-1.70} & \textcolor{red}{-0.11} & \textcolor{red}{-0.57} & \textcolor{red}{-3.07} & \textcolor{red}{-5.93} & \textcolor{red}{-4.65} & \textcolor{red}{-10.47} & \textcolor{red}{-5.22} & \textcolor{red}{-4.51} \\
        FingerNet \cite{tang2017fingernet} & \textcolor{red}{ -0.50} & \textcolor{red}{ -0.50} & \textcolor{red}{ -1.35} & \textcolor{red}{ -1.70} & \textcolor{red}{ -0.79} & \textcolor{red}{ -1.07} & \textcolor{red}{ -8.99} & \textcolor{red}{ -2.22} & \textcolor{red}{ -1.16} & \textcolor{red}{ -1.94} & \textcolor{red}{ -1.92} & \textcolor{red}{ -1.57} \\
        FingerGAN \cite{FingerGAN} & \textcolor{red}{ -0.65} & \textcolor{red}{ -0.75} & \textcolor{red}{ -4.45} & \textcolor{red}{ -5.75} & \textcolor{red}{ -20.43} & \textcolor{red}{ -4.82} & \textcolor{red}{ -9.18} & \textcolor{red}{ -9.66} & \textcolor{red}{ -17.44} & \textcolor{red}{ -15.90} & \textcolor{red}{ -12.25} & \textcolor{red}{ -9.69} \\
        UNetEnh & \textcolor{red}{-0.10} & \textcolor{red}{-0.10} & \textcolor{red}{ -0.30} & \textcolor{red}{-0.25} & \textcolor{red}{ -0.29} & \textcolor{red}{ -0.32} & \textcolor{red}{ -1.22} & \textcolor{darkgreen}{ +7.83} & \textcolor{darkgreen}{ +5.81} & \textcolor{darkgreen}{  +5.42} & \textcolor{darkgreen}{ +0.60} & \textcolor{darkgreen}{  +0.19} \\
        PriorEnh & 0.00 & \textcolor{darkgreen}{ +0.05} & \textcolor{red}{-0.05} & \textcolor{red}{ -0.15} & \textcolor{darkgreen}{ +0.07} & \textcolor{red}{ -0.11} & \textcolor{darkgreen}{ +0.01} & \textcolor{darkgreen}{  +7.49} & \textcolor{darkgreen}{ +4.26} & \textcolor{darkgreen}{ +0.38} & \textcolor{darkgreen}{  +0.74} & \textcolor{darkgreen}{+0.70} \\
        \bottomrule 
      \end{tabular}}
      \begin{tablenotes}
        \item[$\dagger$] TAR@FAR=0.01\%.
      \end{tablenotes}
  \end{threeparttable}  
\end{table*}

\begin{table}[!t]
  \centering
  \caption{Correct and Spurious Minutiae for 258 latent fingerprints in NIST SD27 before and after enhancement. Minutiae are extracted by VeriFinger v12.0 \cite{nist2020verifinger}.}
  \label{tab:enh_restoration}
  \begin{threeparttable}
    \resizebox{0.8\linewidth}{!}{
    \begin{tabular}{lcc}
      \toprule
      \textbf{Enhancement Method} & \textbf{\# Correct} & \textbf{\# Spurious} \\
      \midrule
      Original Image & 3,296 & \textbf{8,784} \\
      VeriFinger \cite{nist2020verifinger} & 3,275 & \textit{11,409}  \\
      FingerNet \cite{tang2017fingernet} & 3,675 & 16,937 \\
      FingerGAN \cite{FingerGAN} & 3,156 & 62,336 \\
      UNetEnh & \textbf{3,933} & 30,864 \\
      PriorEnh & \textit{3,764} & 23,872 \\
      \bottomrule
    \end{tabular}}
  \end{threeparttable}
\end{table}

\subsection{Evaluation of Enhancement Methods} \label{sec:enhancement_ex}
In this section, we compare our enhancement methods, UNetEnh and PriorEnh, with representative approaches including VeriFinger v12.0 \cite{nist2020verifinger}, FingerNet \cite{tang2017fingernet}, and FingerGAN \cite{FingerGAN}. For FingerNet and FingerGAN, we use the officially released pretrained models, while VeriFinger is evaluated using the commercial API we obtained through purchase. As shown in Fig.~\ref{fig:enh_qualitive}, results on three fingerprint types—latent (THU Latent10K, NIST SD27) and contactless (PolyU CL2CB)—demonstrate clear visual differences. VeriFinger and FingerGAN tend to misinterpret background noise as ridge patterns, while FingerNet often exhibits blocky artifacts that compromise continuity. In contrast, our methods enhance contrast while preserving the original ridge structures, without introducing hallucinated textures. This structural fidelity contributes to better compatibility with descriptor extraction and ultimately improved matching performance. 

We further evaluate how different enhancement methods affect the performance of descriptor-based matching. In this experiment, we fix the pose alignment method to that of Duan et al. \cite{duan2023estimating} and use the fixed-length dense descriptor for representation, varying only the enhancement module to assess its impact on overall matching accuracy. Enhancement methods such as VeriFinger \cite{nist2020verifinger}, FingerNet \cite{tang2017fingernet}, and FingerGAN \cite{FingerGAN} have proven effective in restoring local ridge structures. However, they substantially modify the fingerprint modality by imposing strong texture priors or altering ridge patterns, which often introduce noise or blurring artifacts. As shown in Tab.~\ref{tab:enhancement_comp}, such modifications adversely affect the matching of dense representations due to their high sensitivity to global texture consistency, leading to consistent performance degradation across all evaluated datasets. In contrast, our proposed UNetEnh offers limited improvements on clean contact-based fingerprints but yields notable gains on noisy latent and contactless fingerprints. PriorEnh, which incorporates prior ridge information, further improves performance across both low-quality and high-quality datasets. We also evaluate a variant where the descriptor extraction is retrained on enhanced images corresponding to each enhancement method. The overall conclusions remain consistent, as detailed in the supplementary material. Overall, both UNetEnh and PriorEnh maintain stable accuracy on clean contact-based fingerprints while substantially boosting matching robustness under challenging conditions such as noise and modality gaps. 

Moreover, we evaluate ridge restoration quality using minutiae annotations from Feng et al. \cite{feng2013orientation} on NIST SD27. Extacted minutiae are obtained by VeriFinger. As shown in Tab.~\ref{tab:enh_restoration}, UNetEnh and PriorEnh recover the most correct minutiae, though they also introduce more spurious ones. This is mainly due to retained background textures when noise cannot be fully suppressed, which may mislead VeriFinger. Nonetheless, our applied dense representations remains robust by focusing on foreground regions.

\begin{table*}
  \centering
  \caption{Matching accuracy (\%) on several fingerprint datasets with different methods combinations. TAR@FAR = 0.1\% is reported.}
  \label{tab:complementarity_quantitative}
  \renewcommand{\arraystretch}{1.2}
  \begin{threeparttable}
    \resizebox{\textwidth}{!}{
      \begin{tabular}{
        c*{1}{>{\centering\arraybackslash}p{0.08\linewidth}}
        c*{1}{>{\centering\arraybackslash}p{0.08\linewidth}}|
        c*{1}{>{\centering\arraybackslash}p{0.05\linewidth}}
        c*{1}{>{\centering\arraybackslash}p{0.05\linewidth}}
        c*{4}{>{\centering\arraybackslash}p{0.05\linewidth}}
        c*{4}{>{\centering\arraybackslash}p{0.05\linewidth}}
        }
        \toprule
        \multicolumn{2}{c}{\textbf{Pose Estimation}} & \multicolumn{2}{c|}{\textbf{Enhancement}} & \multicolumn{1}{c}{\textbf{FVC02}} & \multicolumn{1}{c}{\textbf{FVC04}} & \multicolumn{1}{c}{\textbf{FVC06}} & \multicolumn{1}{c}{\textbf{PolyU}} & \multicolumn{2}{c}{\textbf{NIST SD27}} & \multicolumn{2}{c}{\textbf{THU Latent10K}} \\
        \cmidrule(lr){1-2} \cmidrule(lr){3-4} \cmidrule(lr){5-8} \cmidrule(lr){9-10} \cmidrule(lr){11-12}  
        \textbf{Duan \cite{duan2023estimating}} & \textbf{Guan \cite{guan2025robust}} & \textbf{PriorEnh} & \textbf{UNetEnh}  & \multicolumn{4}{c}{\textbf{TAR}} & \textbf{Rank-1} & \textbf{TAR} & \textbf{Rank-1} & \textbf{TAR} \\
        \midrule
        $\checkmark$ &  &  &  & 95.86 & 99.50 & 89.52 & 88.62 & 51.94 & 62.02 & 82.46 & 90.79 \\
        $\checkmark$ & & $\checkmark$ & & 95.93 & 99.39 & 89.63 & 96.11 & 56.20 & 62.40 & 83.20 & 91.49 \\
        $\checkmark$ & & $\checkmark$ & $\checkmark$ & 96.25 & 99.36 & 89.03 & 97.28 & 59.69 & 65.89 & 83.84 & 92.04 \\
        & $\checkmark$ &  &  & 91.32 & \textbf{99.57} & 88.11 & 88.17 & 52.71 & 65.12 & 82.94 & 91.46 \\
        & $\checkmark$ & $\checkmark$ & & 93.21 & 99.36 & 88.16 & 95.98 & 53.49 & 62.79 & 83.75 & 92.10 \\
        & $\checkmark$ & $\checkmark$ & $\checkmark$ & 93.75 & 99.39 & 87.35 & 97.07 & 57.75 & 66.28 & 83.95 & 91.98 \\
        \rowcolor{gray!20}
        $\checkmark$ & $\checkmark$ & $\checkmark$ & $\checkmark$ & \textbf{96.36} & \textbf{99.57} & \textbf{89.95} & \textbf{98.11} & \textbf{60.85} & \textbf{69.38} & \textbf{85.12} & \textbf{93.33} \\
        \bottomrule
      \end{tabular}}
  \end{threeparttable}
\end{table*}

\begin{figure}[!t]
  \centering
  \includegraphics[width=\linewidth]{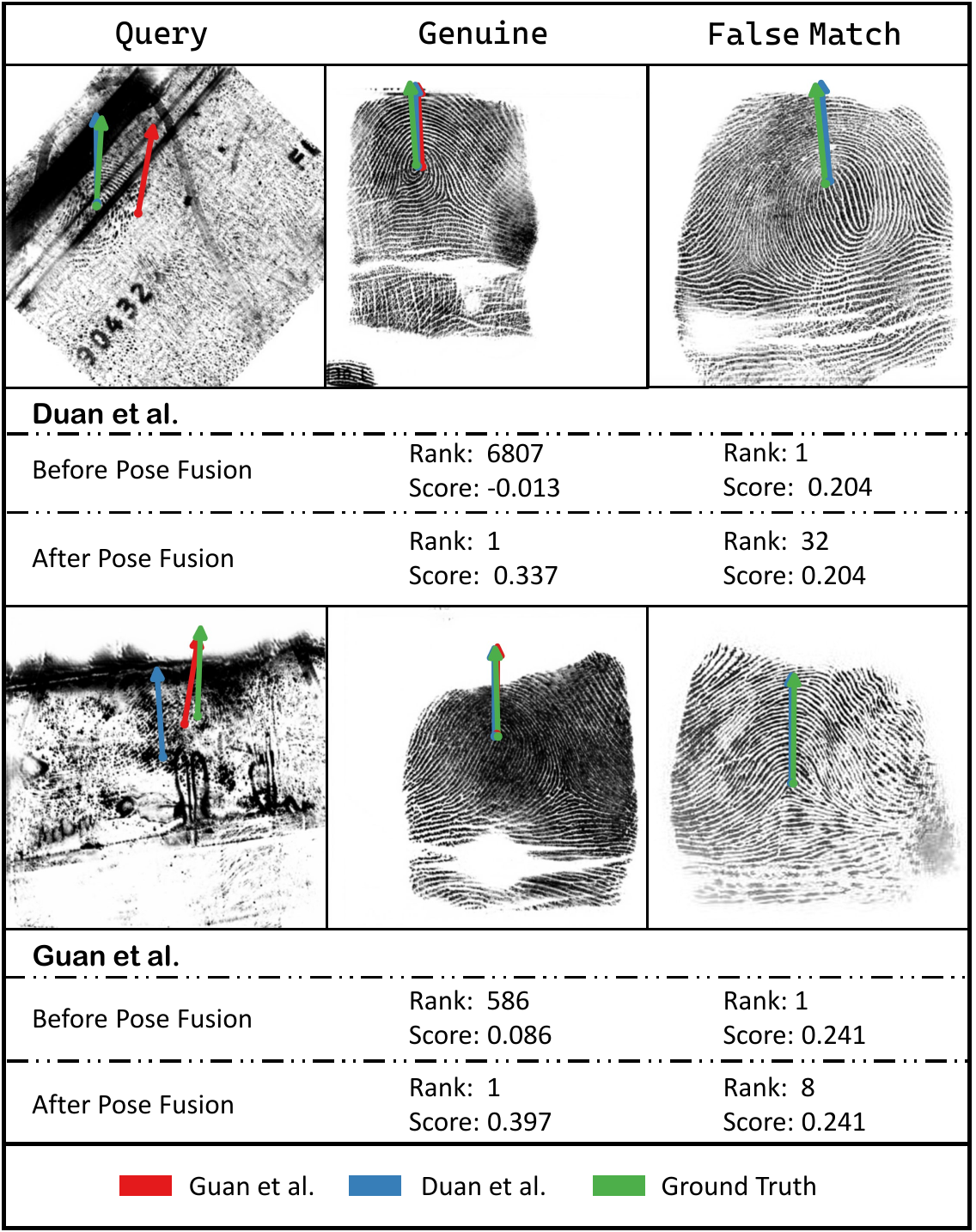}
  \caption{Illustration of the effectiveness of pose fusion. The top and bottom examples show two query cases with their genuine and false matches. Arrows indicate pose estimation from Duan et al. \cite{duan2023estimating} (blue), Guan et al. \cite{guan2025robust} (red), and the ground truth (green).}
  \label{fig:pose_complement}
\end{figure}

\subsection{Complementary Effects of Pose and Enhancement Strategies}
This section analyzes the complementarity of dual pose estimation and enhancement strategies in FLARE and their contribution to improving descriptor matching performance.
Tab. \ref{tab:enhancement_comp} and Tab. \ref{tab:enh_restoration} show that UNetEnh and PriorEnh exhibit different strengths in recovering ridge details and supporting descriptor extraction across varying image conditions. For pose estimation, Fig.~\ref{fig:pose_complement} illustrates that when individual methods fail due to different types of estimation errors, fusing their results can effectively correct misalignments and recover the true match, thereby improving the overall matching accuracy.

Building on these intuitive examples, we now present quantitative results to demonstrate the incremental benefits brought by combining pose estimation and enhancement strategies. Tab.~\ref{tab:complementarity_quantitative} presents a step-by-step quantitative evaluation across various fingerprint datasets. Starting from the baseline setup using a single pose estimation and no enhancement, we incrementally incorporate different pose estimators and enhancement modules. As shown in the table, each component contributes to performance gains on multiple datasets. The final configuration, combining both Duan et al.~\cite{duan2023estimating} and Guan et al.~\cite{guan2025robust} for pose estimation, along with both UNetEnh and PriorEnh for enhancement, achieves the best results across all evaluation sets. These findings confirm that the complementary nature of the selected methods leads to more robust and accurate fingerprint matching under diverse conditions. 

\subsection{Assessment of Space and Computational Complexity}
\label{sec:effi}

We further evaluate the time and space complexity of representative minutiae-based and fixed-length representation methods. In the preprocessing stage, minutiae-based approaches incur the cost of detecting minutiae, whereas VeriFinger\cite{nist2020verifinger} does not require separate preprocessing because minutiae extraction is integrated into its template generation procedure. MultiScale \cite{gu2022latent} and FDD \cite{FDD} adopt the pose estimation method of Duan et al. \cite{duan2023estimating}, while FLARE performs preprocessing through four combinations of pose estimation and enhancement. The detailed results are presented in Tab.~\ref{tab:effi}. All measurements were conducted on a Linux workstation equipped with dual Intel Xeon Platinum 8488C processors (96 cores, 192 threads) and an NVIDIA RTX 4090 GPU.

For dense representations (regarding FDD \cite{FDD}), we maintain compact template sizes and achieve extraction and matching speeds comparable to those of other efficient fixed-length methods. Although FLARE uses four pose-enhancement combinations and consequently stores four templates per fingerprint, leading to a moderate increase in storage and a slight reduction in extraction and matching speed compared with FDD, its overall efficiency remains substantially higher than that of minutiae-based methods. FLARE therefore preserves the key advantage of fixed-length representations in large-scale deployments. Furthermore, when the templates are binarized, the storage requirement decreases significantly and the matching speed increases, with only a minor impact on recognition except for extremely low-quality fingerprints, as shown in Tab.~\ref{tab:compared_method}.

\begin{table}[!t]
  \centering
  \caption{Comparison of Storage Cost and Computational Efficiency.}
  \label{tab:effi}
  \resizebox{\linewidth}{!}{
  \begin{threeparttable}
  \begin{tabular}{lccccc
    }
    \toprule
    \textbf{Method}\tnote{$\dagger$} & 
    \makecell[c]{\textbf{Template}\tnote{$\ddagger$} \\ (KB)} & 
    \makecell[c]{\textbf{Preprocess} \\ (ms)} & 
    \makecell[c]{\textbf{Extraction} \\ (ms)} & 
    \makecell[c]{\textbf{Matching} \\ (s/pair)} &
    \makecell[c]{\textbf{Latency (1:1)}\tnote{$^*$} \\ (ms)}\\
    \midrule
    VeriFinger \cite{nist2020verifinger} & 7.051 & --- & 1995.21 & $2.43 \times 10^{-3}$ & 1997.64 \\
    DMD \cite{pan2024latent} & 182.947 & 1960.78 & 24.20 & $5.52 \times 10^{-4}$ & 1985.53 \\ 
    \hhline
    DeepPrint \cite{DeepPrint} & 0.384 & --- & 0.92 & $4.98 \times 10^{-9}$ & 0.92 \\
    AFRNet \cite{grosz2024afrnet} & 1.536 & --- & 0.53 & $1.02 \times 10^{-8}$ & 0.53 \\
    MultiScale \cite{gu2022latent} & 10.240 & 1.70 & 6.57 & $1.69 \times 10^{-6}$ & 8.27 \\
    \hhline
    FDD \cite{FDD} & 6.176 & 1.70 & 0.33 & $2.88 \times 10^{-8}$ & 2.03 \\
    FDD (binary) \cite{FDD} & 0.416 & 1.70 & 0.33 & $5.21 \times 10^{-9}$ & 2.03 \\
    FLARE & 24.704 & 3.47 & 1.30 & $4.15 \times 10^{-8}$ & 4.77 \\
    FLARE (binary) & 1.664 & 3.47 & 1.30 & $2.26 \times 10^{-8}$ & 4.77\\
    \bottomrule
  \end{tabular}
  \begin{tablenotes}
    \item[$\dagger$] Average template size, fingerprint preprocessing time, template extraction time, and matching time are measured on the NIST SD4 dataset for all  methods. Efficiency metrics are presented as the actual computational time: milliseconds (ms) per template for preprocessing and extraction, and seconds (s) per pair for matching.
    \item[$\ddagger$] All floating-point feature representations are stored in float16 format, except for VeriFinger, which uses its own proprietary format.
    \item[$^*$] The ``Latency (1:1)" column denotes the end-to-end latency for a single verification task (Preprocessing + Extraction + Matching).
  \end{tablenotes}
  \end{threeparttable}}
\end{table}

\section{Discussions} \label{sec:discussions}
\begin{figure}[!t]
  \centering
  \subfloat[]{\includegraphics[width=0.3\linewidth]{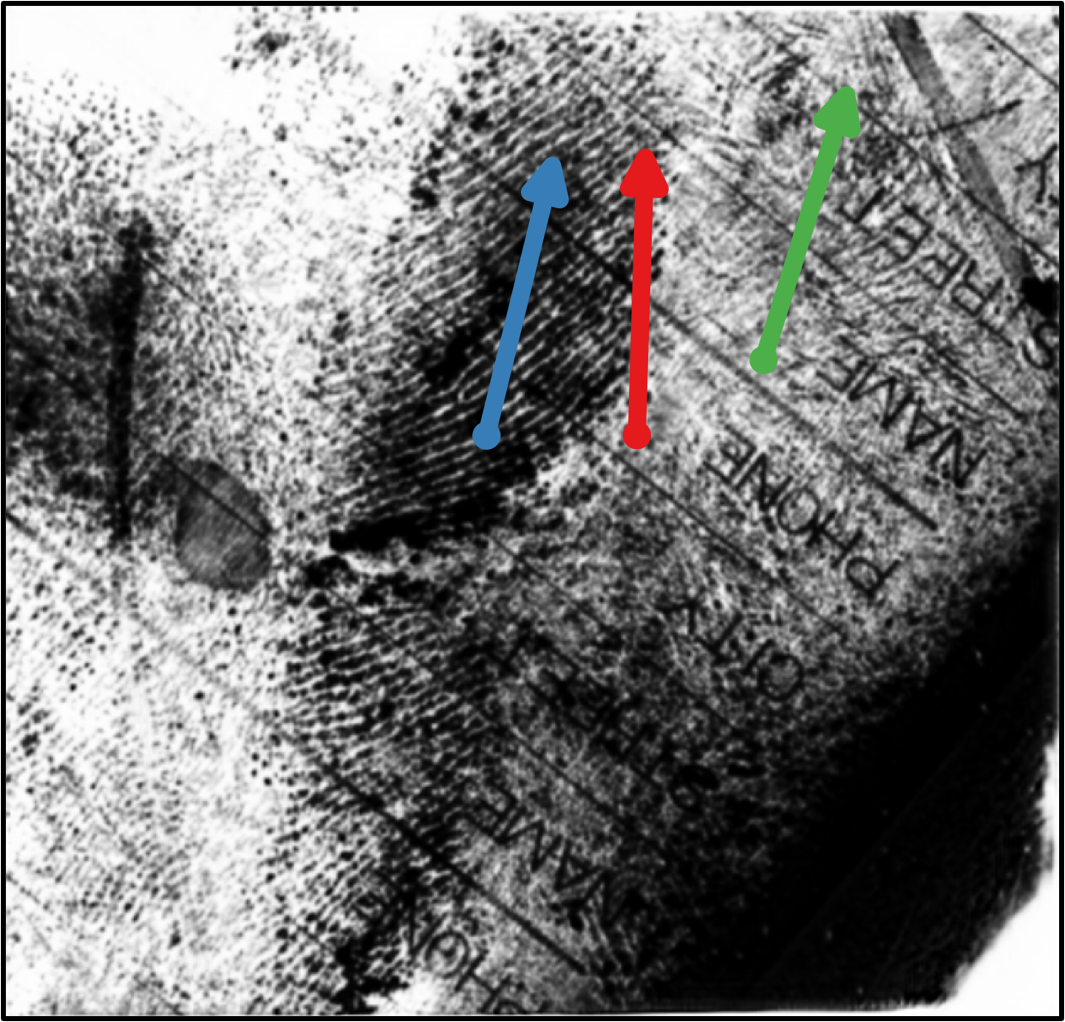}} \hfil
  \subfloat[]{\includegraphics[width=0.3\linewidth]{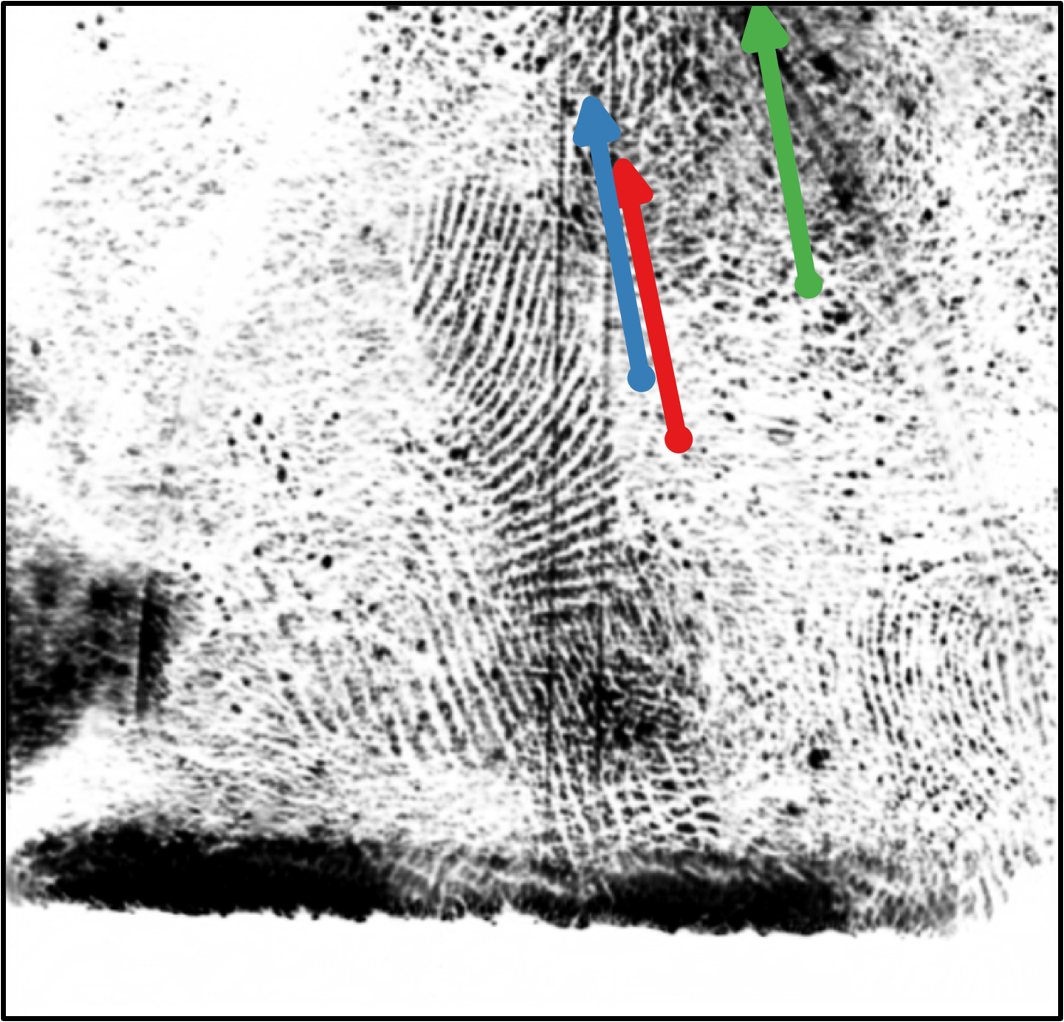}} \hfil
  \subfloat[]{\includegraphics[width=0.3\linewidth]{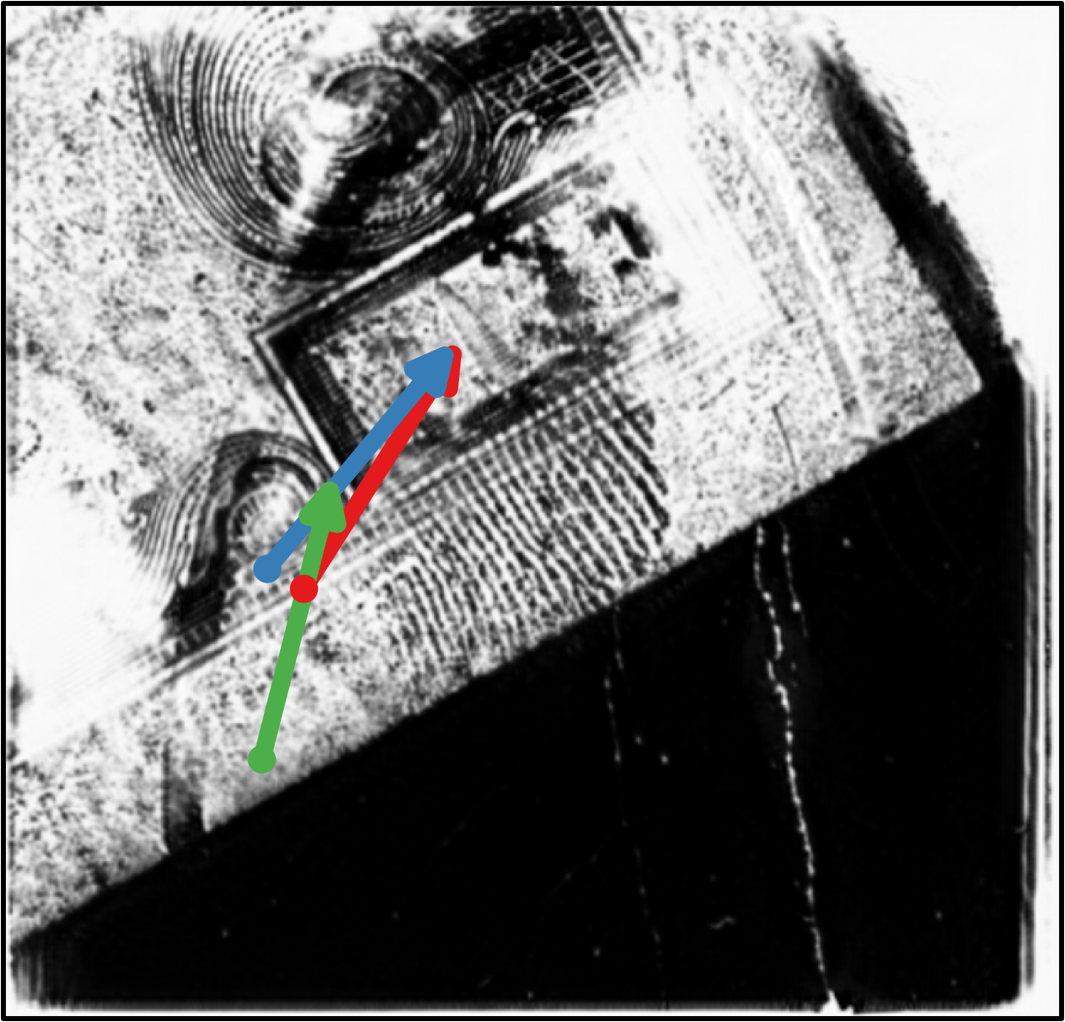}} \hfil
  \caption{Examples of erroneous pose estimation and enhancement with insufficient background suppression. Pose predictions are based on the original fingerprint images, and the orientation arrows follow the same definition as in Fig.~\ref{fig:pose_complement}.}
  \label{fig:failure_case}
\end{figure}

Although FLARE achieves state-of-the-art performance on multiple fingerprint datasets, there is still room for improvement. As shown in Fig.~\ref{fig:failure_case}, current pose estimation methods can fail when fingerprints are severely degraded or incomplete, such as in latent prints, leading to incorrect alignment and failed matching. While our enhancement modules improve ridge clarity without introducing false ridge patterns, they may sharpen background textures. This can lead to over-detection of spurious minutiae, as shown in Tab.~\ref{tab:enh_restoration}, and may interfere with pose estimation, particularly for local voting-based methods, when applied to enhanced images. Therefore, FLARE estimates pose from the original images. Future work will explore more robust pose estimation methods for damaged fingerprints and enhance the denoising ability of the enhancement network.

Regarding the dense descriptor, its spatial sensitivity significantly contributes to high matching accuracy once fingerprint alignment is achieved. However, in cases with minimal overlapping regions—such as partially captured fingerprints—this spatial rigidity may introduce matching ambiguity, potentially leading to incorrect identification when non-mated fingerprints share similar local patterns. To address this, we plan to enhance the fixed-length dense representation framework with overlap-aware strategies that take the extent of the overlapping area into account, improving robustness in more general cases. As discussed in Sec.~\ref{sec:effi}, FLARE retains the efficiency advantages of fixed-length representations, yet its current time and space complexity are not the lowest among fixed-length approaches due to the use of multiple pose-enhancement combinations. While this design offers clear accuracy advantages, future work will explore more lightweight preprocessing pipelines and further optimization of dense feature extraction and matching to improve runtime efficiency. We also plan to investigate compression and quantization techniques for dense representations to reduce storage requirements while keeping any potential performance degradation minimal.

\section{Conclusion}
In this paper, we propose FLARE, a fingerprint matching framework that integrates pose estimation, enhancement, and dense fixed-length descriptor extraction. FLARE standardizes fingerprint alignment through pose estimation, enhances ridge clarity while preserving structural fidelity, and extracts spatially aligned dense descriptors for robust matching. Beyond proposing an effective framework, this work systematically explores and evaluates the collaborative effects of combining multiple pose estimation and enhancement strategies on fixed-length fingerprint representations. Experimental results on various types of fingerprints demonstrate that FLARE consistently achieves strong matching performance. The proposed enhancement modules, UNetEnh and PriorEnh, further improve matching accuracy, particularly on low-quality and contactless fingerprints. Moreover, the integration of multiple pose estimation and enhancement methods is shown to effectively improve matching performance across diverse fingerprint conditions. Overall, FLARE establishes a unified and scalable solution for fixed-length fingerprint representation and matching across diverse conditions, providing a practical foundation for reliable and accurate fingerprint recognition in real-world applications.












\bibliographystyle{IEEEtran}
\bibliography{IEEEabrv,refs}

\clearpage
\setcounter{page}{1}
\twocolumn[
  \begin{@twocolumnfalse}
    \begin{center}
      \textbf{\large Supplementary Material}
    \end{center}
  \end{@twocolumnfalse}
]

\section{Supplementary Implementation Details for Pose and Enhancement Augmentation}
In the main paper, we describe a background-pattern augmentation strategy used during the training of the fingerprint pose estimation module, in which background patterns from the MSRA-TD500 dataset \cite{MSRA_TD500} are randomly inserted into fingerprint images. This augmentation is designed to improve the robustness of pose estimation when applied to latent fingerprints that contain strong background interference. Tab.~\ref{tab:bg_aug} provides the corresponding quantitative results. Using FDD \cite{FDD} as the baseline, we train its pose estimation module under different augmentation settings. The results show that models trained with background-pattern augmentation consistently achieve better descriptor-level matching performance than those trained without it, with particularly notable improvements on latent fingerprint datasets.

To simulate background interference during pose estimation training, a texture image randomly selected from the MSRA-TD500 dataset is blended with the input fingerprint. The background image is first randomly scaled (random from $0.5\times$ to $2\times$), rotated (betweem $-180\degree$ and $180\degree$), and spatially shifted by up to 50\% of its original dimensions, ensuring diverse and realistic background conditions before blending. Let $\omega \in [0.2, 0.8]$ denote a randomly sampled mixing coefficient. The augmented fingerprint $\hat{f}$ is generated as $\hat{f}=\omega b + (1-\omega) f$, where $f$ is the original fingerprint image, and $b$ is the transformed background pattern.

To train the enhancement module, we simulate low-quality fingerprints using a set of handcrafted degradations that reflect common distortions in practical acquisition. Ridge degradations are generated by applying Gaussian blur (kernel size 2-6, $\sigma$ 1-5), morphological dilation (2$\times$2) to mimic dry-finger artifacts, and morphological erosion (3$\times$3) to emulate moist or heavy-pressing effects. External interference is synthesized by inserting background patterns, where grayscale comparison ensures that ridge regions remain dominant, overlaying 1-5 randomly scaled handwritten digits at arbitrary locations, and adding line disturbances composed of 4-10 straight lines (width 2-5) and 3-7 curved lines (width 3-10) with random curvature and orientation.

\begin{table*}[!t]
  \centering
  \caption{Matching Performance (\%) of FDD With and Without Background-Pattern Augmentation for Training the Pose Estimation Module. Unless otherwise specified, TAR@FAR = 0.1\% is reported.}
  \label{tab:bg_aug}
  \renewcommand{\arraystretch}{1.2}
  \begin{threeparttable}
    \resizebox{\textwidth}{!}{
      \begin{tabular}{l
        c*{2}{>{\centering\arraybackslash}p{0.05\linewidth}}
        c*{1}{>{\centering\arraybackslash}p{0.05\linewidth}}
        c*{1}{>{\centering\arraybackslash}p{0.05\linewidth}}
        c*{4}{>{\centering\arraybackslash}p{0.05\linewidth}}
        c*{4}{>{\centering\arraybackslash}p{0.05\linewidth}}}
        \toprule
        \multirow{2}{*}{\textbf{\makecell[l]{Enh.\\Method}}} & \multicolumn{2}{c}{\textbf{NIST SD4}} & \multicolumn{2}{c}{\textbf{N2N Plain}} & \multicolumn{1}{c}{\textbf{FVC02}} & \multicolumn{1}{c}{\textbf{FVC04}} & \multicolumn{1}{c}{\textbf{FVC06}} & \multicolumn{1}{c}{\textbf{PolyU}} & \multicolumn{2}{c}{\textbf{NIST SD27}} & \multicolumn{2}{c}{\textbf{THU Latent10K}} \\
        \cmidrule(lr){2-3} \cmidrule(lr){4-5} \cmidrule(lr){6-9} \cmidrule(lr){10-11} \cmidrule(lr){12-13}  
        & \textbf{Rank-1} & \textbf{TAR}\tnote{$\dagger$} & \textbf{Rank-1} & \textbf{TAR}\tnote{$\dagger$} & \multicolumn{4}{c}{\textbf{TAR}} & \textbf{Rank-1} & \textbf{TAR} & \textbf{Rank-1} & \textbf{TAR} \\
        \midrule
        w/o bkg & 99.55 & 99.40 & 98.30 & 98.20 & 94.00 & 99.46 & 88.86 & 88.48 & 41.86 & 46.12 & 77.25 & 85.39 \\
        w/ bkg & 99.75 & 99.60 & 98.65 & 98.75 & 95.86 & 99.50 & 89.62 & 88.62 & 51.94 & 62.02 & 82.46 & 90.79 \\
        \bottomrule 
      \end{tabular}}
      \begin{tablenotes}
        \item[$\dagger$] TAR@FAR=0.01\%.
      \end{tablenotes}
  \end{threeparttable}  
\end{table*}

\section{Detailed network structure and hyperparameter settings for PriorEnh and FDRN}
Tab. \ref{tab:prior_hyper} presents the hyperparameter settings for PriorEnh. Both $E_\text{H}$ in the first stage and $E_\text{L}$ in the second stage share the same architecture, while $D_\text{H}$ corresponds to the symmetric UNet-based upsampling architecture. The detailed structure of FDRN is shown in Tab. \ref{tab:fdrn_Structure}.


\section{Experiments}

\begin{figure}[!t]
  \centering
  \subfloat[Effect of Translation\label{fig:fdd_trans_rob}]{\includegraphics[width=0.45\linewidth]{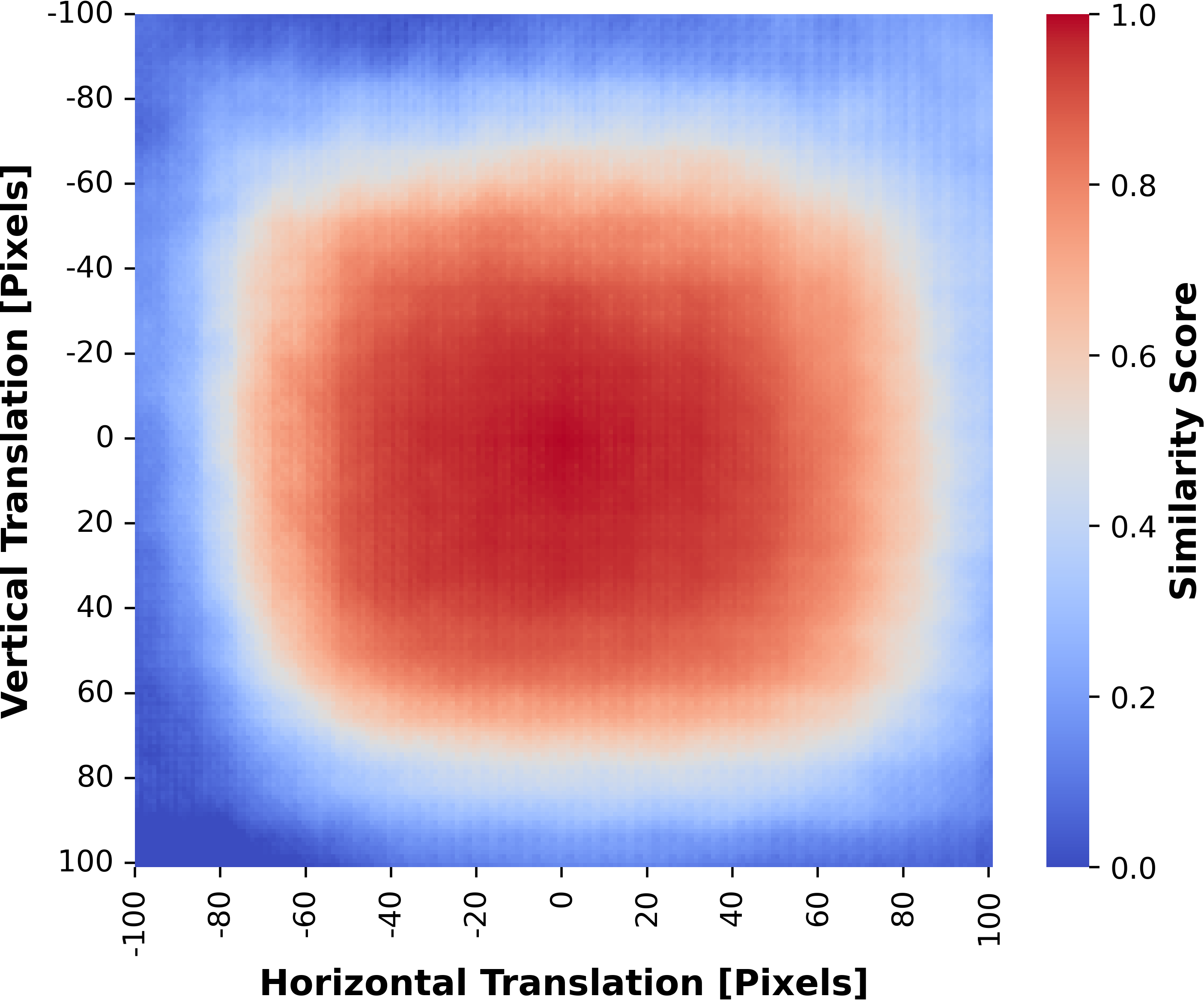}} \hfil
  \subfloat[Effect of Rotation\label{fig:fdd_rot_rob}]{\includegraphics[width=0.45\linewidth]{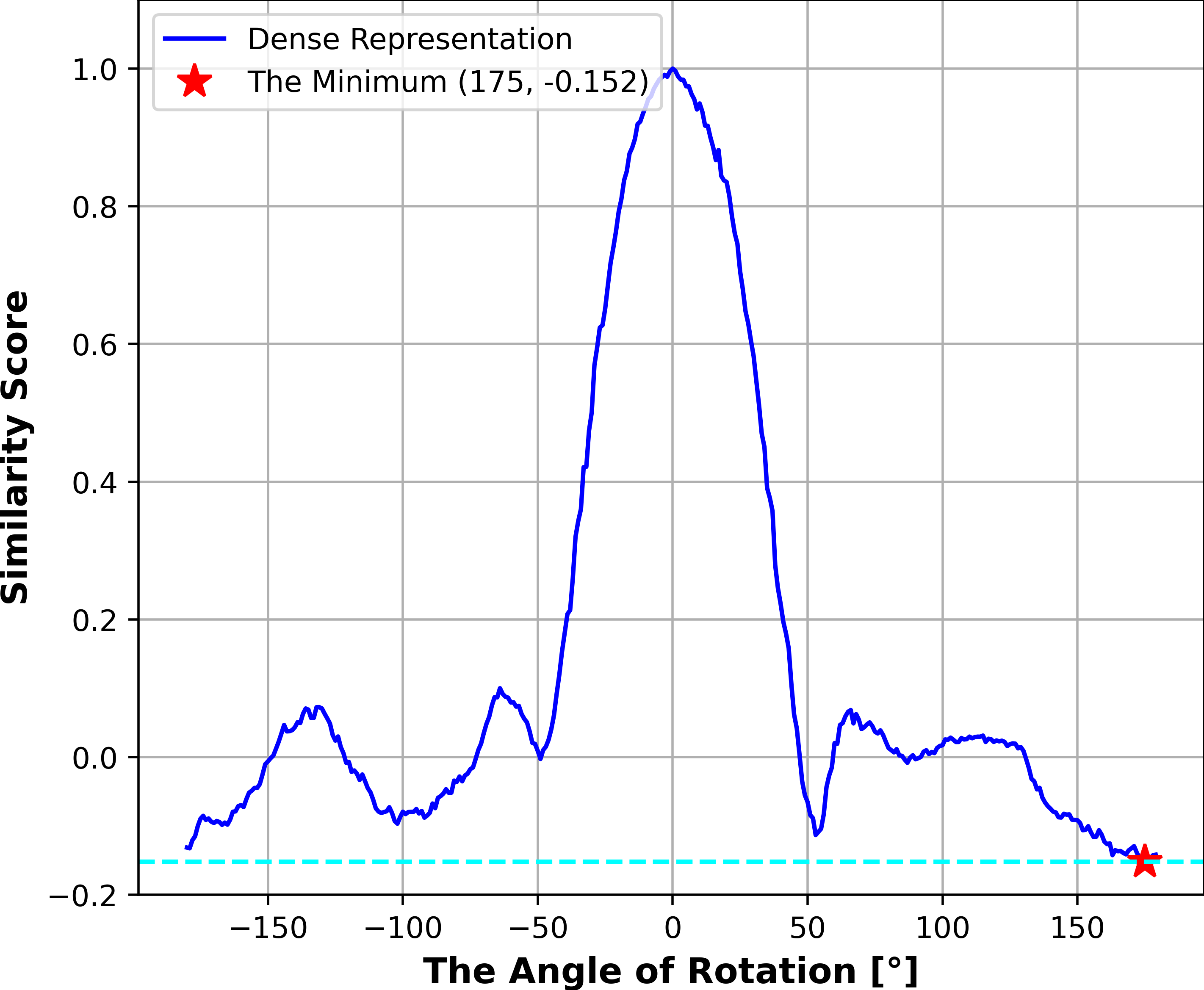}} \hfil
  \caption{The effect of pose variations on dense representations.}
  \label{fig:fdd_pose_robust}
\end{figure}

\begin{figure}[!t]
  \subfloat[FLARE]{\includegraphics[width=0.5\linewidth]{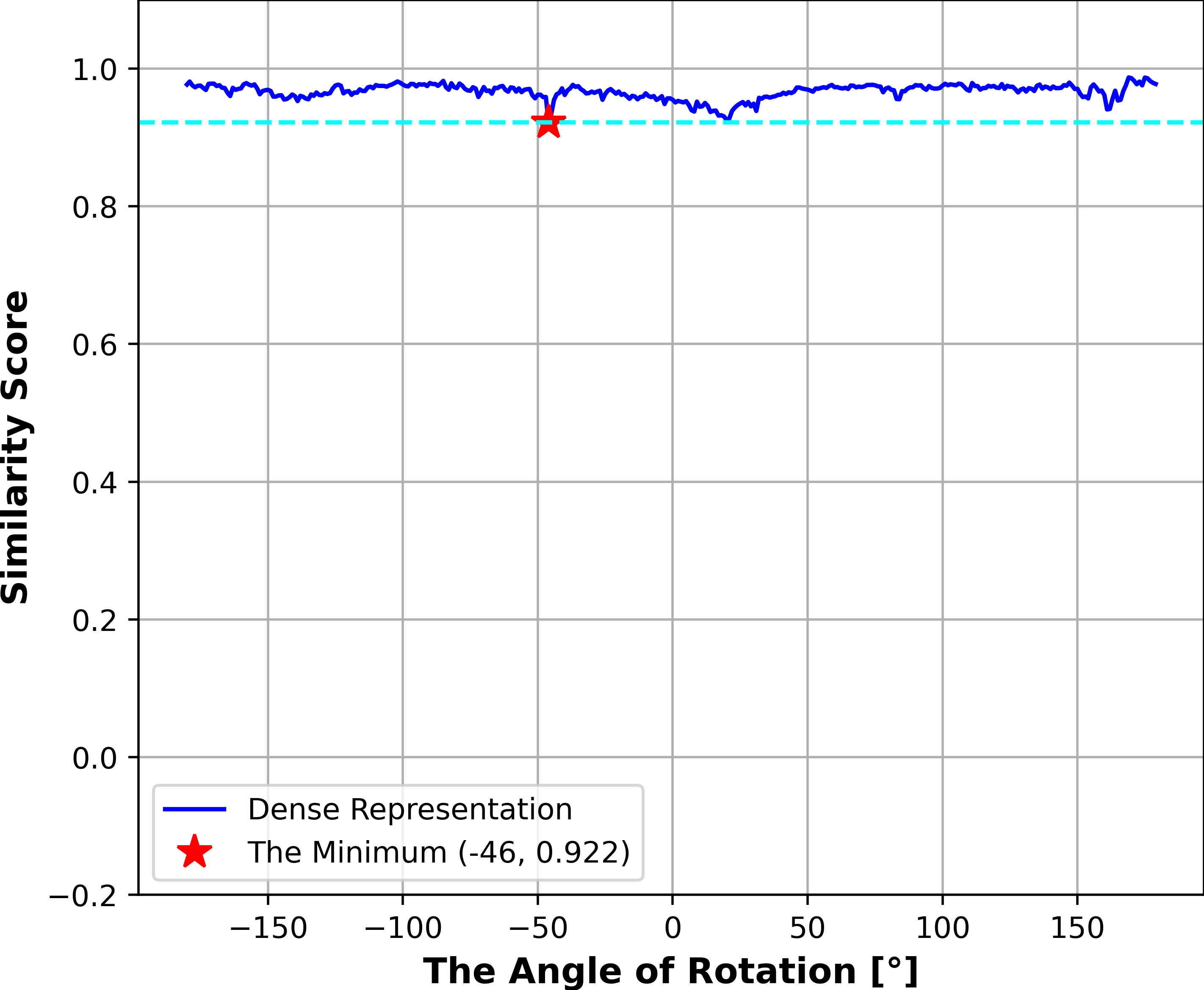}} \hfil
  \subfloat[DeepPrint\cite{DeepPrint}]{\includegraphics[width=0.5\linewidth]{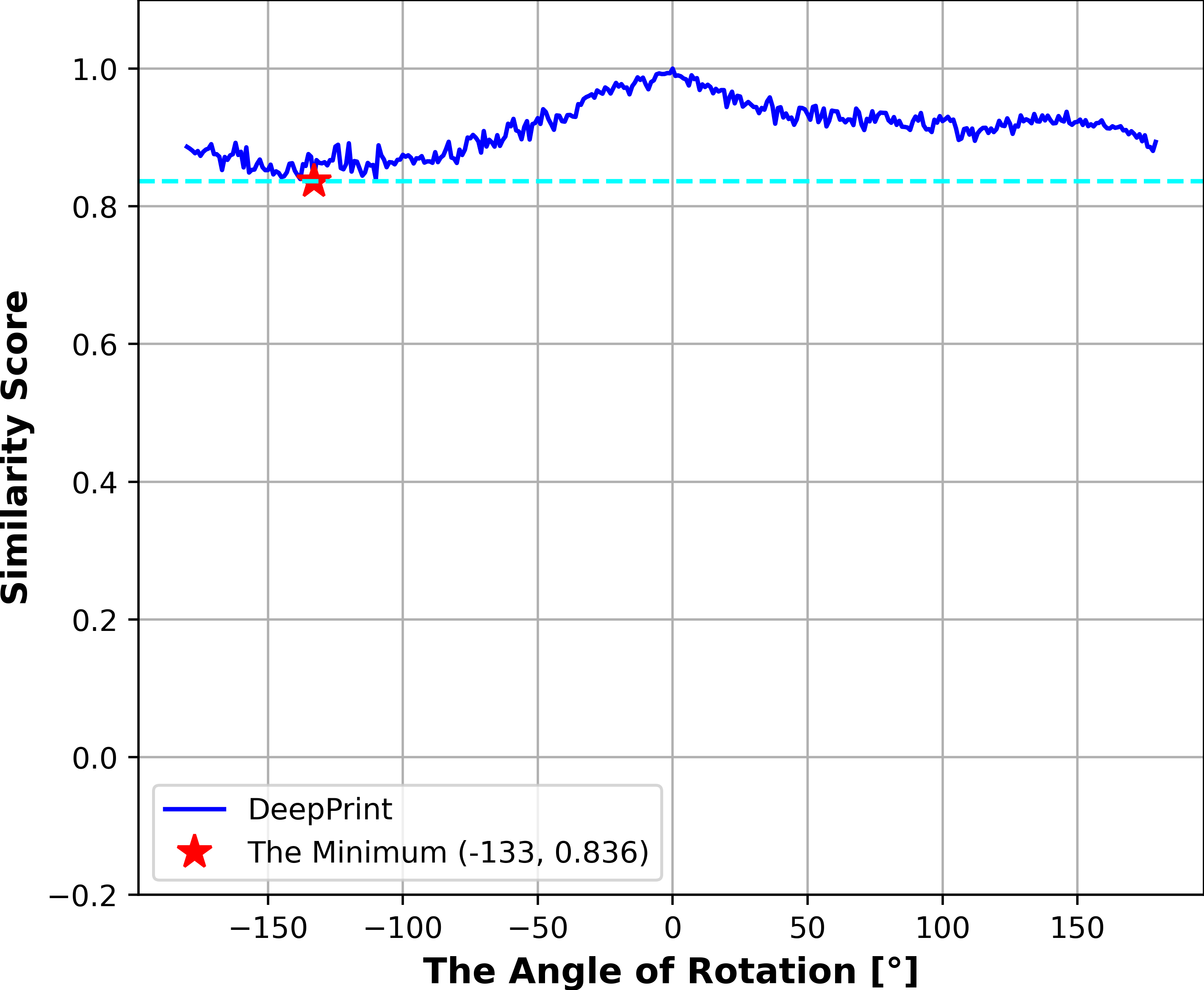}} \hfil
  \caption{The effect of rotation variations on (a) FLARE and (b) DeepPrint\cite{DeepPrint}.}
  \label{fig:pose_robust_rotate}
\end{figure}

\begin{figure}[!t] 
  \subfloat[FLARE]{\includegraphics[width=0.48\linewidth]{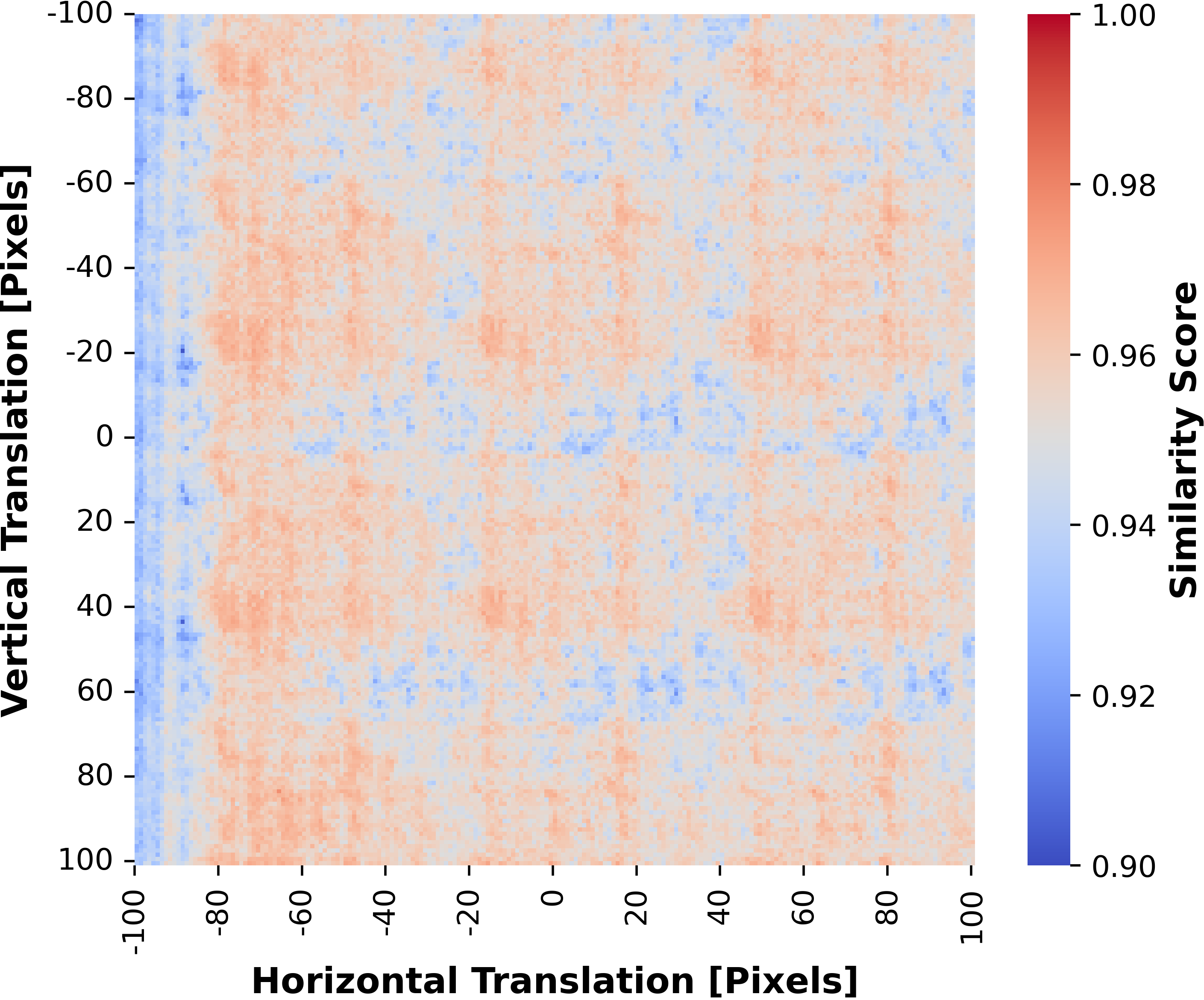}} \hfil
  \subfloat[DeepPrint\cite{DeepPrint}]{\includegraphics[width=0.48\linewidth]{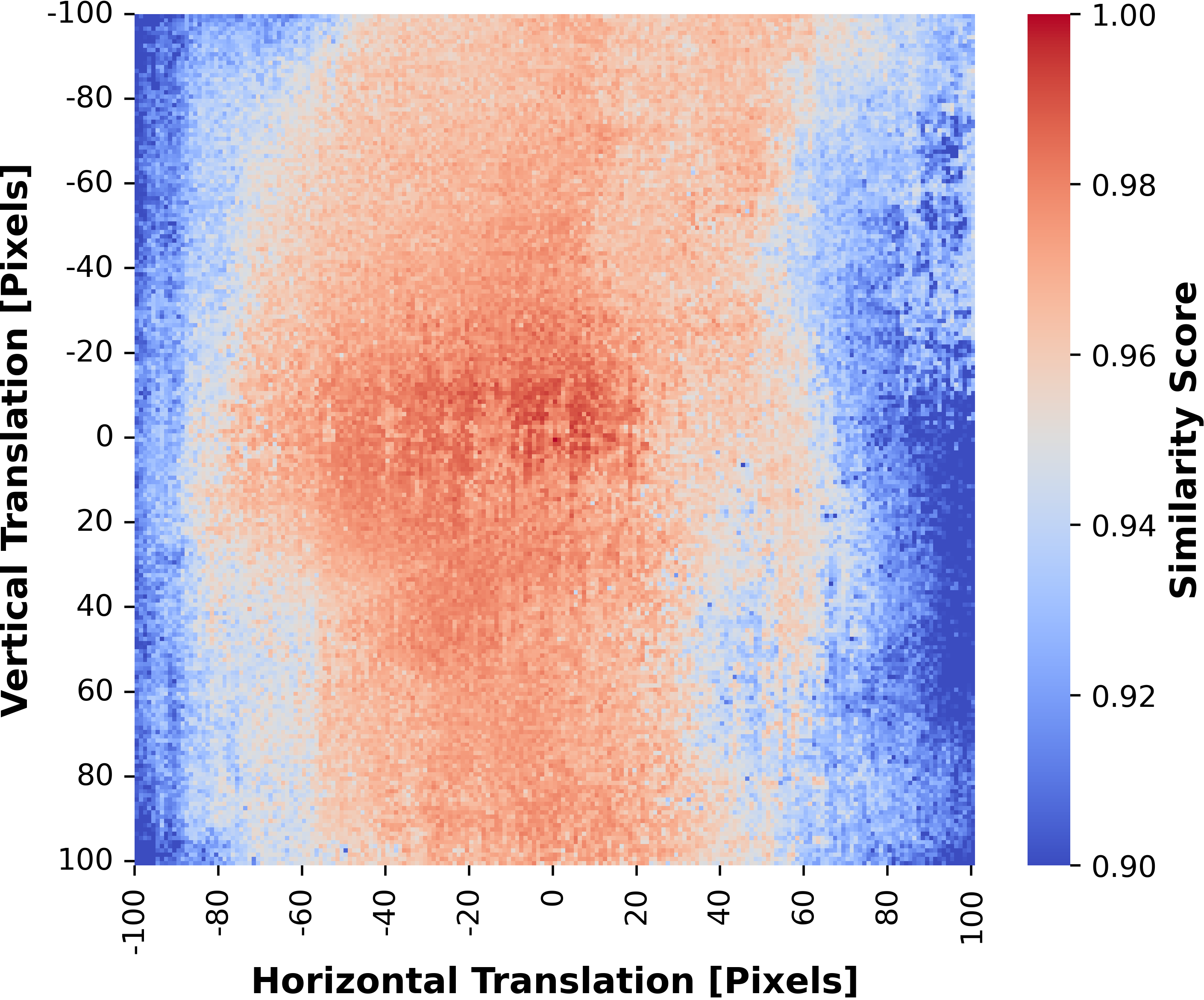}} \hfil
  \caption{The effect of translation variations on (a) FLARE and (b) DeepPrint\cite{DeepPrint}.}
  \label{fig:pose_robust_trans}
\end{figure}

\subsection{Robustness to 2D Pose Variations} 
Variations in the 2D pose of fingerprint images can affect descriptor extraction and thus impact matching accuracy. Since our dense descriptor is defined in a canonical fingerprint coordinate system with embedded spatial encoding, it is inherently sensitive to large pose deviations. To evaluate this, we select a low-quality, limited-area example from the N2N Plain dataset (left image of Fig.~\ref{fig:n2nplain_ex}) and apply controlled translation and rotation. As shown in Fig.~\ref{fig:fdd_trans_rob}, the descriptor exhibits moderate robustness to translation—likely due to the inherent shift tolerance of CNNs—but suffers a sharp performance drop under rotation (Fig.~\ref{fig:fdd_rot_rob}). These results underscore the need for accurate pose alignment, which we address using precise pose estimation strategies in our framework.

In FLARE, pose robustness is achieved through explicit alignment using a pretrained pose estimation module, whereas methods like DeepPrint \cite{DeepPrint} rely on data augmentation to improve the robustness of descriptor extraction during training. To compare these strategies, we evaluate FLARE using the pose estimation method of Guan et al. \cite{guan2025robust} together with the original FDD \cite{FDD}, and compare it against DeepPrint \cite{DeepPrint}. For a fair comparison, both models are trained with the same extensive pose augmentation settings, including random rotations in $[-180^\circ, 180^\circ]$ and translations of up to $\pm100$ pixels. Notably, the FDD model remains unchanged and does not benefit from augmented training. As shown in Fig.~\ref{fig:pose_robust_rotate}, FLARE maintains high descriptor similarity scores across a wide range of rotations ($-180^\circ$ to $180^\circ$), while DeepPrint shows a clear performance drop as the rotation angle increases. A similar trend is observed in Fig.~\ref{fig:pose_robust_trans}, where FLARE demonstrates stronger translation robustness compared to DeepPrint. These results demonstrate that explicit pose correction leads to greater robustness than relying on large pose augmentation during descriptor training. Considering that extreme pose variations are uncommon in practical applications, all other experiments in this paper adopt the augmentation strategy described in Sec.~\ref{sec:implementation}.

\subsection{More Evaluated Results on Fingeprint Matchings}
\begin{figure*}[!t]
  \centering
  \subfloat[NIST SD4]{\includegraphics[width=.45\linewidth]{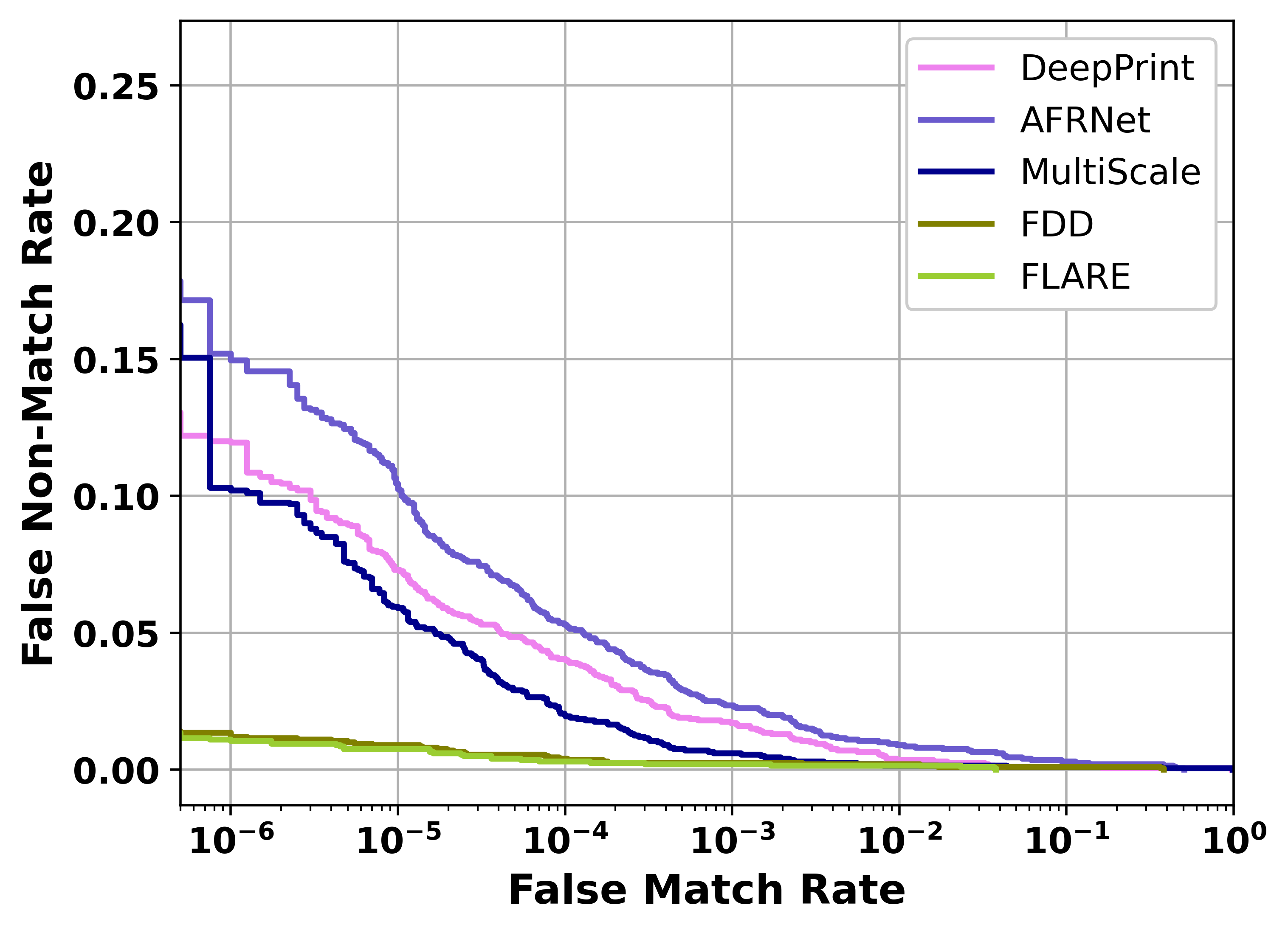}} \hfil
  \subfloat[N2N Plain]{\includegraphics[width=.45\linewidth]{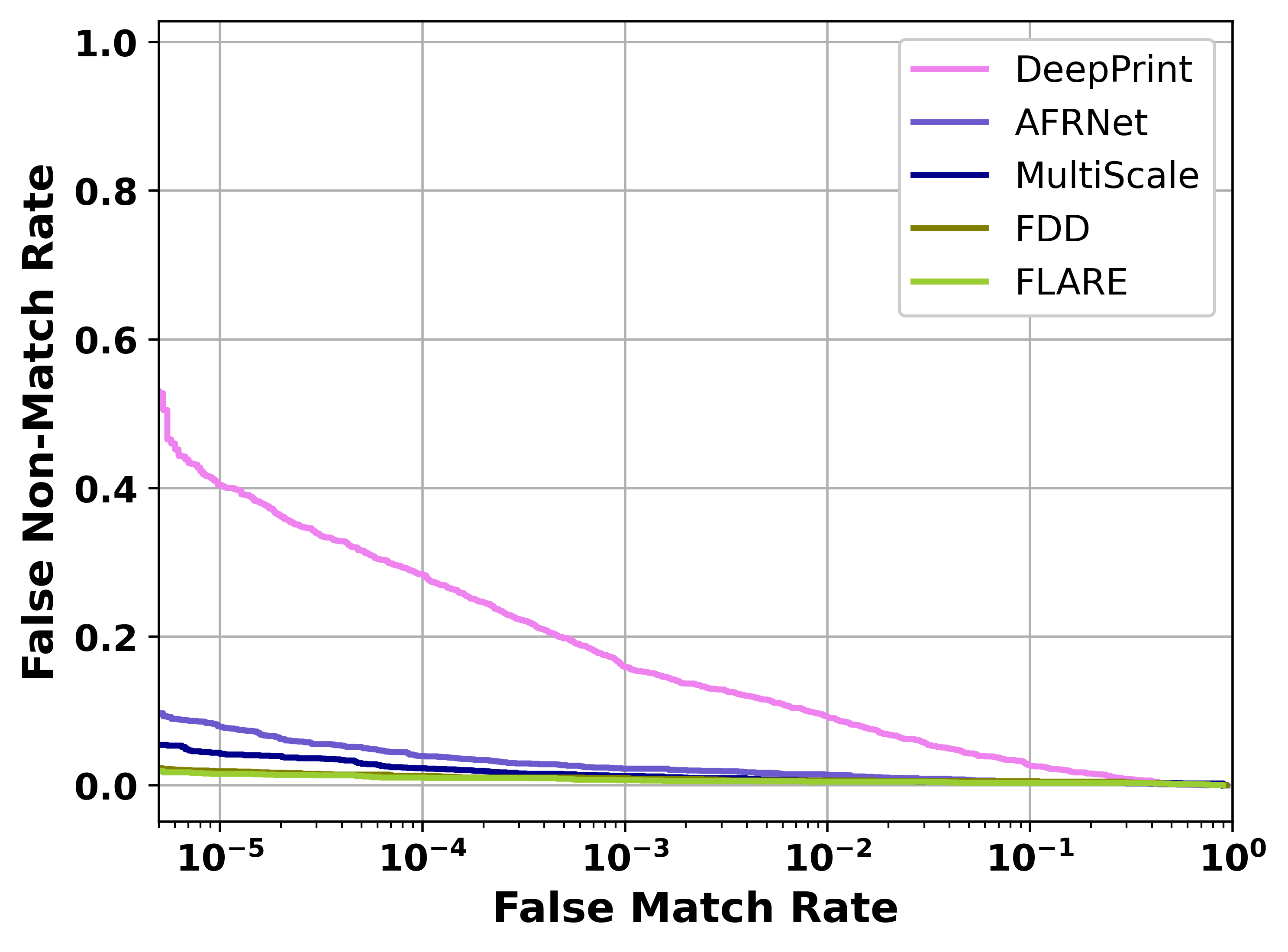}} \hfil
  \subfloat[FVC2002 DB3A]{\includegraphics[width=.3\linewidth]{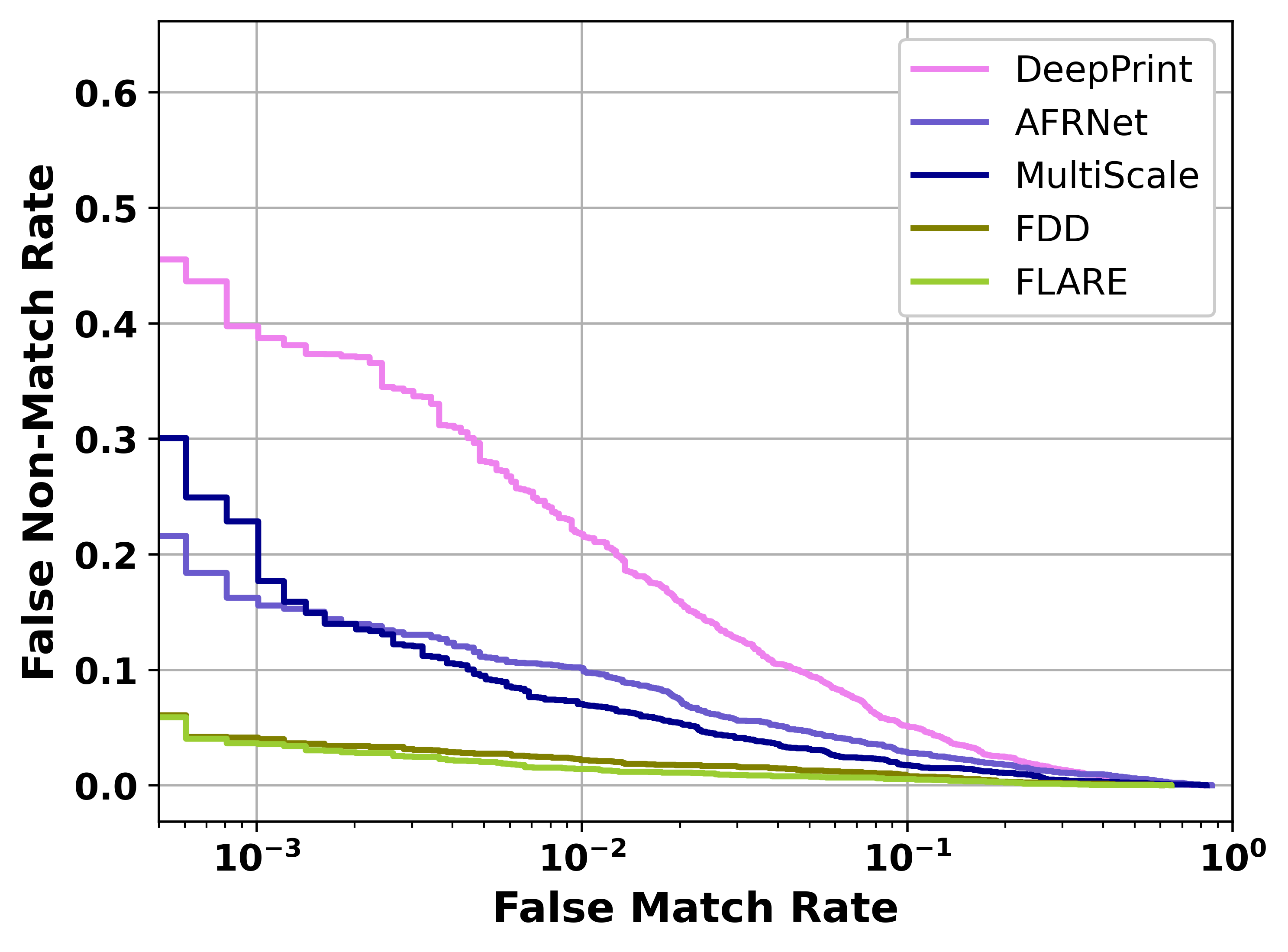}} \hfil
  \subfloat[FVC2004 DB1A]{\includegraphics[width=.3\linewidth]{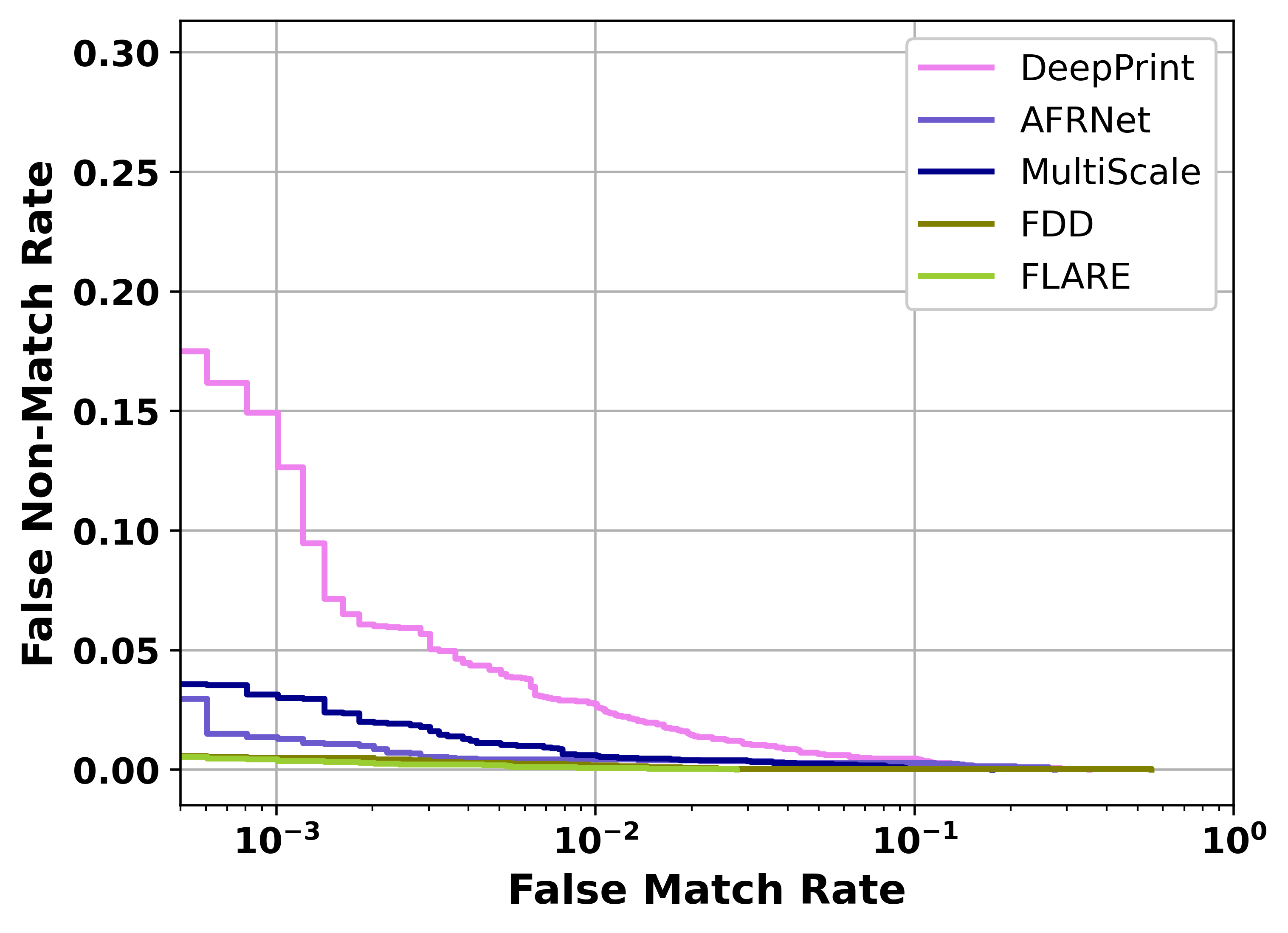}} \hfil
  \subfloat[FVC2006 DB1A]{\includegraphics[width=.3\linewidth]{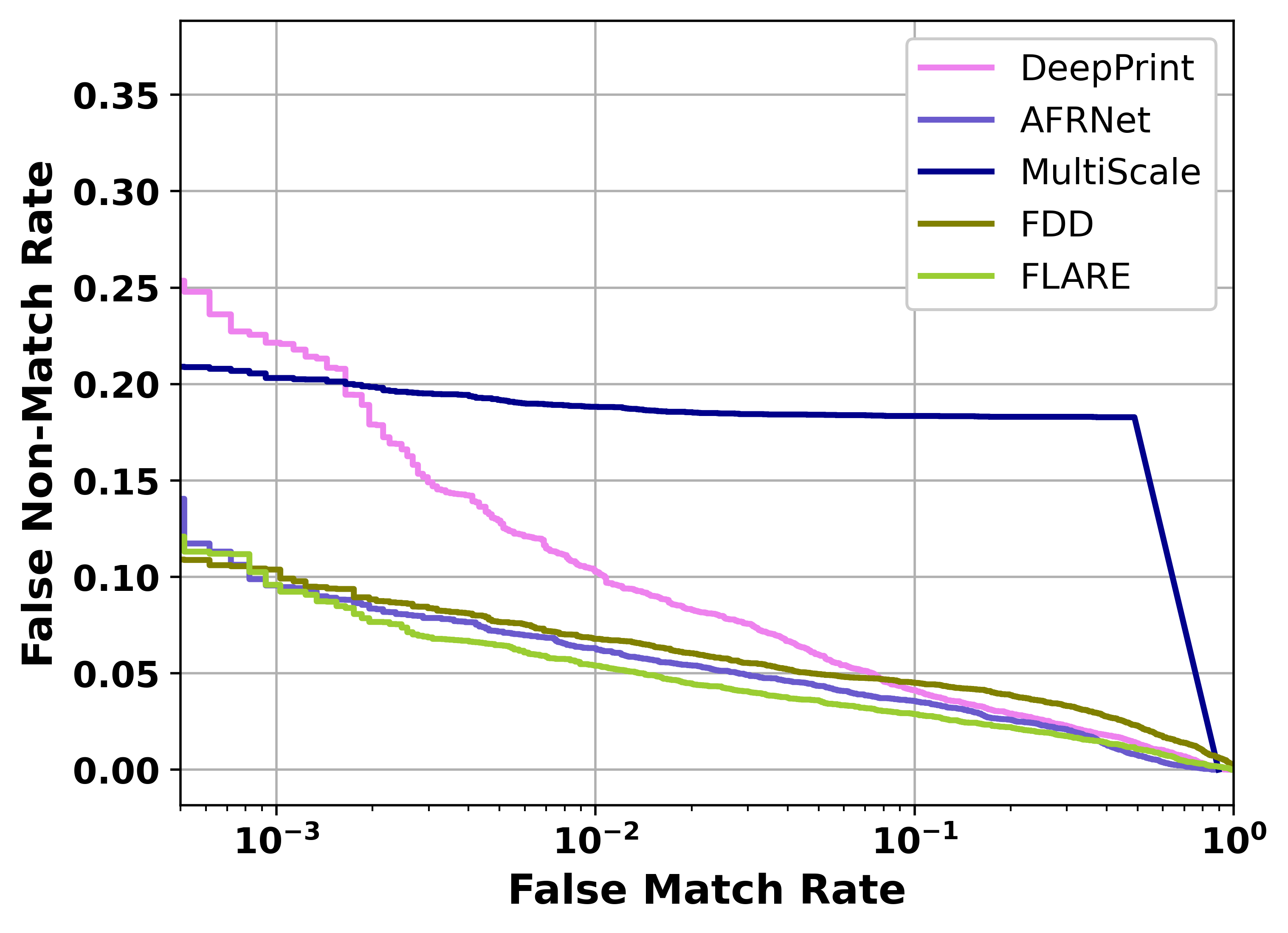}} \hfil
  \caption{Open-set identification performance of all methods, presented with DET curves for (a) rolled fingerprints matching on NIST SD4, (b, d) plain fingerprints matching on N2N Plain and FVC2004 DB1A, and (c, e) partial fingerprints matching on FVC2002 DB3A and FVC2006 DB1A, respectively.}
  \label{fig:more_det}
\end{figure*}

\begin{figure*}[!t]
  \centering
  \subfloat[\label{fig:fdd_ctl1}]{\includegraphics[width=.33\linewidth]{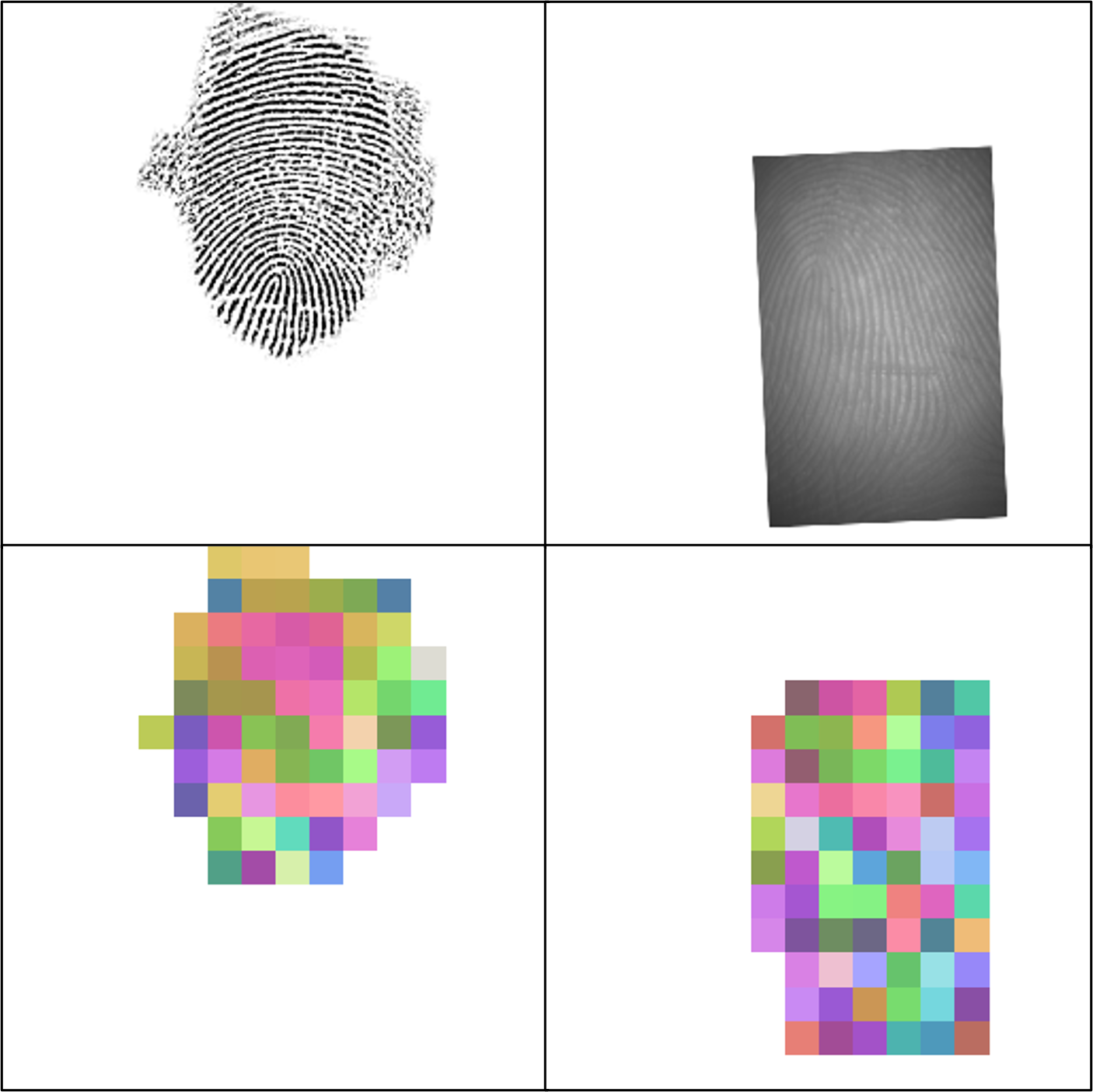}} \hfil
  \subfloat[\label{fig:fdd_ctl3}]{\includegraphics[width=.33\linewidth]{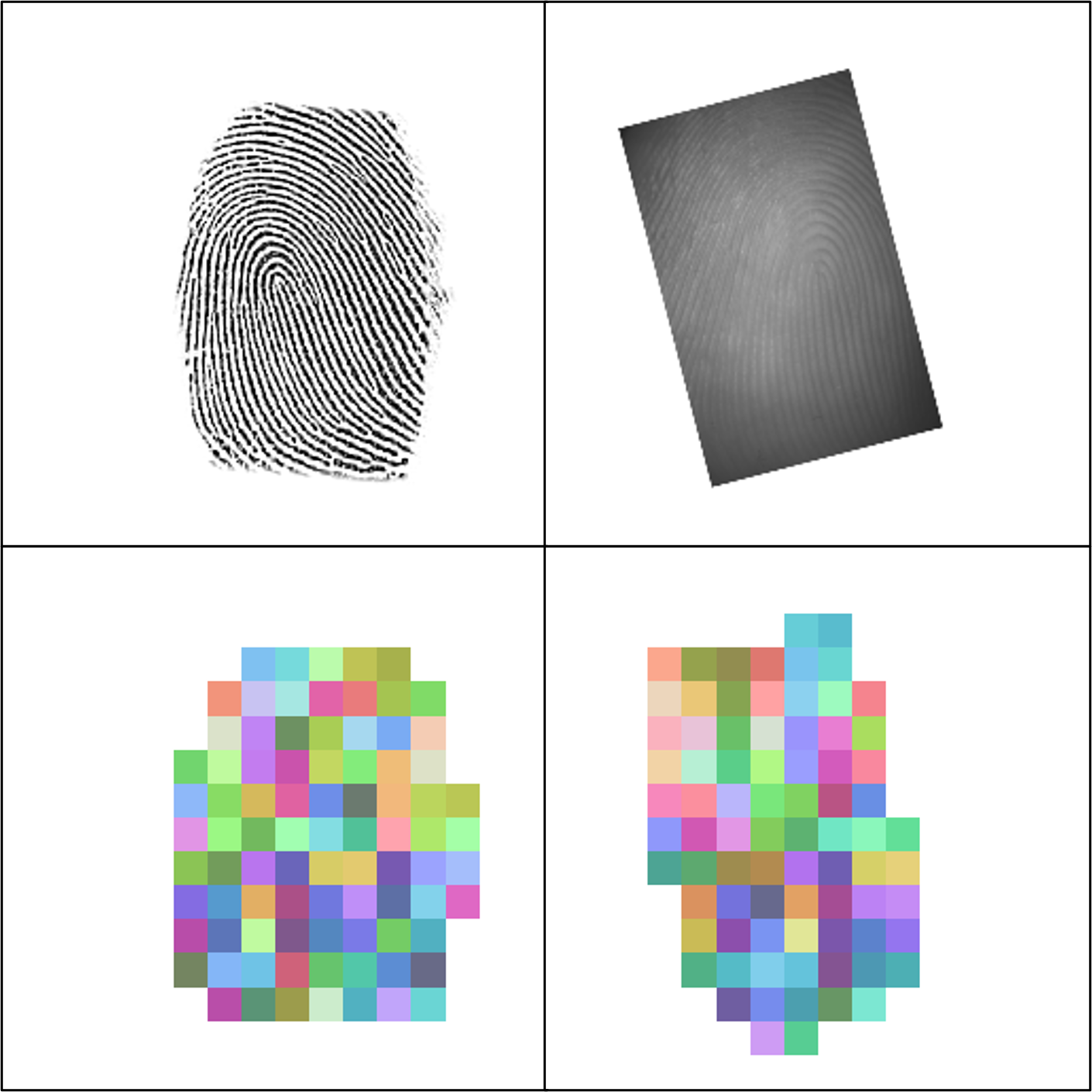}} \hfil
  \subfloat[\label{fig:fdd_latent3}]{\includegraphics[width=.33\linewidth]{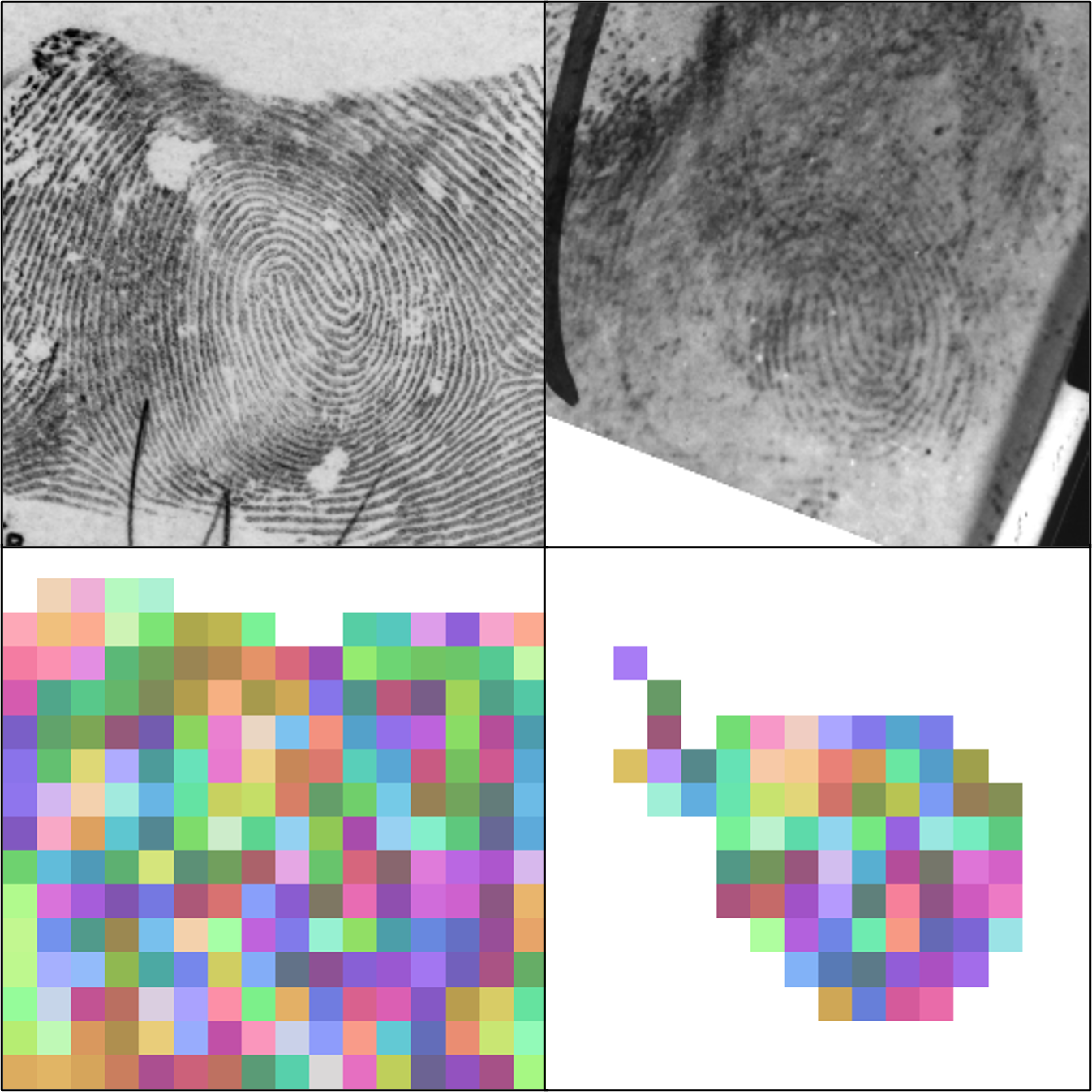}} \hfil
  \subfloat[\label{fig:fdd_latent4}]{\includegraphics[width=.33\linewidth]{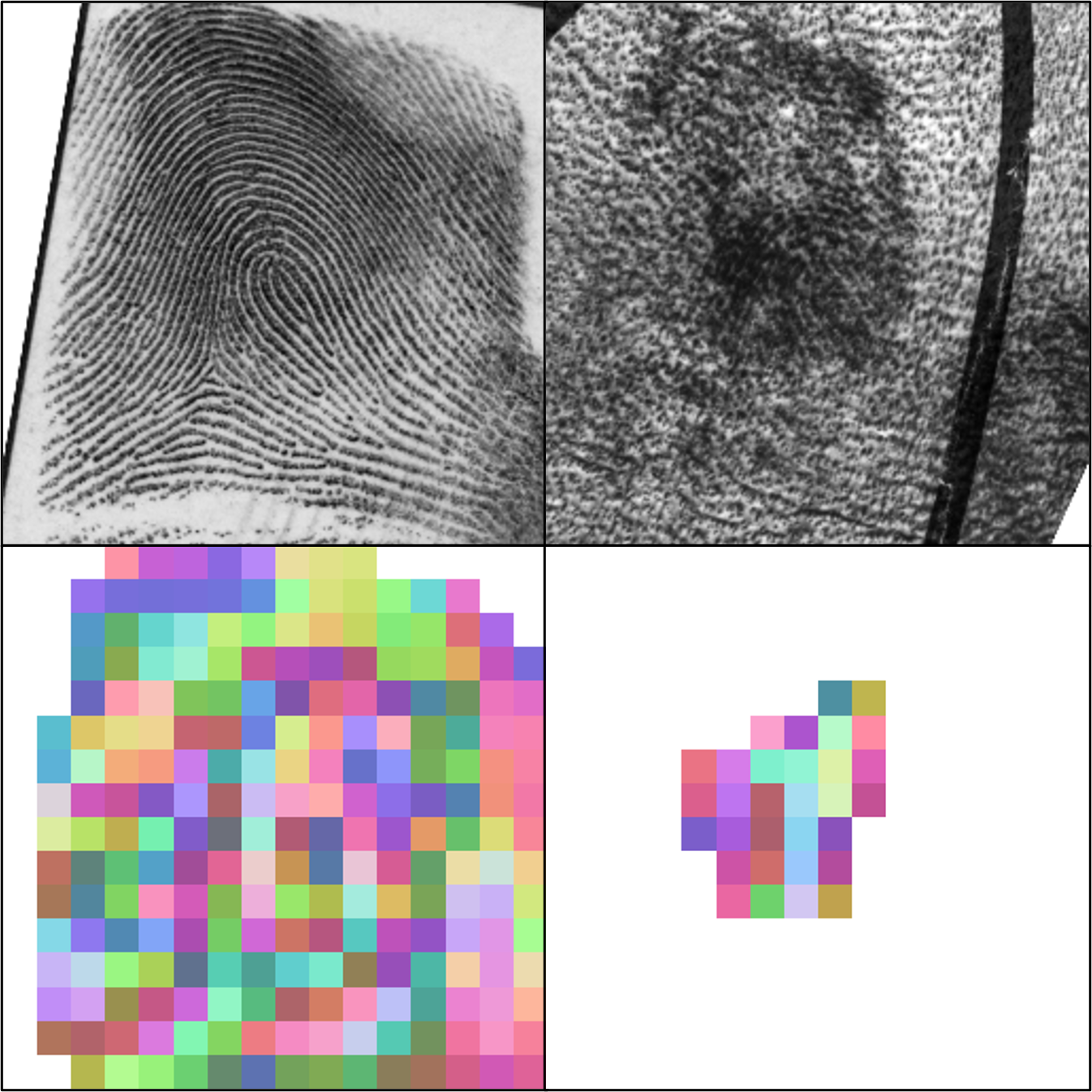}} \hfil
  \subfloat[\label{fig:fdd_latent5}]{\includegraphics[width=.33\linewidth]{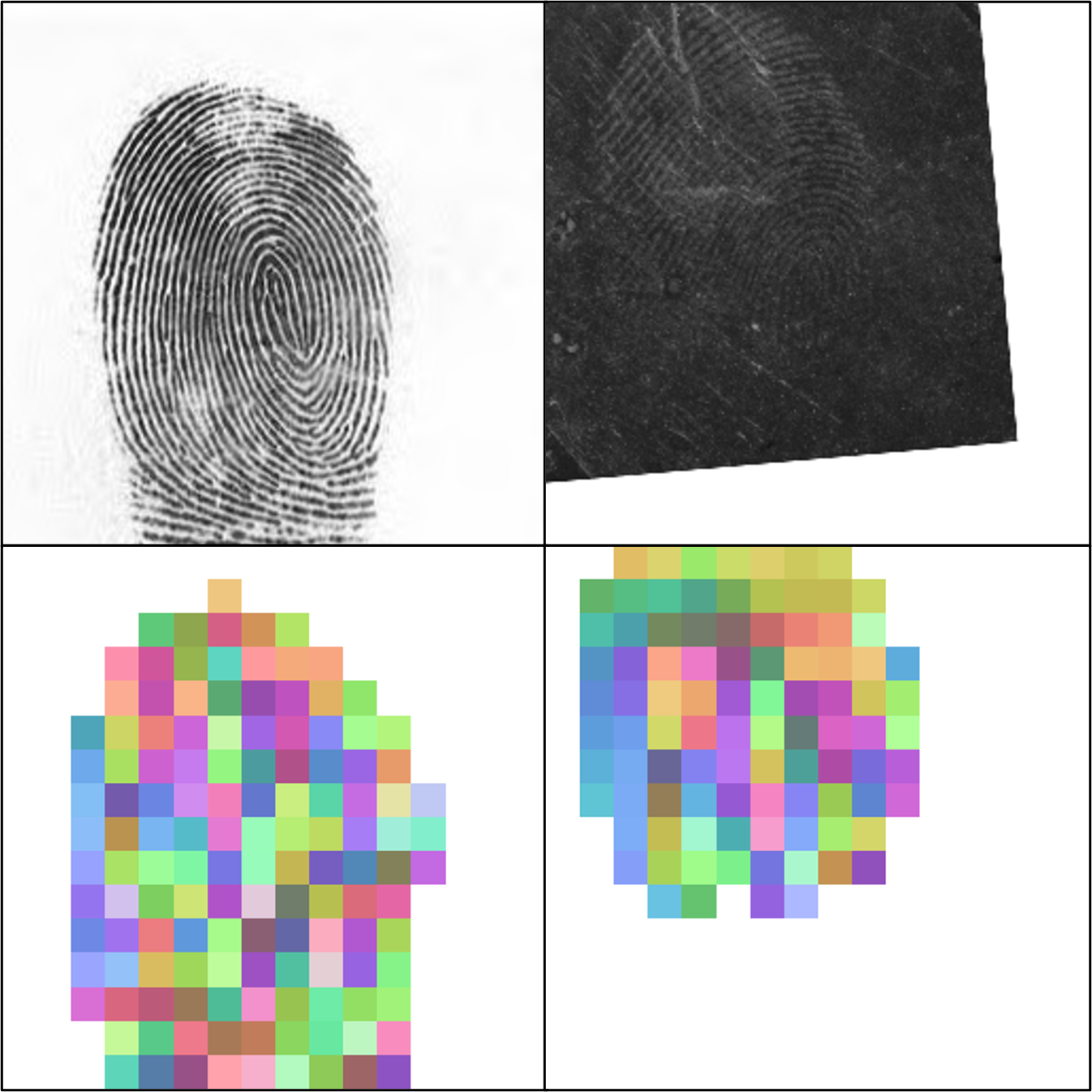}} \hfil
  \subfloat[\label{fig:fdd_latent6}]{\includegraphics[width=.33\linewidth]{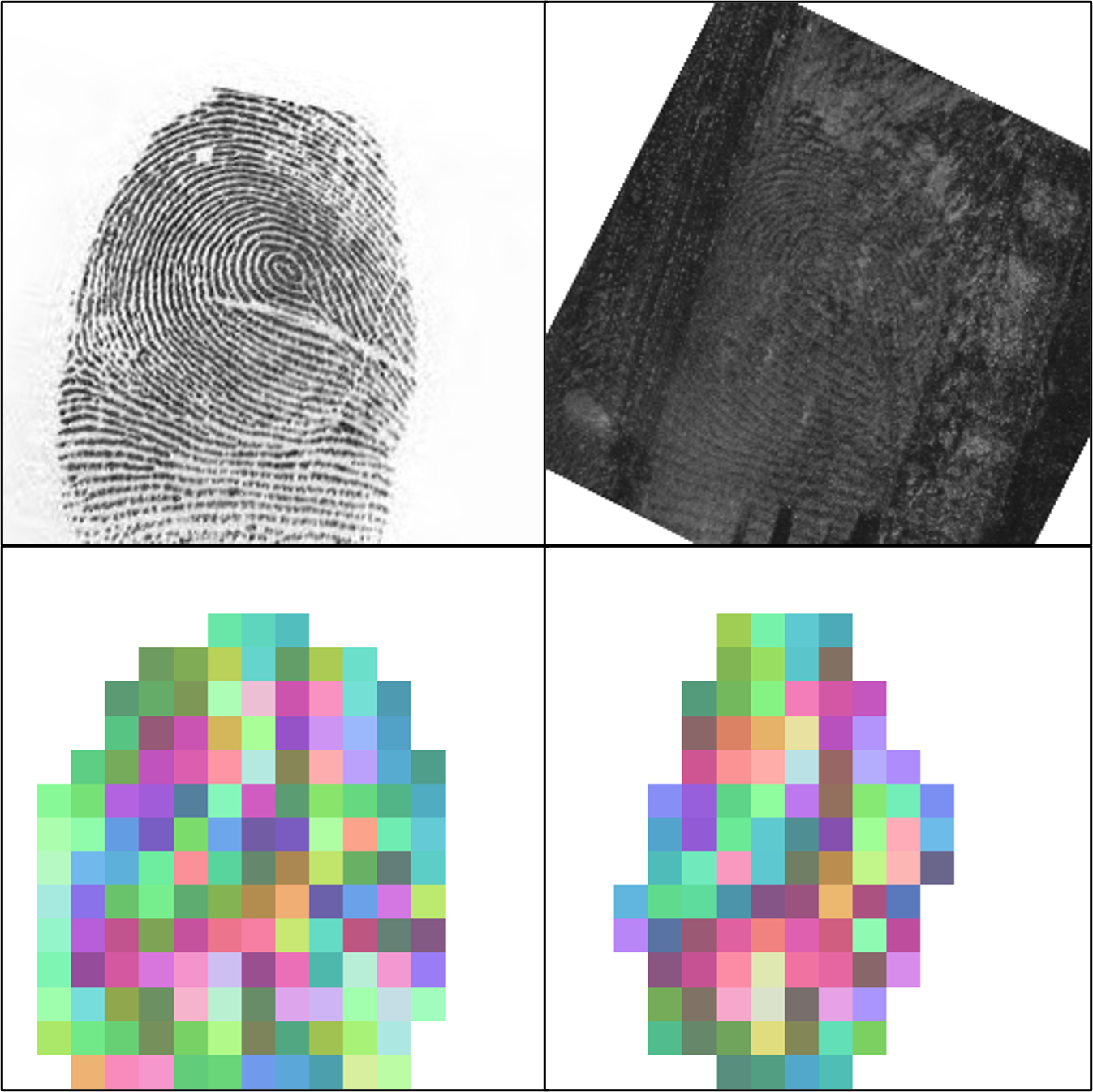}} \hfil
  \caption{Examples of fixed-length dense representations extracted from genuine pairs of additional contactless and latent fingerprint samples. The fingerprint images shown in the figure have been aligned based on their estimated poses.}
  \label{fig:fdd_illus_suppl}
\end{figure*}

Fig.~\ref{fig:more_det} presents additional DET curves comparing all methods on rolled, plain, and partial fingerprint matching tasks. The results consistently demonstrate that FLARE outperforms the other methods across nearly all evaluated datasets. And Fig.~\ref{fig:fdd_illus_suppl}, provided as a supplement to Fig.~\ref{fig:fdd_illus}, offers additional examples that further illustrate the behavior of the dense representation in more challenging scenarios.

\subsection{More Evaluated Results about Enhancement Methods}
In the main text, we evaluated the performance of the Fixed-Length Dense Descriptor (FDD) \cite{FDD} trained on original fingerprint images when applied to various enhanced images. Here, we further assess the performance of FDD models retrained using images enhanced by their corresponding methods. This is based on the assumption that both UNetEnh and PriorEnh perform enhancement in the original image domain, and thus should still use descriptor extraction networks trained on original images. As shown in Table~\ref{tab:enhancement_comp_retrained}, our proposed enhancement method demonstrates superior performance compared to the other enhancement approaches. When FDDs are retrained using images enhanced by their corresponding methods, performance improves on high-quality datasets such as NIST SD4, N2N Plain, and FVC2004 DB1A, compared to using the original FDD model trained solely on unenhanced images. However, on more challenging datasets—such as partial, latent, and contactless fingerprints—the retrained models often perform worse than those trained on original images. Furthermore, for all evaluated datasets, these enhancement-specific models \cite{nist2020verifinger,tang2017fingernet,FingerGAN} still fail to surpass the performance achieved by applying the original model directly to unenhanced images. In contrast, our proposed enhancement method consistently improves matching accuracy across both high-quality and challenging datasets.

\begin{table}
  \centering
  \caption{The hyperparameters for the PriorEnh.}
  \label{tab:prior_hyper}
  \begin{tabular}{lc}
    \toprule
    \textbf{Settings} & \textbf{PriorEnh $512\times512$}  \\
    \midrule
    Compression Ratio & 4\\
    Latent Feature Shape & $128 \times 128 \times 3$ \\
    Codebook Length $l_c$ & 4096 \\
    Channels & 64 \\
    Depth & 2 \\
    Channel Multiplier & 1,2,4 \\
    Attention resolutions & 16 \\
    GatedPixelCNN Embedding dim & 64 \\
    \bottomrule
  \end{tabular}
\end{table}

\begin{table*}[!t]
  \centering
  \begin{threeparttable}
      \caption{Detailed Network Structure of FDRN}
      \label{tab:fdrn_Structure}
      \begin{tabularx}{\textwidth}{p{.3\linewidth}<{\centering}|p{.35\linewidth}<{\centering}|p{.35\linewidth}<{\centering}}
          \toprule
          \multicolumn{3}{c}{\textbf{$\mathcal{E}$}} \\
          \midrule
          \makecell[c]{Type} & \makecell[c]{Output Size \\(spatial scales, channels)} & \makecell[c]{\# of Layer} \\
          \midrule
          $7 \times 7$ Conv. Layer & $/2$, 64 & 1 \\
          \midrule
          $3 \times 3$ Residue Block & 1, 64 & 3 \\
          \midrule
          $3 \times 3$ Residue Block & $/2$, 128 & 4 \\
          \bottomrule
      \end{tabularx}
      \vspace{0.02cm}
      \begin{tabularx}{\textwidth}{p{.3\linewidth}<{\centering}|p{.35\linewidth}<{\centering}|p{.35\linewidth}<{\centering}}
          \toprule
          \multicolumn{3}{c}{\textbf{$\mathcal{E}_\text{m}$\textbackslash$\mathcal{E}_\text{s}$}} \\
          \midrule
          \makecell[c]{Type} & \makecell[c]{Output Size \\(spatial scales, channels)} & \makecell[c]{\# of Layer} \\
          \midrule
          $3 \times 3$ Residue Block\tnote{a} & $/2$, 256 & 6 \\
          \midrule
          $3 \times 3$ Residue Block\tnote{b} & $/2$, 512 & 3 \\
          \bottomrule
      \end{tabularx}
      \vspace{0.02cm}
      \begin{tabularx}{\textwidth}{p{.3\linewidth}<{\centering}|p{.35\linewidth}<{\centering}|p{.35\linewidth}<{\centering}}
          \toprule
          \multicolumn{3}{c}{\textbf{$\mathcal{D}_\text{m}$}} \\
          \midrule
          \makecell[c]{Type} & \makecell[c]{Output Size \\(spatial scales, channels)} & \makecell[c]{\# of Layer} \\
          \midrule
          $3 \times 3$ Conv. Layer & 1, 128 & 6 \\
          \midrule
          $4 \times 4$ Deconv. Layer & $2$, 64 & 2 \\
          \midrule
          $3 \times 3$ Conv. Layer & 1, 6 & 1 \\
          \bottomrule
      \end{tabularx}
      \vspace{0.02cm}
      \begin{tabularx}{\textwidth}{p{.3\linewidth}<{\centering}|p{.35\linewidth}<{\centering}|p{.35\linewidth}<{\centering}}
          \toprule
          \multicolumn{3}{c}{\textbf{$\mathcal{D}_\text{s}$}} \\
          \midrule
          \makecell[c]{Type} & \makecell[c]{Output Size \\(spatial scales, channels)} & \makecell[c]{\# of Layer} \\
          \midrule
          $3 \times 3$ Conv. Layer & 1, 512 & 1 \\
          \midrule
          $1 \times 1$ Conv. Layer & 1, 1 & 1 \\
          \bottomrule
      \end{tabularx}
      \vspace{0.02cm}
      \begin{tabularx}{\textwidth}{p{.3\linewidth}<{\centering}|p{.35\linewidth}<{\centering}|p{.35\linewidth}<{\centering}}
          \toprule
          \multicolumn{3}{c}{\textbf{$\mathcal{D}_{f_\text{m}}$\textbackslash $\mathcal{D}_{f_\text{s}}$}} \\
          \midrule
          \makecell[c]{Type} & \makecell[c]{Output Size \\(spatial scales, channels)} & \makecell[c]{\# of Layer} \\
          \midrule
          $3 \times 3$ Conv. Layer & 1, 512 & 1 \\
          \midrule
          $1 \times 1$ Conv. Layer & 1, 6 & 1 \\
          \bottomrule
      \end{tabularx}
      \begin{tablenotes}
          \item[a] $\mathcal{D}_\text{m}$ connects from here.
          \item[b] The branches for $\mathcal{D}_\text{s}$ and $\mathcal{D}_{f_\text{m}}$\textbackslash $\mathcal{D}_{f_\text{s}}$ connect from here.
      \end{tablenotes}
  \end{threeparttable}
\end{table*}

\begin{table*}[!t]
  \centering
  \caption{Matching Performance (\%) with Different Enhancement Methods. FDD \cite{FDD} is retrained based on the enhancement images. Values are reported as relative differences from the original image baseline. Unless otherwise specified, TAR@FAR = 0.1\% is reported.
  }
  \label{tab:enhancement_comp_retrained}
  \renewcommand{\arraystretch}{1.2}
  \begin{threeparttable}
    \resizebox{\textwidth}{!}{
      \begin{tabular}{l
        c*{2}{>{\centering\arraybackslash}p{0.05\linewidth}}
        c*{1}{>{\centering\arraybackslash}p{0.05\linewidth}}
        c*{1}{>{\centering\arraybackslash}p{0.05\linewidth}}
        c*{4}{>{\centering\arraybackslash}p{0.05\linewidth}}
        c*{4}{>{\centering\arraybackslash}p{0.05\linewidth}}}
        \toprule
        \multirow{2}{*}{\textbf{\makecell[l]{Enh.\\Method}}} & \multicolumn{2}{c}{\textbf{NIST SD4}} & \multicolumn{2}{c}{\textbf{N2N Plain}} & \multicolumn{1}{c}{\textbf{FVC02}} & \multicolumn{1}{c}{\textbf{FVC04}} & \multicolumn{1}{c}{\textbf{FVC06}} & \multicolumn{1}{c}{\textbf{PolyU}} & \multicolumn{2}{c}{\textbf{NIST SD27}} & \multicolumn{2}{c}{\textbf{THU Latent10K}} \\
        \cmidrule(lr){2-3} \cmidrule(lr){4-5} \cmidrule(lr){6-9} \cmidrule(lr){10-11} \cmidrule(lr){12-13}  
        & \textbf{Rank-1} & \textbf{TAR}\tnote{$\dagger$} & \textbf{Rank-1} & \textbf{TAR}\tnote{$\dagger$} & \multicolumn{4}{c}{\textbf{TAR}} & \textbf{Rank-1} & \textbf{TAR} & \textbf{Rank-1} & \textbf{TAR} \\
        \midrule
        Orginal Image & 99.75 & 99.60 & 98.65 & 98.75 & 95.86 & 99.50 & 89.62 & 88.62 & 51.94 & 62.02 & 82.46 & 90.79 \\
        \hhline
        VeriFinger\tnote{$\star$} \cite{nist2020verifinger} & \textcolor{red}{-0.35} & \textcolor{red}{-0.55} & \textcolor{red}{-0.95} & \textcolor{red}{-1.20} & \textcolor{red}{-5.5} & \textcolor{red}{-0.68} & \textcolor{red}{-11.03} & \textcolor{red}{-15.41} & \textcolor{red}{-15.51} & \textcolor{red}{-19.38} & \textcolor{red}{-7.78} & \textcolor{red}{-6.55} \\
        FingerNet\tnote{$\star$} \cite{tang2017fingernet} & \textcolor{red}{ -0.30} & \textcolor{red}{ -0.10} & \textcolor{red}{ -0.30} & \textcolor{red}{ -0.55} & \textcolor{red}{ -4.72} & \textcolor{red}{ -0.68} & \textcolor{red}{ -11.03} & \textcolor{red}{ -19.75} & \textcolor{red}{ -6.59} & \textcolor{red}{ -4.27} & \textcolor{red}{ -4.36} & \textcolor{red}{ -3.71} \\
        FingerGAN\tnote{$\star$} \cite{FingerGAN} & \textcolor{red}{ -0.25} & \textcolor{red}{ -0.10} & \textcolor{red}{ -0.30} & \textcolor{red}{ -0.55} & \textcolor{red}{ -21.68} & \textcolor{red}{ -2.57} & \textcolor{red}{ -37.48} & \textcolor{red}{ -10.59} & \textcolor{red}{ -7.75} & \textcolor{red}{ -9.69} & \textcolor{red}{ -5.79} & \textcolor{red}{ -4.83} \\
        UNetEnh & \textcolor{red}{-0.10} & \textcolor{red}{-0.10} & \textcolor{red}{ -0.30} & \textcolor{red}{-0.25} & \textcolor{red}{ -0.29} & \textcolor{red}{ -0.32} & \textcolor{red}{ -1.22} & \textcolor{darkgreen}{ +7.83} & \textcolor{darkgreen}{ +5.81} & \textcolor{darkgreen}{  +5.42} & \textcolor{darkgreen}{ +0.60} & \textcolor{darkgreen}{  +0.19} \\
        PriorEnh & 0.00 & \textcolor{darkgreen}{ +0.05} & \textcolor{red}{-0.05} & \textcolor{red}{ -0.15} & \textcolor{darkgreen}{ +0.07} & \textcolor{red}{ -0.11} & \textcolor{darkgreen}{ +0.01} & \textcolor{darkgreen}{  +7.49} & \textcolor{darkgreen}{ +4.26} & \textcolor{darkgreen}{ +0.38} & \textcolor{darkgreen}{  +0.74} & \textcolor{darkgreen}{+0.70} \\
        \bottomrule 
      \end{tabular}}
      \begin{tablenotes}
        \item[$\dagger$] TAR@FAR=0.01\%.
        \item[$\star$] The FDD \cite{FDD} is retrained using images enhanced by each corresponding method, while the training set remains the same—24,000 pairs of rolled fingerprints from the NIST SD14 dataset.
      \end{tablenotes}
  \end{threeparttable}  
\end{table*}

\end{document}